\RequirePackage{fix-cm}
\documentclass[twocolumn]{svjour3}          
\smartqed  
\usepackage{graphicx}
\graphicspath{{img/jpg/}{img/pdf/}{img/eps/}{img/png/}}
\DeclareGraphicsExtensions{.jpg,.pdf,.eps,.png}
\usepackage[caption=false,font=footnotesize]{subfig}
\usepackage{stfloats}

\usepackage[authoryear]{natbib}
\usepackage{hhline}
\usepackage{siunitx}
\usepackage{hyperref}

\usepackage{algorithmic}
\usepackage[ruled,norelsize,linesnumbered]{algorithm2e}

\usepackage{amsmath}
\usepackage{amssymb}

\usepackage{booktabs}

\usepackage[defaultcolor=blue]{changes}
\usepackage{color}

\begin{document}
\emergencystretch 3.3em

\title{SMORES-EP, a Modular Robot with Parallel Self-assembly
  \thanks{This work was funded by NSF grant number CNS-1329620.}
}


\author{Chao Liu \and
  Qian Lin \and
  Hyun Kim \and
  Mark Yim
}


\institute{Chao Liu, Hyun Kim, and Mark Yim \at
              GRASP Laboratory and The Department of Mechanical
              Engineering and Applied Mechanics\\
              University of Pennsylvania\\
              220 South 33rd St.,\\
              Philadelphia, PA, USA.\\
              \email{\{chaoliu, hkim8, yim\}@seas.upenn.edu}           
           \and
           Qian Lin \at
           The Department of Mechanical Engineering\\
           Massachusetts Institute of Technology\\
           Cambridge, MA, USA.\\
           \email{qalin@mit.edu}
}

\date{Received: date / Accepted: date}

\maketitle

\begin{abstract}
  Self-assembly of modular robotic systems enables the construction of
  complex robotic configurations to adapt to different tasks. This
  paper presents a framework for SMORES types of modular robots to
  efficiently self-assemble into tree topologies. These modular robots
  form kinematic chains that have been shown to be capable of a large
  variety of manipulation and locomotion tasks, yet they can
  reconfigure using a mobile reconfiguration. A desired kinematic
  topology can be mapped onto a planar pattern with optimal module
  assignment based on the modules' locations, then the mobile
  reconfiguration assembly process can be executed in parallel. A
  docking controller is developed to guarantee the success of docking
  processes. A hybrid control architecture is designed to handle a
  large number of modules and complex behaviors of each individual,
  and achieve efficient and robust self-assembly actions. The
  framework is demonstrated in both hardware and simulation on the
  SMORES-EP platform.

\keywords{Cellular and Modular Robots \and Distributed Robot Systems
  \and Path Planning for Multiple Mobile Robots or Agents \and Motion Control}
\end{abstract}

\section{Introduction}
\label{intro}
It is common in nature that groups of individuals can join together to
overcome the limited capability of individuals, especially for insects
who often need to collaborate in large groups to accomplish
tasks. This collective behavior in nature has inspired similar
approaches for robotic systems.  Modular robots are composed of
numerous simple building blocks, or modules, that can be joined into
different connected morphologies depending on task requirements.
Determining the sequence of motions to form a goal configuration of
modules is called {\em self-assembly planning}
\citep{Seo-assembly-planning-icra-2016}.

Self-assembly is related to self-reconfiguration for modular
robots. There are in general three styles of self-reconfiguration
among the variety of self-reconfigurable robot architectures: lattice,
chain and mobile style~\citep{Yim-review-ram-2007}. Lattice style
\citep{Chirikjian-metamorphic-robot-icra-1994} reconfiguration occurs
with modules rearranging themselves to positions on a virtual lattice
while maintaining a single connected component. Chain style
\citep{Yim-polybot-icra-2000} occurs between modules forming kinematic
chains (links are connected by joints) again in a single connected
component. The mobile style \citep{Fukuda-cebot-ras-1991} occurs where
modules can separate from each other and move on the environment
before joining. Self-assembly is most related to the mobile style
reconfiguration as pieces are assembled as separate units.

A particularly useful self-assembly example collaborative behavior
can be shown for systems that can achieve mobile reconfiguration. Each
individual module can locomote around an area, however it may discover
it cannot cross a gap that is larger than one module. Instead multiple
modules can join to form a snake configuration making it long enough
to cross over. Other applications include reaching tall spaces or
rapid simultaneous exploration. All of these behaviors are similar to
the behaviors shown in ants.

An individual module usually has limited motion capability due to its
design. Our goal is to allow modular robots to self-assemble into
desired morphologies to perform complex tasks, such as dexterous
manipulation, climbing stairs, navigation over complex terrains, and
quick exploration. An individual module is not capable of handling
these tasks because of its motion limitations.  There are several
challenges needed to be addressed:
\begin{enumerate}
\item Efficiency is important, especially for modular robotic systems
  that include a large number of modules;
\item Physical constraints have to be considered for real hardware
  applications;
\item Accurate docking is required.
\end{enumerate}
Many modules can be involved in order to handle various scenarios and
an efficient framework is important. The framework should output
efficient assembly sequences and also provide an efficient
architecture to control modules so that actions can be executed
efficiently. A motion controller is necessary to accurately command
multiple modules to move around with the hardware performance and
motion constraints being considered. Docking actions have to be
carefully executed and some connection can be coupled with module
motions.

One example of a modular reconfigurable robot is the
SMORES~\citep{Davey-smores-iros-2012} and SMORES-EP systems. SMORES-EP
uses EP-Face connectors~\citep{Tosun-ep-face-iros-2016}. Each SMORES
module has four active rotational degrees-of-freedom (DOF), pan, tilt,
and left/right wheels. It has differential wheeled drive using its
left and right wheels.  The SMORES modules are unique among
self-reconfigurable systems in that they can reconfigure in any of the
three styles noted earlier. The modules are individually mobile with
the wheels. The system can form serial chains which has been shown to
be one of the more useful configurations for doing things like
manipulation of objects~\citep{Liu-control-planning-irc2020} or
different styles of locomotion.  While lattice style reconfiguration
mechanisms have been studied in
depth~\citep{Stoy-lattice-construct-2006}, the mobile form of
reconfiguration similar to self-assembly has not been.

In this paper, a parallel self-assembly algorithm for kinematic
topologies (the connectivity of the modules) as well as controllers
for hardware execution are presented using SMORES-EP modules. A target
kinematic structure can be simply defined as a graph where each vertex
is a module and each edge is the connection between adjacent
modules. Given the current locations of all modules, every individual
is mapped to a module in the target configuration in an optimal way
with respect to some predefined cost (e.g. the total moving distance)
by solving a task assignment problem. Then the assembly actions can be
executed in parallel for efficiency while satisfying the physical
hardware constraints. Docking is difficult for modular robots and a
motion controller is also developed to guarantee the success of
docking process. The control architecture of our system is hybrid:
distributed at low level actions and centralized at high level
planning. In our system, each SMORES-EP module has its own processor
to handle hardware-level control and communication, and a central
computer is used for tracking modules' poses and high-level control
and planning. This architecture makes use of distributed computing
power and also achieves efficiency for communication and planning
through centralized information processing. The effectiveness and
robustness of the framework is demonstrated in both the simulation
environment and the real world. Three tasks are presented and every
task contains two simulation tests and one hardware test. The proposed
framework can be easily extended to other mobile type modular robots
as long as the topology is defined.

The paper is organized as follows. Section~\ref{sec:related} reviews
relevant work on self-assembly with modular
robots. Section~\ref{sec:hardware} introduces the hardware and
software we developed to achieve robust self-assembly
activities. Section~\ref{sec:config-task} discusses some concepts and
mathematical models that are necessary to describe a modular robot
configuration and claims our self-assembly problem. The parallel
assembly algorithm is presented in Section~\ref{sec:assembly} as well
as the controller to ensure the docking process. Three hardware
experiments are shown in Section~\ref{sec:exp} to demonstrate the
effectiveness of our framework. Finally, Section~\ref{sec:conclude}
concludes and presents future work.

\section{Related Work}
\label{sec:related}

Modular robots have been an active research field over several
decades, and a general review of these systems is done
by~\cite{Seo-review-modular-robots-annurev-2019}.

\subsection{Modular Robots with Self-assembly}
\label{sec:systems}

Various modular robotic systems have been developed with self-assembly
capability. For mobile self-reconfiguration, CEBOT is one of the
first~\citep{Fukuda-cebot-ras-1991}, with a heterogeneous system
composed of modules (cells) with different functions, such as mobile
cells, bending joint cells, rotating joint cells, and sliding joint
cells. The mobile cells are able to approach stationary cells for
docking using a cone-shaped coupling mechanism so as to further
self-assemble into a manipulation structure. However, its flexibility
is limited to the single function and motion capability of each
module. SMC Rover~\citep{Motomura-smc-rover-ar-2005} is a similar
heterogeneous system that consists of one main body and multiple wheel
units. The main body is stationary and equipped with some functional
components (e.g. solar battery cells). Each wheel unit has a wheel for
locomotion and an arm for manipulation. These wheel units can dock
with the main body to form a vehicle or undock to execute missions by
themselves.

Gunryu~\citep{Hirose-gunryu-assembly-ras-1996} is a robotic system
composed of multiple mobile vehicles and each vehicle has an arm
installed as a connecting mechanism. These vehicles can be connected
as snake-like robots to enhance the locomotion capability of the
system, such as moving over a cliff and climbing stairs. A similar
system is presented in~\citep{Brown-millibot-tmech-2002} but using a
simpler docking mechanism. Their docking capability can extend their
applications but the morphologies formed are limited to single chain
topology. Swarm-bot~\citep{Gross-swarmbot-assembly-conf-2006} modules
also use robotic arms as a connecting mechanism, but are equipped with
a surrounding ring for mating with grippers, which enables a module to
link with more than one modules. However, the assembly behavior is
limited to simple 2-dimensional structure for cooperative locomotion
due to the shape of its modules. They cannot form dextrous kinematics
structures to perform manipulation tasks
(e.g.~\cite{Liu-control-planning-irc2020}) or locomotion behaviors
(e.g.~\cite{Kamimura-mtran-locomotion-icra-2003}).

Some modular robots are designed mainly for forming two-dimensional
patterns or building structures by self-assembly. Units in regular
shape (cube or sphere) but without locomotion ability are presented to form
two-dimensional or three dimensional lattice structures in stochastic
ways
~\citep{White-stochastic-celluar-rss-2005,Haghighat-lily-assembly-icra-2015}.
A similar system performing two-dimensional stochastic self-organizing
is introduced in~\citep{Klavins-self-organize-2006}. Larger scale
planar systems with nearly one thousand modules have been shown with
kilobots~\citep{Rubenstein-kilobot-icra-2012} to generate different
planar patterns, and each module has very limited locomotion
ability. Tactically Expandable Maritime Platform (T.E.M.P.) is a
system with a large number of modular
boats~\citep{Darpa-temp-icra-2014}. These boats can self-assemble into
modular sea bases in a brick wall pattern. A similar idea was
implemented using quadrotors which can form floating structures in
midair~\citep{modquad-hardware-icra-2018}. These systems are
mainly for building structures and not
applicable to complex locomotion and manipulation tasks.

SUPERBOT is a hybrid modular robotic system with limited locomotion
capability of each module~\citep{Salemi-superbot-iros-2006}. A SUPERBOT
module is formed by two linked cubes with three twist DOFs, which
allows a module to crawl on the ground. This design makes
self-assembly possible but also challenging because it is difficult to
control the pose of the module body. Modular clusters of CKBot modules
are able to perform self-reassembly after
explosion~\citep{Yim-ckbot-assembly-iros-2007}. Each cluster consists
of four CKBot modules~\citep{Yim-ckbot-2009}, a camera module, and
magnet faces. Similar to SUPERBOT, each cluster can execute snake-like
locomotion which makes the self-assembly more difficult than wheeled
locomotion. A heterogeneous modular robotic system for self-assembly
is shown by~\cite{Liu-decentralized-self-assembly-2014}. Modules are
equipped with different locomotion actuators but cannot perform gait
locomotion and manipulation tasks. {\color{black}Sambot~\citep{Wei-sambot-tmech-2014} system, similar to Swarm-bot, is composed of multiple mobile robots with an active docking mechanism and a bending joint.} The cubic shape allows more dexterous morphologies than Swarm-bot, such as a walker or a manipulator.

\subsection{Docking Interface and Execution}
\label{sec:docking-design}

Docking is a necessary, but usually difficult task for modular robots,
which occurs at certain module faces, or connectors. There are many
connector designs that can be either gendered or ungendered. The
general requirements for modular robot docking interfaces are high
strength, high speed of docking/undocking, low power consumption, and
large area-of-acceptance~\citep{Eckenstein-xface-iros-2012}.

Mechanical devices or structural hook-type connectors are widely
designed for docking activities. Robotic arms and grippers are used in
SMC Rover~\citep{Motomura-smc-rover-ar-2005},
Gunryu~\citep{Hirose-gunryu-assembly-ras-1996}, and
Swarm-bot~\citep{Gross-swarmbot-assembly-conf-2006}. The mechanical
design is straightforward and easy to be strong, but difficult to be
compact, which limits the robot flexibility. {\color{black}Structural
  hook-type connector are used in CEBOT~\citep{Fukuda-cebot-ras-1991},
  Millibot trains~\citep{Brown-millibot-tmech-2002},
  T.E.M.P.~\citep{Darpa-temp-icra-2014}, M-TRAN
  III~\citep{Murata-mtran-visual-docking-iros-2006}, and
  Sambot~\citep{Wei-sambot-tmech-2014}.} The SINGO
connector~\citep{Shen-singo-connector-icra-2009} for SUPERBOT is
genderless (there is no polarity and any two connectors can be mated),
and capable of disconnection even when one module is unresponsive,
allowing for self-repair. These docking systems are mechanically
complicated, and usually require large amount of space. Accurate
positioning is also crucial for these connectors resulting in small
area-of-acceptance. The design complexity may also cause failure over
time.

Magnets exist in many modular robotic systems for docking. Permanent
magnet interfaces are easy to implement and have relatively large
area-of-acceptance. Modules can be easily docked as long as they are
close to each other within some distance, with the misalignment
adjusted automatically. However, extra actuation is needed for
undocking. M-TRAN~\citep{Murata-mtran-trm-2002} uses permanent magnets
for latching, and undocks using shape memory alloy (SMA) coils to
generate the required large force. However, it takes minutes for these
coils to cool leading to slow response, and this connector is not
energy efficient. Permanent magnets are also used in programmable
parts~\citep{Klavins-self-organize-2006} for latching. This docking
interface is gendered. The original SMORES system utilizes permanent
magnets as docking interfaces as well and a unique docking key for
undocking~\citep{Davey-smores-iros-2012}. Permanent magnets are placed
at the frame corners of a flying system called ModQuad. Here modules
dock by adjoining the frame corner magnets and undock with aggressive
maneuvers of vehicle
structures~\citep{Saldana-modquad-undocking-ral-2019}.

The docking interface design can significantly affect the docking
execution. In general, we aim to make the level of positioning
precision required for docking as low as possible, or extra
high-quality sensors are needed to provide precise pose
feedback. Connector area-of-acceptance and some preferred properties
are studied by~\cite{Eckenstein-connector-acceptance-iros-2017}
and~\cite{Nilsson-connector-analysis-icra-2002}. Infrared (IR) sensors
are installed on the docking faces for assisting alignment of
hook-type connectors, such as
PolyBot~\citep{Yim-polybot-docking-tmech-2002},
CONRO~\citep{Rubenstein-conro-docking-icra-2004}, and the
heterogeneous system in SYMBRION
project~\citep{Liu-decentralized-self-assembly-2014}. Two mobile
robots are shown to dock using a visual-based system using a
black-and-white camera to track a visual
target~\citep{Bererton-visual-docking-2001}. Vision-based approaches
were developed for M-TRAN
III~\citep{Murata-mtran-visual-docking-iros-2006} by placing a camera
module on a cluster of modules that can detect LED signals. However,
guidance by this visual feedback is not sufficient to realize
positional precision for its mechanical docking interface and extra
efforts from the cluster are required for connection. Similarly,
Swarm-bot applies a omni-directional camera to detect docking slots
with a ring of LEDs, but cannot guarantee the success of docking. In
comparison, the docking process can be easier with magnet-based
connectors. Using a similar visual-based solution with M-TRAN III,
CKBot clusters with magnet faces show more robust and easier docking
procedures~\citep{Yim-ckbot-assembly-iros-2007}. The large
area-of-acceptance of magnet docking faces can also be seen by
stochastic self-assembly modular
robots~\citep{
White-stochastic-celluar-rss-2005,
  Haghighat-lily-assembly-icra-2015}.

\subsection{Self-assembly Algorithm}
\label{sec:algorithm-review}

Stochastic self-assembly algorithms are for modules that are not
self-actuated but use external actuation, such as fluid flow
{\color{black}{\citep{
White-stochastic-celluar-rss-2005,Klavins-self-organize-2006,
Mermoud-stochastic-lily-assembly-icra-2012,Fox-probabilistic-assembly-tac-2015,Haghighat-lily-stochastic-assembly-2017}}}. Static
planar structures can be assembled by building blocks which are
supplied by mobile
robots {\color{black}\citep{Werfel-3d-construct-science-2014}}. Self-assembly
solutions for mobile-type modular robots typically use modules that do
not have non-holonomic constraints and aim for planar structures
{\color{black}{\citep{Klavins-self-assembly-2002,Murray-epuck-assembly-2013,Seo-assembly-planning-icra-2016,Li-assembly-2d-2016,Saldana-decentralized-assembly-2d-iros-2017}}}. An
approach to solve configuration formation is presented
by~\cite{Nelson-formation-ar-2019} but in sequential manner which
makes the formation process slow. Assembling structures in
3-dimensional space is shown by~\cite{Werfel-3d-construct-ijrr-2008}
and~\cite{Tolley-stochastic-assembly-ijrr-2011} but does not deal with
the physical constraints of land-mobile platforms.

Our work differs from structural self-assembly in that the assembly
goal is to build a movable kinematic topology, such as a multi-limbed
form. The target morphologies are similar to those built by chain-type
modular robots. For example, reconfiguration for Polybot in a
kinematic topology was presented
by~\cite{Yim-chain-reconfiguration-2000}. However in chain-type
reconfiguration, modules remain connected in a single connected
component during the process. For lattice-type modular robots,
reconfiguration usually happens in 3-dimensional
space~\citep{Brandt-atron-lattice-reconfig-iros-2006}, but these
techniques are not applicable to the self-assembly problems of mobile
class of modular
robots. \citet{Daudelin-perception-modular-robot-scirobotics-2018}
showed some simple self-assembly behaviors using SMORES-EP modules in
which only two modules need to move and dock, and the planning and the
docking are manually created and not efficient. The lack of robustness
also leads to high possibility of failure. A distributed mobile-type
reconfiguration algorithm for SMORES-EP is introduced
by~\cite{Liu-smores-reconfig-ral-2019} which focuses on topology
reconfiguration while minimizing the number of reconfiguration
actions, though it does not consider the hardware control and
planning. In this paper, we extend the work
by~\cite{Liu-smores-reconfig-ral-2019}. We not only address navigation
and docking control and planning, but also demonstrate the algorithm
on the real hardware system with multiple experiments to show its
effectiveness and robustness. The parallel self-assembly framework can
also be used to make the sequential reconfiguration actions more
efficient by executing these actions in parallel.

\section{Hardware and Software Architecture}
\label{sec:hardware}

\begin{figure}[t]
  \centering
  \includegraphics[width=0.4\textwidth]{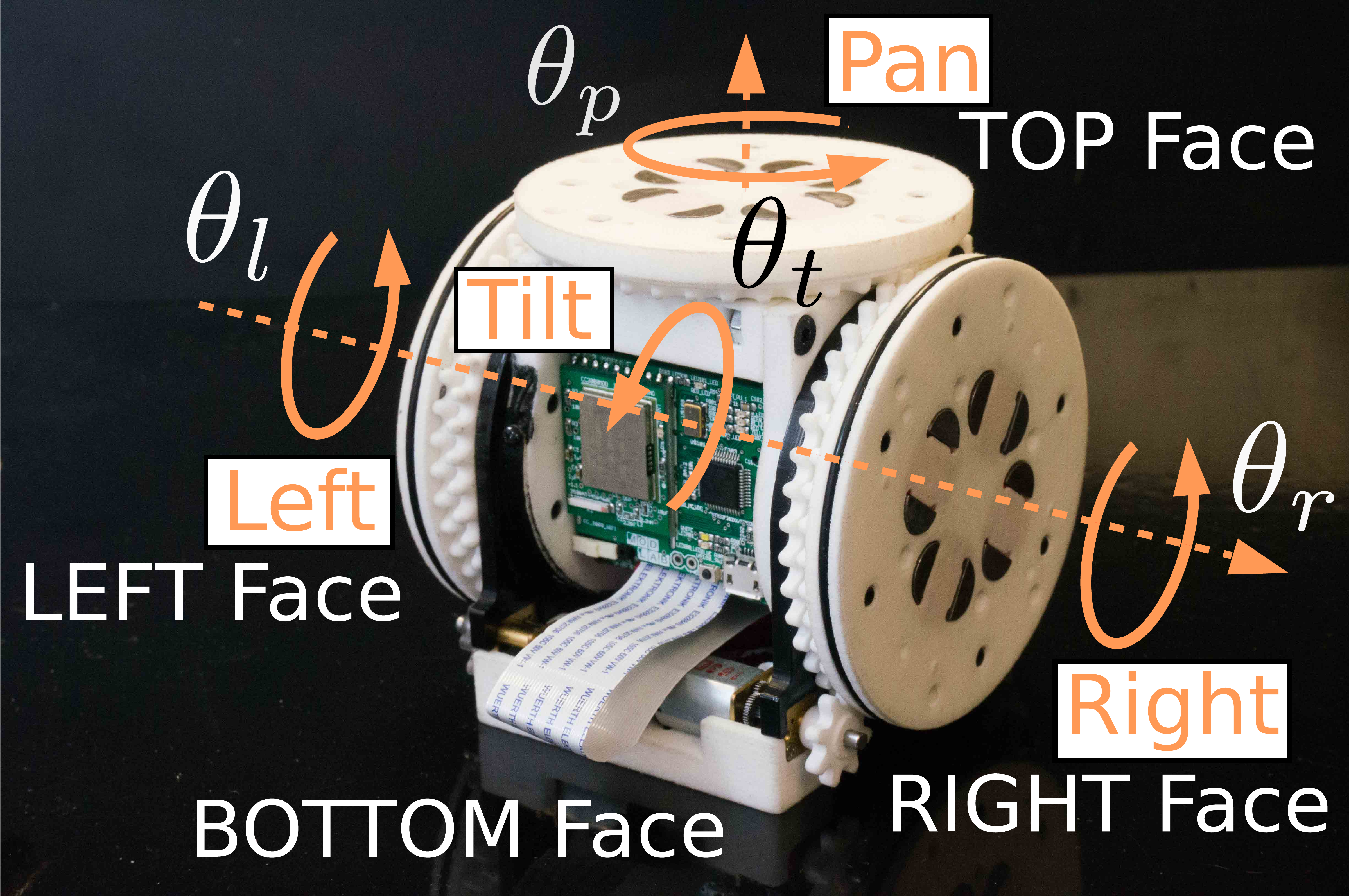}
  \caption{A SMORES-EP module has four active rotational
    degrees of freedom and four connectors using an array of
    electro-permanent (EP) magnets.}
  \label{fig:smores-module}
\end{figure}

\begin{figure}[t]
  \centering
  \begin{subfloat}[]{
      \includegraphics[width=0.39\linewidth]{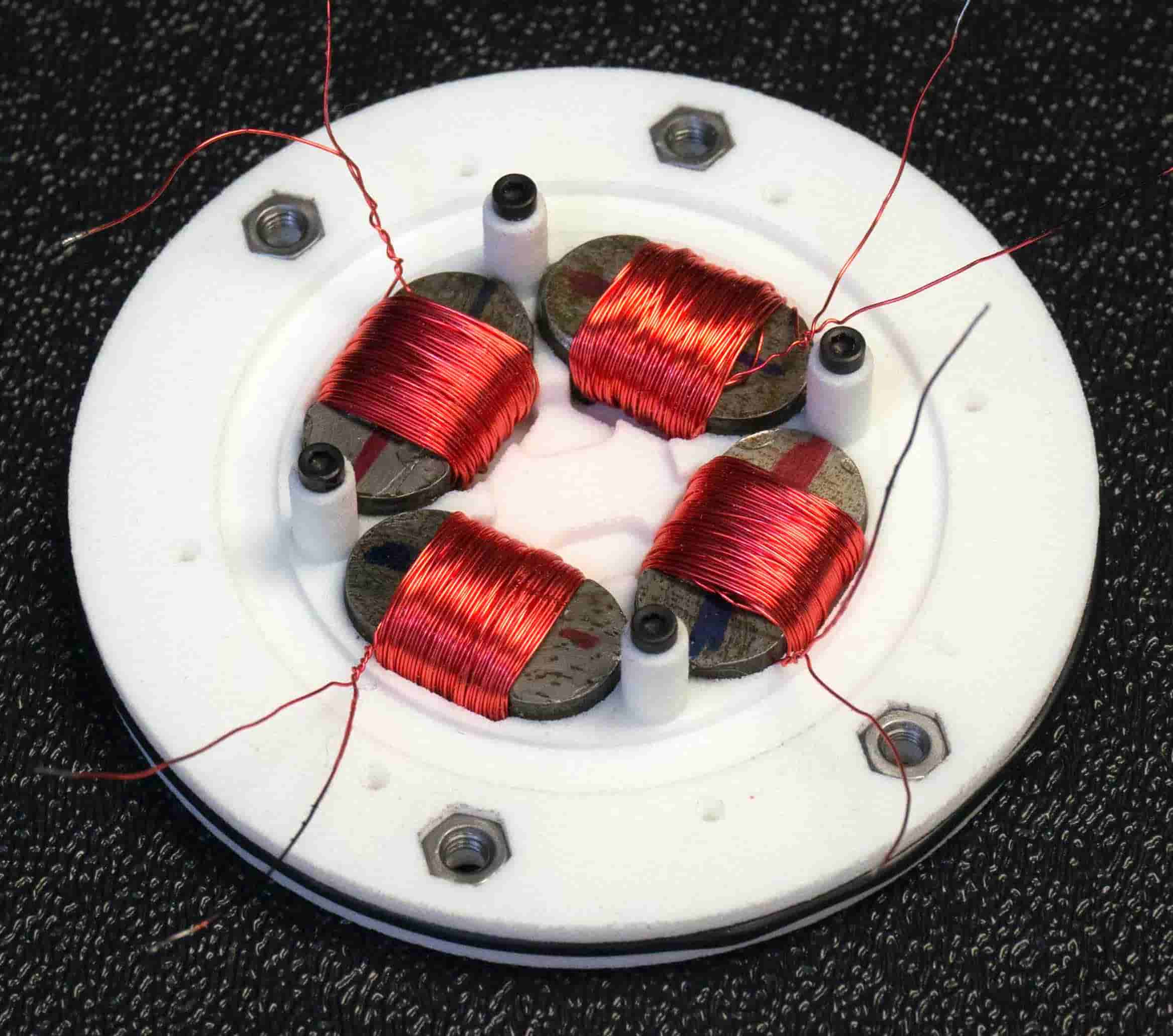}\label{fig:magnets}}
  \end{subfloat}
  \hfil
  \begin{subfloat}[]{
      \includegraphics[width=0.5\linewidth]{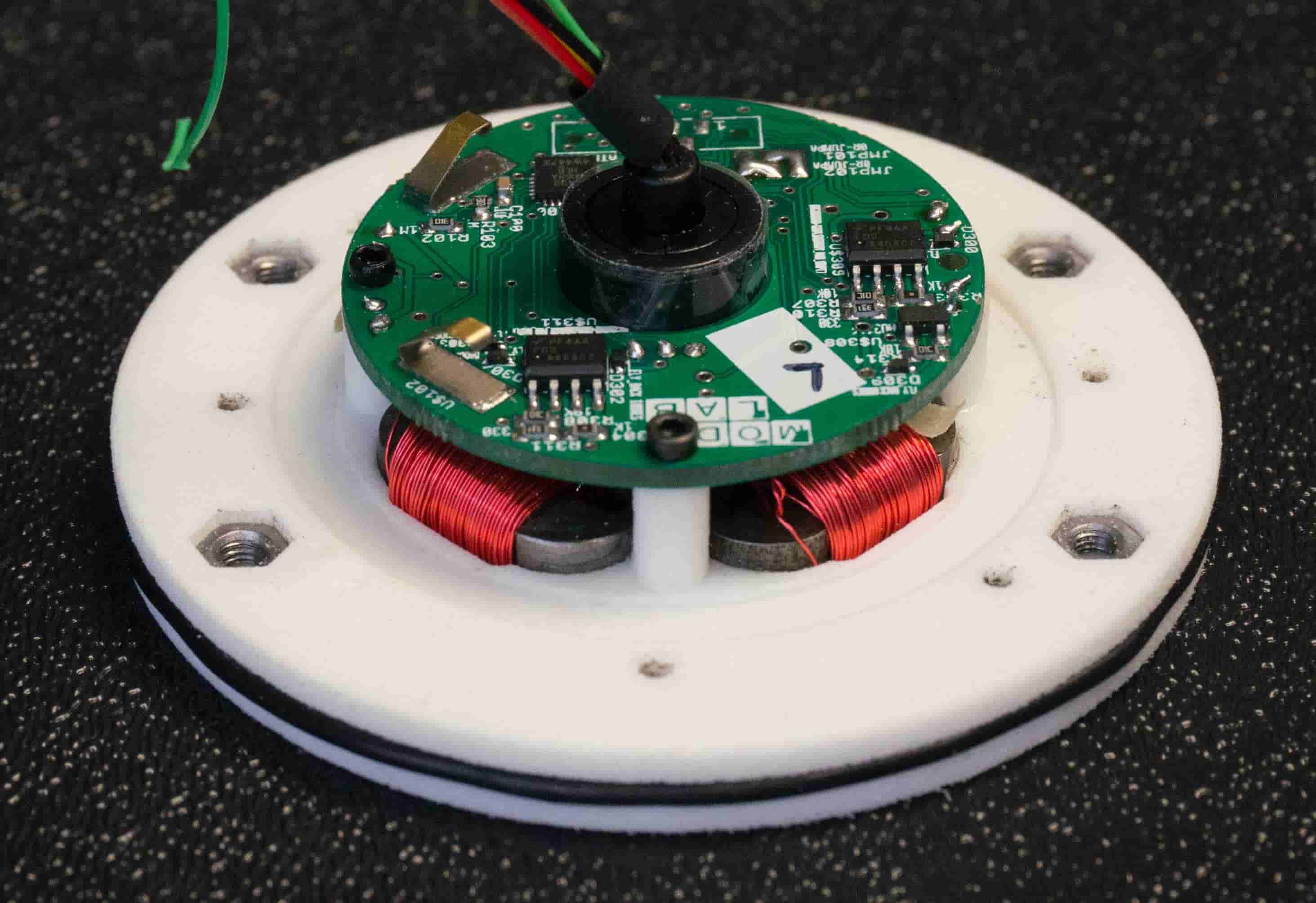}\label{fig:ep-face}}
  \end{subfloat}
  \caption{(a) Internal view of the magnets in the EP-Face. (b) Internal view
    of the EP-Face with circuit board.}\label{fig:smores-face}
\end{figure}

The hardware platform SMORES (\textbf{S}elf-assembly \textbf{MO}dular
\textbf{R}obot for \textbf{E}xtreme \textbf{S}hape-shifting) is first
presented by~\cite{Davey-smores-iros-2012} and SMORES-EP is the
current version where EP refers to the Electro-Permanent
magnets~\citep{Knaian-ep-magnets-mit-2010} as its
connector~\citep{Tosun-ep-face-iros-2016}. The magnets can be switched
on or off with pulses of current unlike electromagnets which require
sustained currents while on. Each module is a four-DOF
(\textit{LEFT DOF}, \textit{RIGHT DOF}, \textit{PAN DOF}, and
\textit{TILT DOF}) system with four connectors (\textit{LEFT Face} or
\textbf{L}, \textit{RIGHT Face} or \textbf{R}, \textit{TOP Face} or
\textbf{T} and \textit{BOTTOM Face} or \textbf{B}) shown in
Fig.~\ref{fig:smores-module}, and it is the size of an
80-\SI{}{mm}-wide cube. In particular, LEFT DOF, RIGHT DOF, and PAN DOF
can continuously rotate to produce a twist motion of docking ports
relative to the module body, and TILT DOF is limited to
$\pm$\SI{90}{\degree} to produce a bending joint. LEFT DOF and RIGHT
DOF can also be used as driving when wheels doing differential drive
locomotion. A customizable sensor as well as its estimation method
are used for tracking the position of each
DOF~\citep{Liu-paintpot-sensing-jmr-2021}. Each connector, called
EP-Face, is equipped with an array of EP magnets arranged in a ring,
with south poles counterclockwise of north as illustrated in
Fig.~\ref{fig:smores-face}.

The EP magnet is driven by three pulses under \SI{11.1}{V} battery
voltage.  A pair of EP-Face connectors can provide as much as
\SI{90}{N} to maintain their docking status and very little energy is
consumed to connect with or disconnect from each other.  Magnetic
forces can draw two modules together through a gap of \SI{4}{mm}
normal, and \SI{7}{mm} parallel to the faces leading to a large area
of acceptance.

Each module has an onboard processor, a Wi-Fi module, and a battery
for hardware-level control and communication. This local processor
handles all EP-Face connectors (generating pulses), all DOFs' position
estimation and control (running four Kalman filters at the same time),
and the Wi-Fi communication with the central computer (UDP
protocol). On the central computer, a robot manager is running for
communication with all modules, including fetching information from an
individual module, sending commands to control EP-Face connectors and
motion of every DOF. Messages related to motion control are
transmitted between the central computer and every module at around
\SI{40}{Hz} mainly for two purposes: (1) reporting the positions of
all involved DOFs when executing manipulation tasks; (2) sending
velocity commands of the left and right wheel when doing differential
drive locomotion. This allows real-time control of a cluster of
modules which is important to handle local scenarios that may cause
collision among moving modules efficiently. All other information and
commands, such as activating an EP-Face connector for docking and
moving a DOF to a specific position, are transmitted as required by
either programs or users.

In our experiment setup, we use a VICON motion capture system to
localize module poses as they are moving on the ground. The system can
provide precise pose estimation at \SI{100}{Hz}. This position data is
used for the navigation of the modules moving as differential-drive
vehicles.

\section{Robot Configuration and Assembly}
\label{sec:config-task}

This section introduces a graph model to describe the topology of a
modular robot system, the definition of our target topology, and the
self-assembly problem.

\subsection{Modular Robot Topology Configuration}
\label{sec:config}

A graph model of modular robot topology was presented
by~\cite{Liu-config-recognition-isrr-2017}. A modular robot topology
can be represented as a graph $G = (V, E)$ where $V$ is the set of
vertices of $G$ representing all modules and $E$ is the set of edges
of $G$ representing all the connections among modules. Graphs with
only one path between each pair of vertices are \textit{trees}. It is
convenient to start with tree topology and, if a configuration has
loops, it can be converted into an acyclic configuration by running a
spanning tree algorithm. Therefore, this work only focuses on
configurations in tree topology.

A tree $G = (V, E)$ can be rooted with respect to a vertex
$\tau\in V$. Given a modular robot configuration in tree topology, the
root is selected as the center of the graph defined
by~\cite{McColm-acyclic-graph-2004}. Let $\mathcal{C}$ be the set of
all connectors of a module that is
$\{\mathrm{\textbf{L}}, \mathrm{\textbf{R}}, \mathrm{\textbf{T}},
\mathrm{\textbf{B}}\}$, and $\mathrm{CN}^v(c)$ denotes the total
number of modules connected to module $v$ via its connector
$c\in \mathcal{C}$. Note that a configuration graph $G$ can be rooted
with respect to any module but $\mathrm{CN}^v(c)$ is invariant under
the selection of the root module. For any vertex $v\in V$ that is not
a leaf with respect to root $\tau$, we denote its child connected via
its connector $c$ as $\hat{v}^c$, the mating connector of $\hat{v}^c$
as $\hat{c}^\prime$, the set of its children as
$\mathcal{N}(v, \tau)$, and the set of $c$ connected with its children
as $\mathcal{C}_d(v)\subseteq\mathcal{C}$. Then the root module $\tau$
has to satisfy the following condition
\begin{equation}
  \label{eq:root_condition}
  \mathrm{CN}^\tau (c) \le \frac{1}{2} \vert V\vert,\ \forall c\in \mathcal{C}
\end{equation}

\begin{algorithm}[t]
  \caption{Root Module Search}\label{alg:root}
  \KwIn{Graph representation $G = (V,E)$}
  \KwOut{Root module $\tau$}
  Root $G = (V,E)$ with respect to a module $v_0$ with height
  of $H$ and the height of $v\in V$ is $h(v)$\;
  Initialize $\mathrm{CN}^v(c)$ for $v\in V$ and $c\in \mathcal{C}$ to
  be zero\;
  Initialize $\mathbf{h} \gets 1$\;
  \While{$\mathbf{h}\le H$}{
    \ForEach{module $v$ with $h(v) = \mathbf{h}$}{
      \ForEach{module $\hat{v}^c\in \mathcal{N}(v,v_{0})$}{
        $\mathrm{CN}^v(c) \leftarrow \sum_{\hat{c}\in \mathcal{C}-\hat{c}^\prime}
        \mathrm{CN}^{\hat{v}^c}(\hat{c}),\ c\in \mathcal{C}_d(v)$\;
        $\mathrm{CN}^{\hat{v}^c}(\hat{c}^\prime) \leftarrow \vert V\vert - 1 - \sum_{\hat{c}\in
          \mathcal{C}_d(\hat{v})}\mathrm{CN}^{\hat{v}^c}(\hat{c})$\;
      }
    }
    $\mathbf{h} \gets \mathbf{h} + 1$;
  }
  $\tau\leftarrow v\in V$ such that $\mathrm{CN}^{\tau}(c)\le \frac{1}{2}\vert V\vert$ $\forall c\in
  \mathcal{C}$
\end{algorithm}
A linear-time algorithm to compute the root of a modular robot
configuration is shown in Algorithm~\ref{alg:root}. We first root the
graph $G$ with respect to a randomly selected module $v_0$. Then the
root module search can be formulated as a dynamic programming problem
that can be solved in time $\mathcal{O}(|V|)$. The \textit{height} of
a vertex (module) $v$ in $G=(V,E)$ denoted as $h(v)$ is the number of
the edges of the longest downward path to a leaf from $v$ and the
\textit{height} of a rooted graph is the height of its root. This
bottom-up algorithm starts from vertices whose height are
one. Detailed derivation can be found in the work
by~\cite{Liu-config-recognition-isrr-2017}.

\begin{figure}[b]
  \centering
  \subfloat[]{\includegraphics[width=0.2\textwidth]{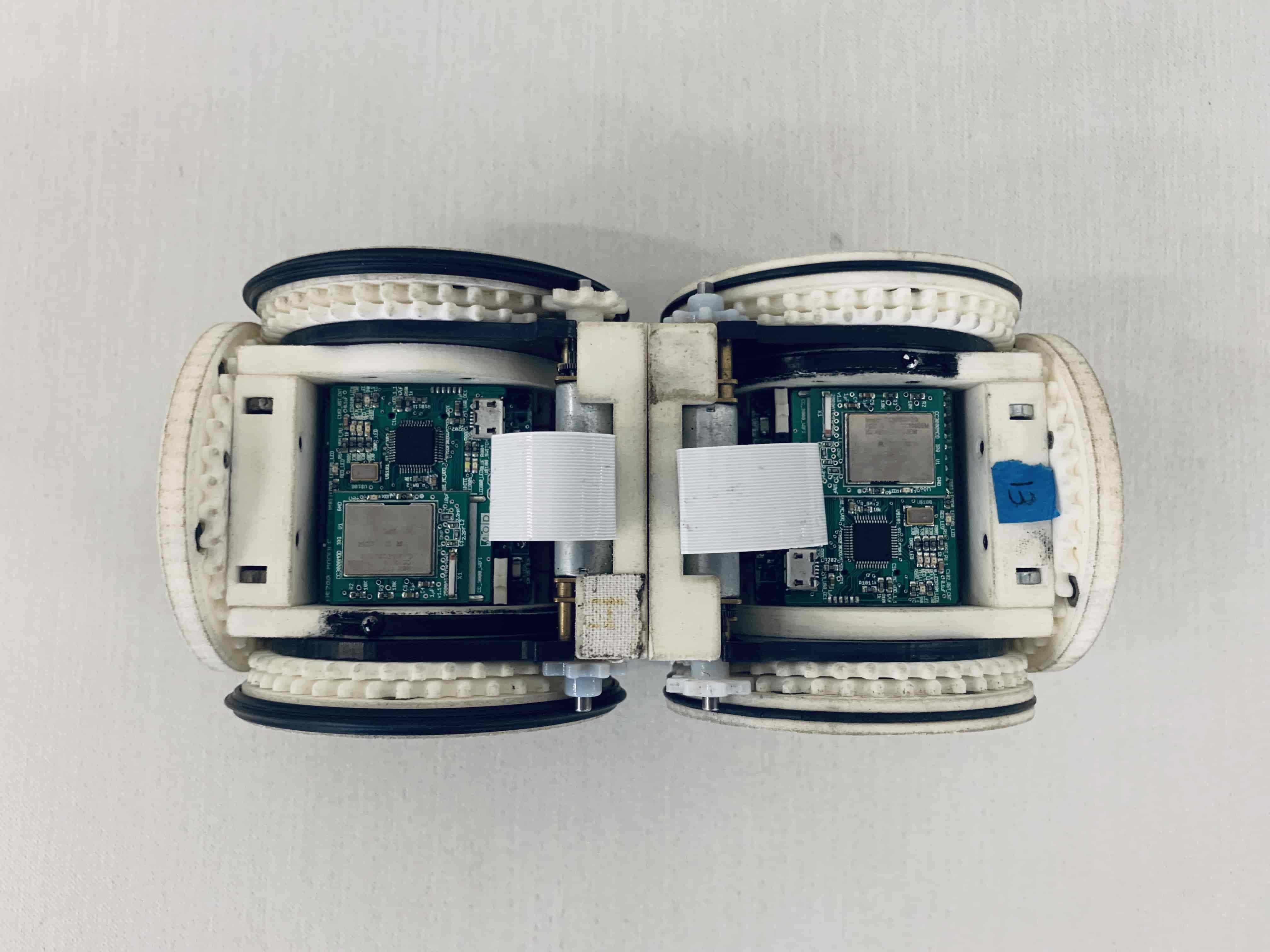}\label{fig:bb-0}}
  \hfil
  \subfloat[]{\includegraphics[width=0.2\textwidth]{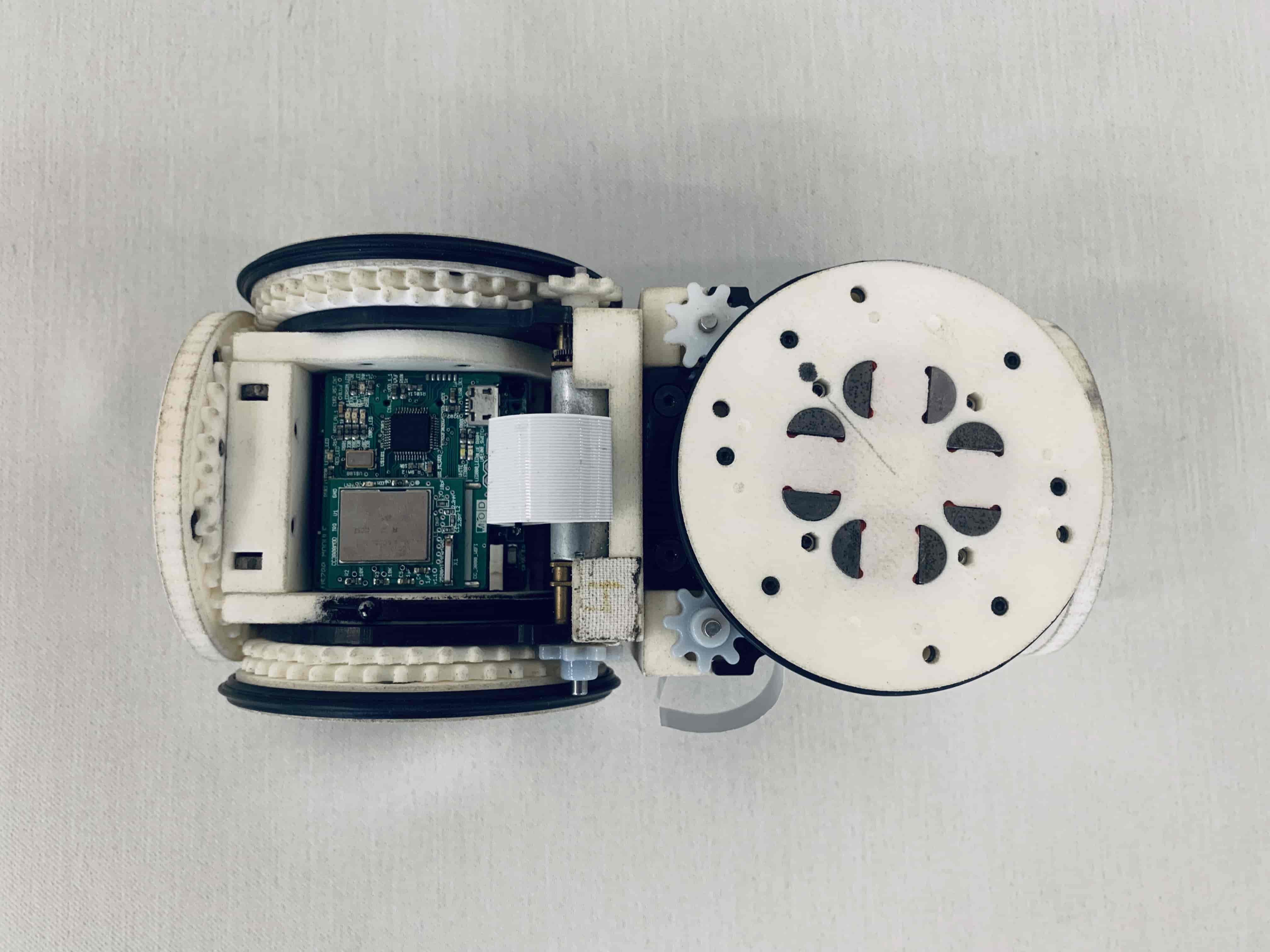}\label{fig:bb-1}}
  \caption{Connection between two BOTTOM Faces: (a)
    \textit{Orientation} is 0 and (b) \textit{Orientation} is 1.}
  \label{fig:smores-bottom}
\end{figure}

The \textit{depth} of a vertex (module) $v$ in $G=(V,E)$ denoted as
$d(v)$ can be defined as the number of the edges from $v$ to the root
$\tau$. There are multiple ways to connect module $u$ and module $v$
denoted as $\mathrm{connect}(u,v)$ from module $u$'s point of view and
$\mathrm{connect}(v, u)$ from module $v$'s point of view. Each
connection has three attributes: \textit{Face}, \textit{Face2Con}, and
\textit{Orientation}~\citep{Liu-config-recognition-isrr-2017}. Some
seemingly different connections are actually equivalent in
topology. In a SMORES-EP configuration, attribute \textit{Orientation}
is trivial for the connections among LEFT Face, RIGHT Face, and TOP
Face since the faces can rotate about the symmetry axis. However
connections between two BOTTOM Faces cannot rotate, so
\textit{Orientation} mounted at \SI{90}{\degree} increments (two of
which are symmetric) needs to be considered
(\textit{Orientation}$\in \left[0, 1\right]$) shown in
Fig.~\ref{fig:smores-bottom}.

\begin{figure}[b!]
  \centering
  \subfloat[]{\includegraphics[height=0.15\textwidth]{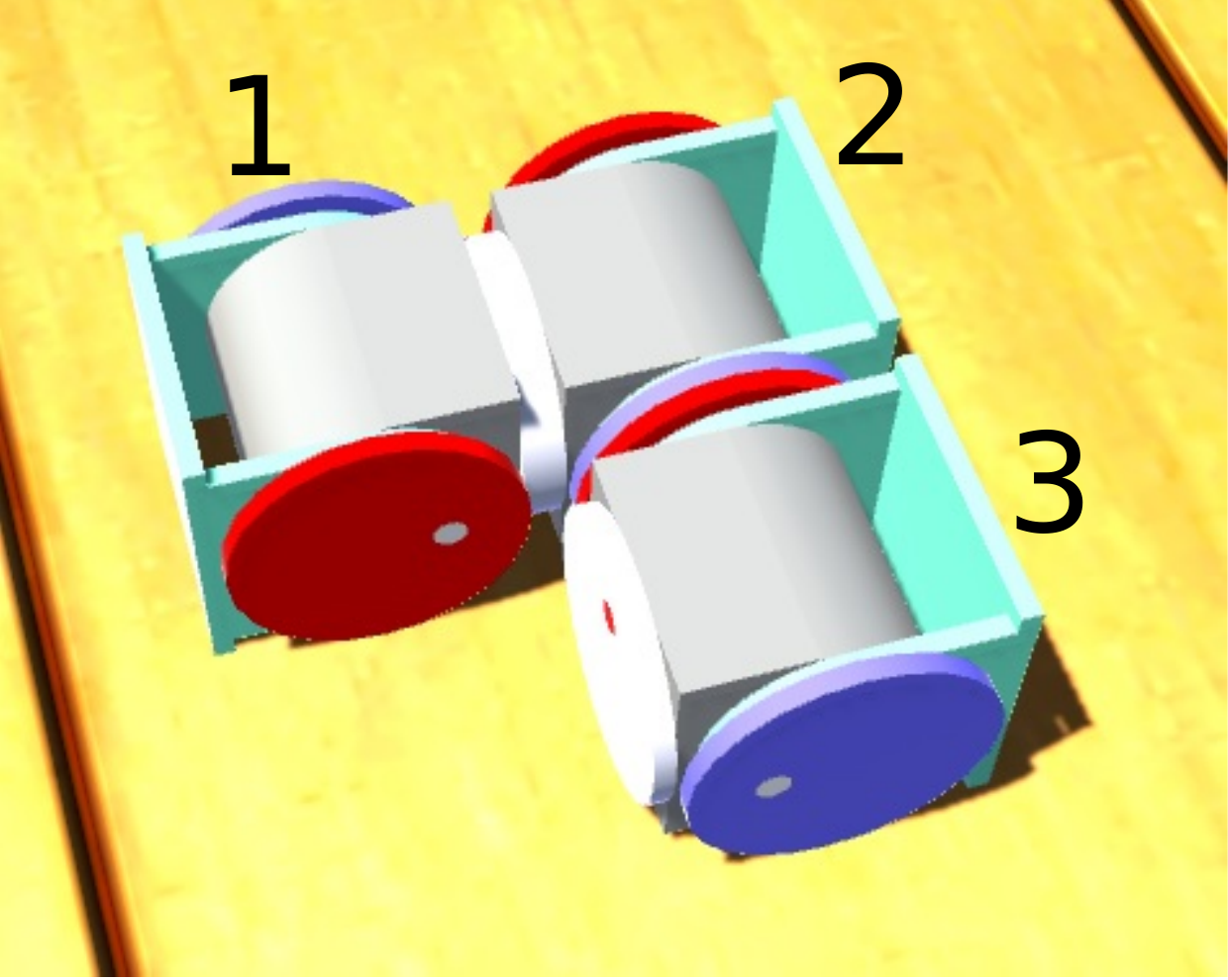}\label{fig:3module}}
  \hfil
  \subfloat[]{\includegraphics[height=0.15\textwidth]{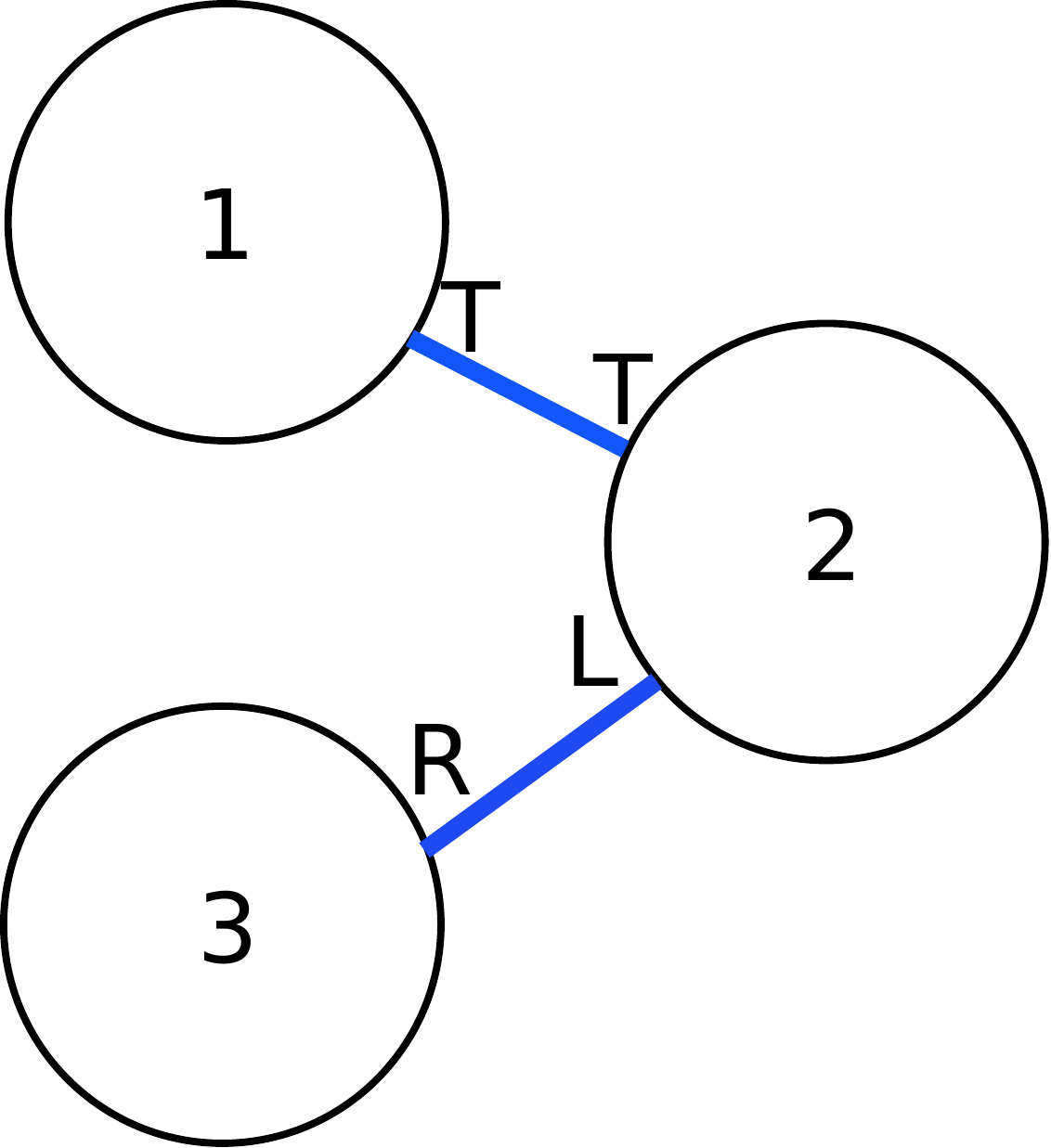}\label{fig:3module-graph}}
  \caption{(a) A topology is built by three SMORES-EP modules. (b) The
    equivalent graph representation of the topology.}
\end{figure}

From here a SMORES-EP configuration can be fully defined. For example,
a three-module configuration in a simulation
environment~\citep{Tosun-modular-robot-design-isrr-2015} is shown in
Fig.~\ref{fig:3module} and its graph representation is shown in
Fig.~\ref{fig:3module-graph}. The connections in this topology can be
written as:
\begin{align*}
  \label{eq:connect-3-modules}
  \mathrm{connect}(1,2) =& \{\mathrm{Face}: \mathrm{TOP\ Face},\\
                         \mathrm{Face2}&\mathrm{Con}: \mathrm{TOP\ Face},
                          \mathrm{Orientation}: \mathtt{Null}\}\\
  \mathrm{connect}(2,1) =& \{\mathrm{Face}:
                           \mathrm{TOP\ Face},\\
                           \mathrm{Face2}&\mathrm{Con}: \mathrm{TOP\ Face},
                           \mathrm{Orientation}: \mathtt{Null}\}\\
  \mathrm{connect}(2,3) =& \{\mathrm{Face}: \mathrm{RIGHT\ Face},\\
                         \mathrm{Face2}&\mathrm{Con}: \mathrm{LEFT\ Face},
                           \mathrm{Orientation}: \mathtt{Null}\}\\
  \mathrm{connect}(3,2) =& \{\mathrm{Face}:
                           \mathrm{LEFT\ Face},\\
                           \mathrm{Face2}&\mathrm{Con}: \mathrm{RIGHT\ Face},
                                        \mathrm{Orientation}: \mathtt{Null}\}
\end{align*}
In addition, for this simple topology, the root module is Module 2.

\subsection{Self-assembly Problem}
\label{sec:task-assign}

Assume there is a team of modules
$\mathcal{M} = \{m_1, m_2, \cdots, m_n\}$ in the Euclidean space
$\mathbb{R}^2$. The state of a module $m_i\in \mathcal{M}$ is defined as
$p_i = \left[x_i, y_i, \theta_i\right]^\intercal$ where
$o_i = \left[x_i, y_i\right]^\intercal$ is the location of the center of
$m_i$ and $\theta_i$ is the orientation of $m_i$. Then the distance between
module $m_i$ and $m_j$ can be derived as $\|o_i - o_j\|$. Every
SMORES-EP module is a cube with a side length of $w$. The assembly
goal is a SMORES-EP tree topology configuration $G=(V,E)$ where
$\vert V\vert = n$. Not all kinematic topology can be built by the self-assembly
process. Only the kinematic topology that can be unfolded onto a plane
can be achieved by a bunch of separated modules on the ground.

\begin{definition}
  \label{def:topology}
  The target kinematic topology that can be self-assembled by
  separated modules is a modular robot configuration $G=(V,E)$ that
  can be fully unfolded to a plane satisfying:
  \begin{enumerate}
  \item $G$ is a connected graph;
  \item The Euclidean distance between two adjacent modules is $w$;
  \item The center of every module occupies a unique location.
  \end{enumerate}
\end{definition}

Our target kinematic topologies are different from static assembled
structures. For modular robots that are aiming for self-assembling
static structures, the connections among the modules are homogeneous
and the positions of the modules are enough to fully define the
target. However, in our case, a module is composed of several joints
and several modules can be connected in many ways to generate
different kinematic chains. Hence, for our targets, we not only need
to move the modules to their desired positions but also adjust their
orientations to construct all connections defined in $E$ of $G$. We
are not interested in all topologies that satisfy the constraints in
Definition~\ref{def:topology}. Some topologies are less useful for
executing tasks. For example, the connection in Fig.~\ref{fig:bb-1}
shows the two BOTTOM Faces are connected with \textit{Orientation}
1. This arrangement unnecessarily constrains the relative orientation
between two modules so it is not considered in our work. In addition,
the target topology is also meant to satisfy hardware constraints,
such as the connector and the actuator limitations when lifting many
modules.

The modules are at arbitrary locations with the constraint that the
distance between any pair of modules $m_i$ and $m_j$ denoted as
$d_{ij}$ is greater than $w$. The kinematic topology self-assembly
problem is stated. Given a target kinematic topology $G = (V,E)$ and a
team of $n$ modules $\mathcal{M} = \{m_1, m_2, \cdots, m_n\}$ where
$n = \vert V\vert$, find a sequence of collision-free assembly actions to form
the target kinematic topology.

\section{Parallel Self-assembly Algorithm}
\label{sec:assembly}

We propose a parallel self-assembly algorithm for a set of modules to
form a desired kinematic topology.

\subsection{Task Assignment}
\label{sec:task}

In order to self-assemble $n$ separated modules into a target
kinematic topology, we need to map the corresponding role of each
module in the target. We model this problem as a task assignment
problem, finding the optimal assignment solution among $n!$ different
assignments with respect to some cost function.

\begin{figure}[t]
  \centering
  \subfloat[]{\includegraphics[height=0.13\textwidth]{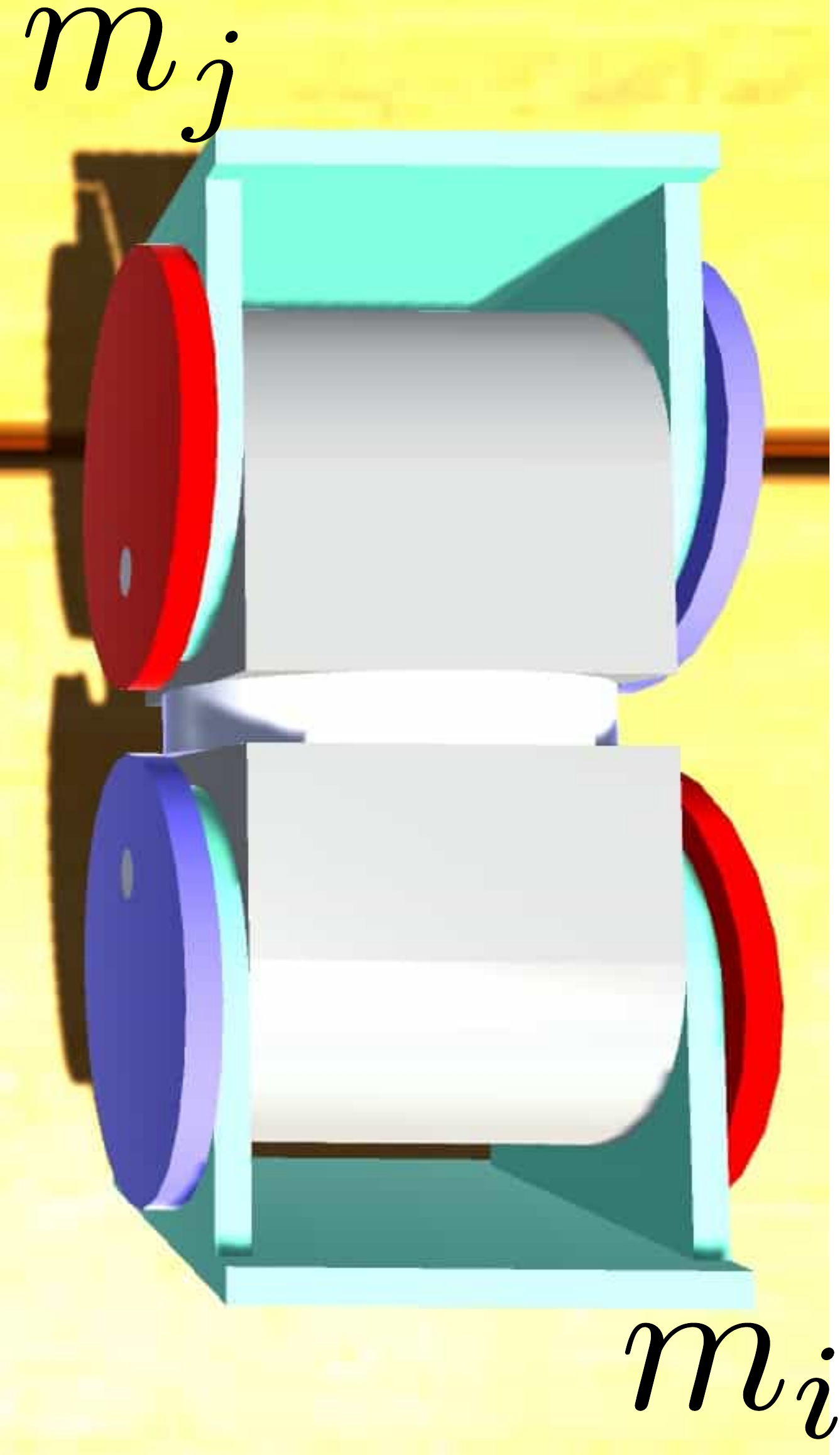}\label{fig:2-module}}
  \hfil
  \subfloat[]{\includegraphics[height=0.13\textwidth]{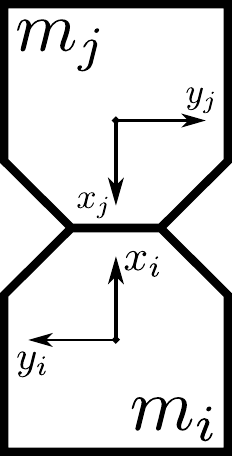}\label{fig:2-module-TT}}
  \hfil
  \subfloat[]{\includegraphics[height=0.13\textwidth]{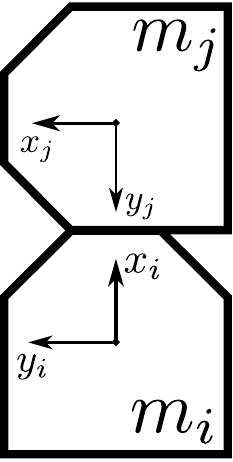}\label{fig:2-module-TL}}
  \hfil
  \subfloat[]{\includegraphics[height=0.13\textwidth]{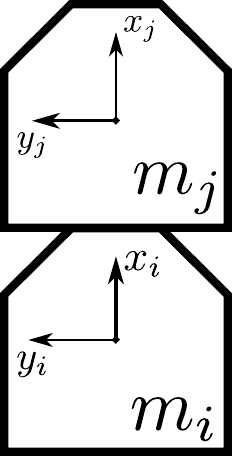}\label{fig:2-module-TB}}
  \hfil
  \subfloat[]{\includegraphics[height=0.13\textwidth]{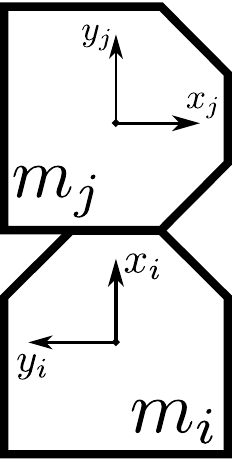}\label{fig:2-module-TR}}
  \caption{(a) $m_i$ TOP Face is connected with $m_j$ TOP Face and (b)
  its kinematic diagram. (c) (d) (e) are the kinematic diagrams of
  three other cases when $m_i$ TOP Face is involved in the connection.}
\end{figure}

Given a target kinematic topology $G = (V, E)$, first check if it
satisfies the requirements in Definition~\ref{def:topology} in order
to be self-assembled and, if so, fully unfold it on the ground. Given
the initial placement of all modules (in our case, all modules are
placed on the ground with their electronics board facing upwards), the
solution to unfold a kinematic topology is unique if the displacement
of all modules remain unchanged. The root module $\tau$ of $G$ can be
computed by Algorithm~\ref{alg:root} in linear time. Then the state of
each module $v_i\in V$ with respect to this root module $\tau$ denoted
as $\bar{p}_i$ after fully unfolding $G$ can be computed in
breadth-first search order starting from $\tau$. The state of a module
is fully determined if the state of its parent module and the involved
connection are known. For example, module $m_i$ is the parent of
module $m_j$, with TOP Face of $m_j$ attached to TOP Face of $m_i$
shown in Fig.~\ref{fig:2-module} (kinematic diagram in
Fig.~\ref{fig:2-module-TT}). There are three more cases with TOP Face
of $m_i$ being involved shown in Fig.~\ref{fig:2-module-TL} ---
Fig.~\ref{fig:2-module-TR}. In breadth-first search order, when
visiting $m_j$, the state of $m_i$ with respect to $\tau$ should
already be known that is
$\bar{p}_i = \left[\bar{x}_i, \bar{y}_i,
  \bar{\theta}_i\right]^{\intercal}$.  The state of $m_j$ with respect
to $m_i$ denoted as $\bar{p}_{ji}$ is determined by the involved
connectors, e.g., $\bar{p}_{ji} = \left[w, 0, \pi\right]^\intercal$
for Fig.~\ref{fig:2-module-TT}. Then the state of $m_j$ with respect
to $\tau$ is
\begin{equation}
  \label{eq:module-state}
  \bar{p}_j = \left[
    \begin{array}{cc}
      R&0\\
      0&1
    \end{array}
  \right]\bar{p}_{ji} + \bar{p}_{i}
\end{equation}
in which $R = \left[
  \begin{array}{cc}
    \cos \bar{\theta}_i&-\sin \bar{\theta}_i\\
    \sin \bar{\theta}_i&\cos \bar{\theta}_i
  \end{array}
\right]$.

Given a set of $n$ modules $\mathcal{M}$, each $m\in \mathcal{M}$ needs
to be mapped by a module $v\in V$ in an optimal way. In our task
assignment problem, we want to minimize the total distance that all
modules have to travel in order to assemble $G$. First, the center
location of all modules can be defined as
$o_c = \left[x_c, y_c\right]^\intercal$ where
$x_c = \sum_{i=1}^n x_i / n$ and $y_c = \sum_{i=1}^n y_i / n$. Then the root
module is selected as
\begin{equation}
  \label{eq:root}
  m_\tau = \underset{m_i\in\mathcal{M}}{\arg\min} \|o_i - o_c\|
\end{equation}

The state of every module $m_i\in \mathcal{M}$ with respect to
$m_\tau$ denoted as $\tilde{p}_i$ can be computed simply with a rigid
body transformation. Obviously
$\tilde{p}_\tau = \left[0, 0, 0\right]^\intercal$ that is the state of
$m_{\tau}$ with respect to itself. Recall that $\bar{o}_i$ is the
location of the center of $v_i$ with respect to $\tau\in V$. Given
$m_\tau$ is mapped to $\tau$, namely
$\|\tilde{o}_\tau - \bar{o}_\tau\| = 0$, the distance between every pair of
$m_i\in \mathcal{M}\setminus \{m_\tau\}$ and
$v_j\in V\setminus \{\tau\}$ is simply $\|\tilde{o}_i - \bar{o}_j\|$ which is the
cost of the task --- moving to the location of $v_j$ --- for module
$m_i$. Other factors can also be included in the cost besides
distance, such as the orientation. The optimal task assignment problem
can be solved by Kuhn-Munkres algorithm or other concurrent assignment
and planning of trajectories algorithms in polynomial
time~\citep{Turpin-capt-icra-2013}. The output is a one-to-one and onto
mapping $f: V\rightarrow \mathcal{M}$ which is used later by a motion planner to
generate the assembly sequence.

\subsection{Parallel Assembly Actions}
\label{sec:assembly-actions}

With mapping $f: V\rightarrow\mathcal{M}$, we can compute the assembly sequence
from the root to the leaves of $G$. In each step, the modules in
$\mathcal{M}$ mapped to the modules in the target kinematic topology
$G$ at the same depth can be executed in a parallel manner. Let $d(G)$
and $d(v)$ be the depth of the rooted graph $G$ and vertex $v\in V$
respectively, then for any vertex with $d(v)> 0$, we denote its parent
connected via its connector $c$ as $\tilde{v}^c$ and the mating
connector of $\tilde{v}^c$ as $\tilde{c}'$. An assembly action is a
tuple of the form $(m_i, c_i, m_j, c_j)$ which means connect $m_i$'s
connector $c_i$ with $m_j$'s connector $c_j$. The parallel assembly
algorithm is shown in Algorithm~\ref{alg:parallel-assembly}. In every
iteration, we first fetch all modules in the current depth from the
target topology, then, with the optimal mapping
$f:V\rightarrow \mathcal{M}$, the corresponding moving modules ($m$) are
determined as well as their parent modules ($\tilde{m}^c$). We then
have a group of assembly actions $A$ composed of
$(m,c, \tilde{m}^c, \tilde{c}^\prime)$.
\begin{algorithm}[t]
  \caption{Parallel Assembly}\label{alg:parallel-assembly}
  \KwIn{Target Kinematic Topology $G=(V, E)$ with root $\tau$ and depth
    $d(G)$, $f:V\rightarrow\mathcal{M}$}
  \KwOut{Parallel Assembly Sequence $\mathcal{A}$}
  $d\leftarrow 1$\;
  \While{$d\le d(G)$}{
    Create empty action queue $A$\;
    $\overline{V}\leftarrow \{v\in V\vert d(v) = d\}$\;
    \For{$v\in\overline{V}$}{
      $m \leftarrow f(v)$\;
      $\tilde{m}^c\leftarrow f(\tilde{v}^c)$\;
      $A.\mathrm{enqueue}((m, c, \tilde{m}^c, \tilde{c}'))$\;
    }
    $\mathcal{A}.\mathrm{enqueue}(A)$\;
    $d\leftarrow d+1$\;
  }
\end{algorithm}

For each group of assembly actions $A\in \mathcal{A}$, all actions can
be executed in parallel except when multiple modules are docking with
different connectors of the root module. The group of assembly actions
$A$ with the root module involved is separated into two subgroups: one
group contains the actions for LEFT Face and RIGHT Face of the root
module which can be executed first. The other group contains the
actions for TOP Face and BOTTOM Face of the root module which are
executed later. This is because the root module is not fixed to the
ground and it is hard to ensure all modules to be attached can
approach the root module simultaneously. When docking with a module
that is not a root module, since this module is already attached to a
group of modules, it is less likely to be pushed by other modules when
executing the docking process. For each assembly action
$a = (m_i, c_i, m_j, c_j)\in \mathcal{A}$, a motion controller first
navigates $m_i$ to a location close to $m_j$ where the distance is
determined by the grid size of the environment. Then module $m_i$
adjusts its pose to align the connector $c_i$ with $c_j$, and finally
it approaches $c_j$ to finish the docking process. A square grid
environment is generated once the root module is determined, with the
root module at the center. The grid serves as a routing graph for
navigating modules.

\subsection{Docking Control}
\label{sec:docking}

As mentioned, docking is a difficult part of self-assembly. We divide
the docking process into three steps to ensure its success:
\textit{navigation}, \textit{pose adjustment}, and \textit{approach}.

\subsubsection{Navigation}
\label{sec:navigate}

When doing self-assembly, each SMORES-EP module $m\in\mathcal{M}$ can
behave as a differential-drive vehicle with the following kinematics
model:
\begin{equation}
  \label{eq:diff-model}
  \dot{\tilde{p}} = \left[
    \begin{array}{c}
      \dot{\tilde{x}}\\\dot{\tilde{y}}\\\dot{\tilde{\theta}}
    \end{array}
  \right] = \left[
    \begin{array}{cc}
      \cos\tilde{\theta}&0\\
      \sin\tilde{\theta}&0\\
      0&1
    \end{array}
  \right]\left[
    \begin{array}{c}
      v\\\omega
    \end{array}
\right]
\end{equation}
in which $v$ is the linear velocity along $x$-axis of the body frame
of $m_i$ and $\omega$ is the angular velocity around $z$-axis of body
frame $m_i$ which are determined by the velocity of LEFT DOF and RIGHT
DOF.  Given a set of assembly actions $A\in \mathcal{A}$ in which all
modules to be attached are at the same depth, the collision-free paths
to navigate all involved modules can be generated by a multi-vehicle
planner, e.g.~\cite{Binder-multi-vehicle-planning-iros-2019}. Then,
$\forall a = (m_i, c_i, m_j, c_j)\in A$, the module $m_i$ is
controlled to follow a path to a location close to $m_j$ along the
routes predefined by the grid environment. Every module is controlled
by a real-time path-following controller and synchronized with other
moving modules. Hence, prioritized planning approaches can be applied
to handle potential collision among moving modules. For
example,~\cite{Binder-multi-vehicle-planning-iros-2019} developed
three simple strategies (\textit{Wait}, \textit{Avoid}, \textit{Push})
combined with rescheduling method to handle possible local collision.

\subsubsection{Pose Adjustment}
\label{sec:pose-adjust}

\begin{figure}[t]
  \centering
  \subfloat[\label{fig:pose-x}]{\includegraphics[width=0.3\textwidth]{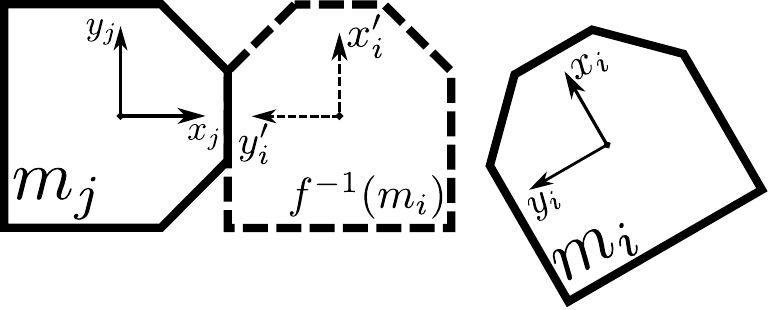}}
  \hfil
  \subfloat[\label{fig:pose-y}]{\includegraphics[width=0.3\textwidth]{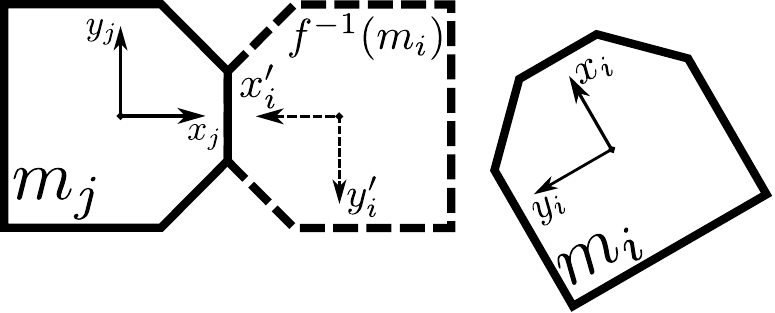}}
  \caption{(a) The assembly action is to connect LEFT Face of $m_i$
    with $m_j$ and $f^{-1}(m_i)$ (the goal pose of $m_i$) is shown in
    dashed line. In this case, we have to align the connector by
    adjusting $x^{\prime}_i$ and $\theta^{\prime}_i$ to zeros. (b) If the assembly
    action is to connect TOP Face of $m_i$ with $m_j$, then we have to
    align the connector by adjusting $y^{\prime}_i$ and
    $\theta^{\prime}_i$ to zeros.}
  \label{fig:pose-adjust}
\end{figure}

Once the navigation procedure is done for all modules involved in
$A\in\mathcal{A}$, $m_i$ starts to adjust its pose to align the involved
connector. If $c_i$ (the connector of $m_i$ to be attached with $m_j$)
is either LEFT Face or RIGHT Face, then adjust $x_i^\prime$ and
$\theta_i^\prime$ to zeros where $x_i^\prime$ is the $x$ location of
$m_i$ with respect to the body frame of $f^{-1}(m_i)$ that is the goal
pose of $m_i$ and $\theta_i^\prime$ is the orientation of $m_i$ with respect to
the body frame of $f^{-1}(m_i)$ (recall that $f$ is the mapping from
$V\rightarrow\mathcal{M}$ derived by solving the task assignment
problem). Otherwise, adjust $y_i^\prime$ and $\theta_i^\prime$ to zeros where
$y_i^\prime$ is the $y$ location of $m_i$ with respect to the body frame of
$f^{-1}(m_i)$. Two cases are shown in Fig.~\ref{fig:pose-adjust}. Note
that this process has nothing to do with $c_j$.  A kinematics model
for the second case can be derived as
\begin{equation}
  \label{eq:pose-kinematics}
  \left[
    \begin{array}{c}
      \dot{y}_i^\prime\\\dot{\theta}_i^\prime
    \end{array}
  \right] = \left[
    \begin{array}{cc}
      \sin\theta_i^\prime&0\\
      0&1
    \end{array}
  \right]\left[
    \begin{array}{c}
      v\\\omega
    \end{array}
  \right]
\end{equation}
A control law to make $y_i^\prime$ and $\theta_i^\prime$ to converge to zeros is
\begin{equation}
  \label{eq:control-law}
  \left[
    \begin{array}{c}
      v\\\omega
    \end{array}
  \right] = \left[
    \begin{array}{cc}
      \sin\theta_i^\prime&0\\
      0&1
    \end{array}
  \right]^{-1} K_{2\times 2} \left[
    \begin{array}{c}
      -y_i^\prime\\-\theta_i^\prime
    \end{array}
  \right]
\end{equation}
where $K$ is positive definite and $K = \mathrm{diag}(2,1)$ in our
experiments. A similar controller can be derived for the first
case. For the pose adjustment controller, we relax the constraint on
one dimension making the system fully actuated. For example, with the
controller in Eq.~\eqref{eq:control-law}, only $y_i^{\prime}$ and
$\theta_i^{\prime}$ are under control and $x_i^\prime$ is no longer constrained, so
$x_i^\prime$ may diverge possibly resulting in collision. However, this
drift is not a concern for our modules which use differential drive
with positioned controlled wheel motion. This is confirmed in our
hardware experiments.

\subsubsection{Approach}
\label{sec:approach}

\begin{figure}[t]
  \centering
  \includegraphics[width=0.4\textwidth]{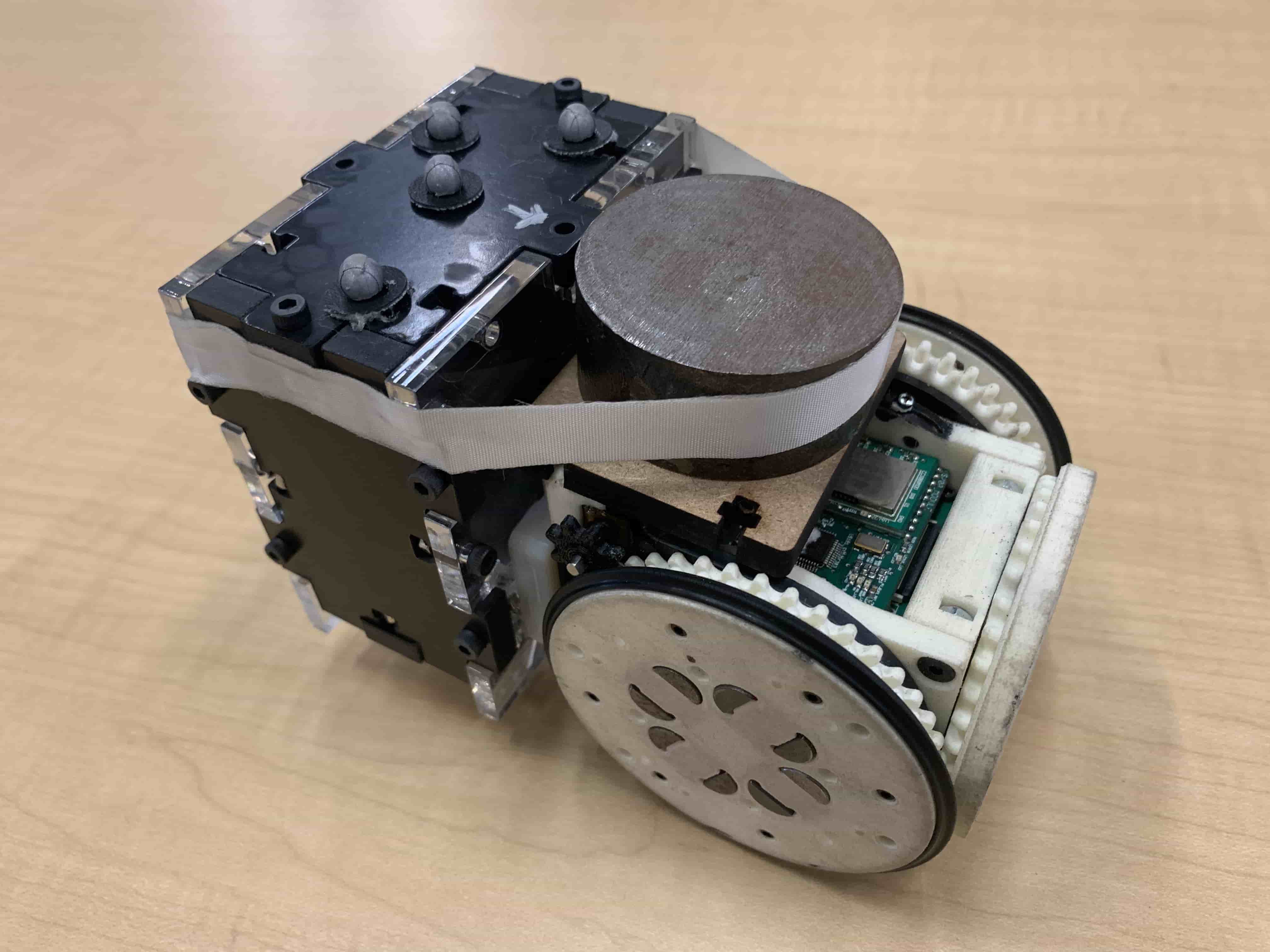}
  \caption{A helping module is a SMORES-EP module equipped with some
    payload so that it can lift another module.}
  \label{fig:helping-module}
\end{figure}

\begin{figure*}[t]
    \centering
    \subfloat[\label{fig:task1-step1}]{\includegraphics[height=0.30\textwidth]{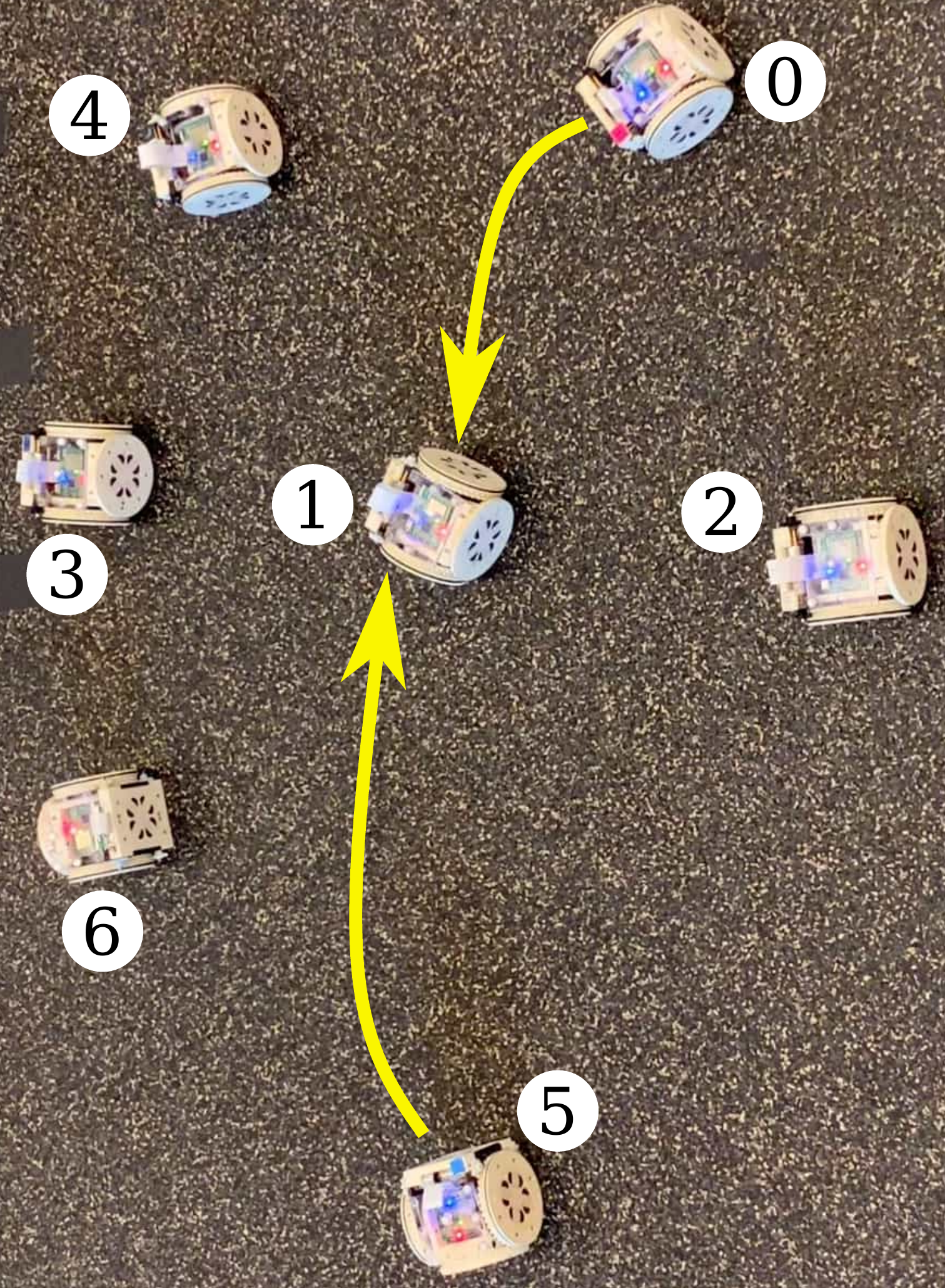}}
    \hfil
    \subfloat[\label{fig:task1-step2}]{\includegraphics[height=0.30\textwidth]{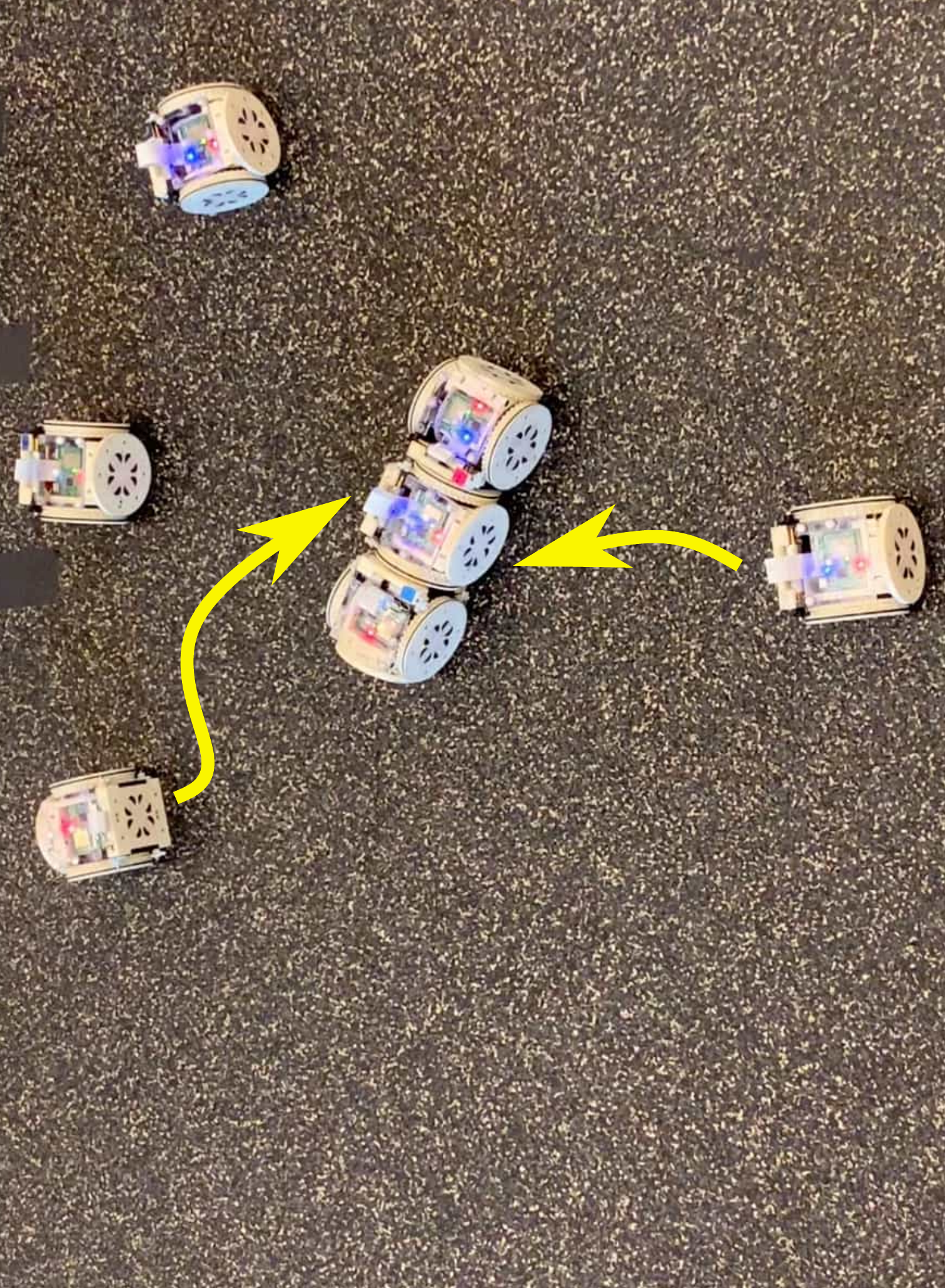}}
    \hfil
    \subfloat[\label{fig:task1-step3}]{\includegraphics[height=0.30\textwidth]{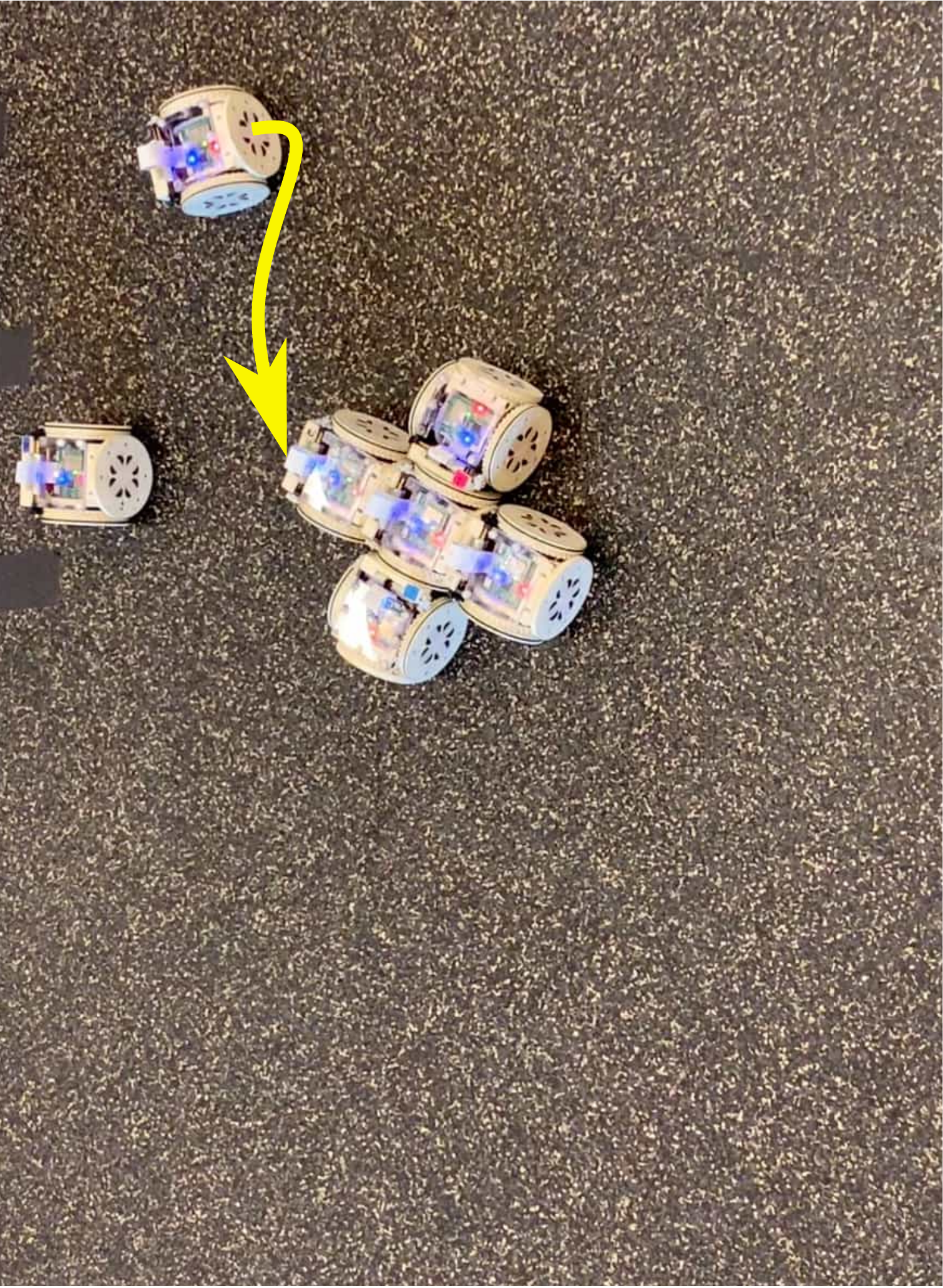}}
    \hfil
    \subfloat[\label{fig:task1-step4}]{\includegraphics[height=0.30\textwidth]{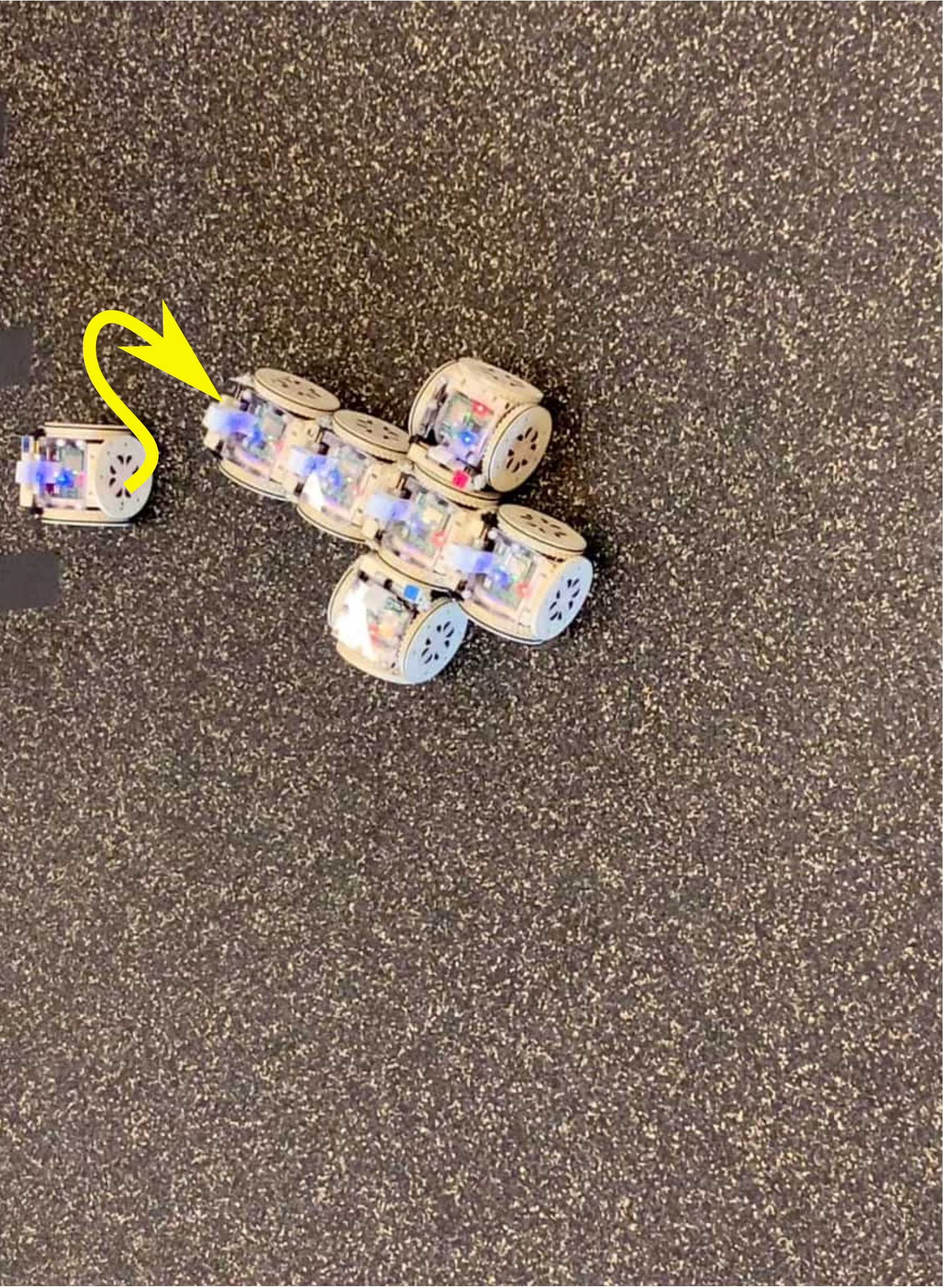}}\\
    \subfloat[\label{fig:task1-assembly}]{\includegraphics[height=0.30\textwidth]{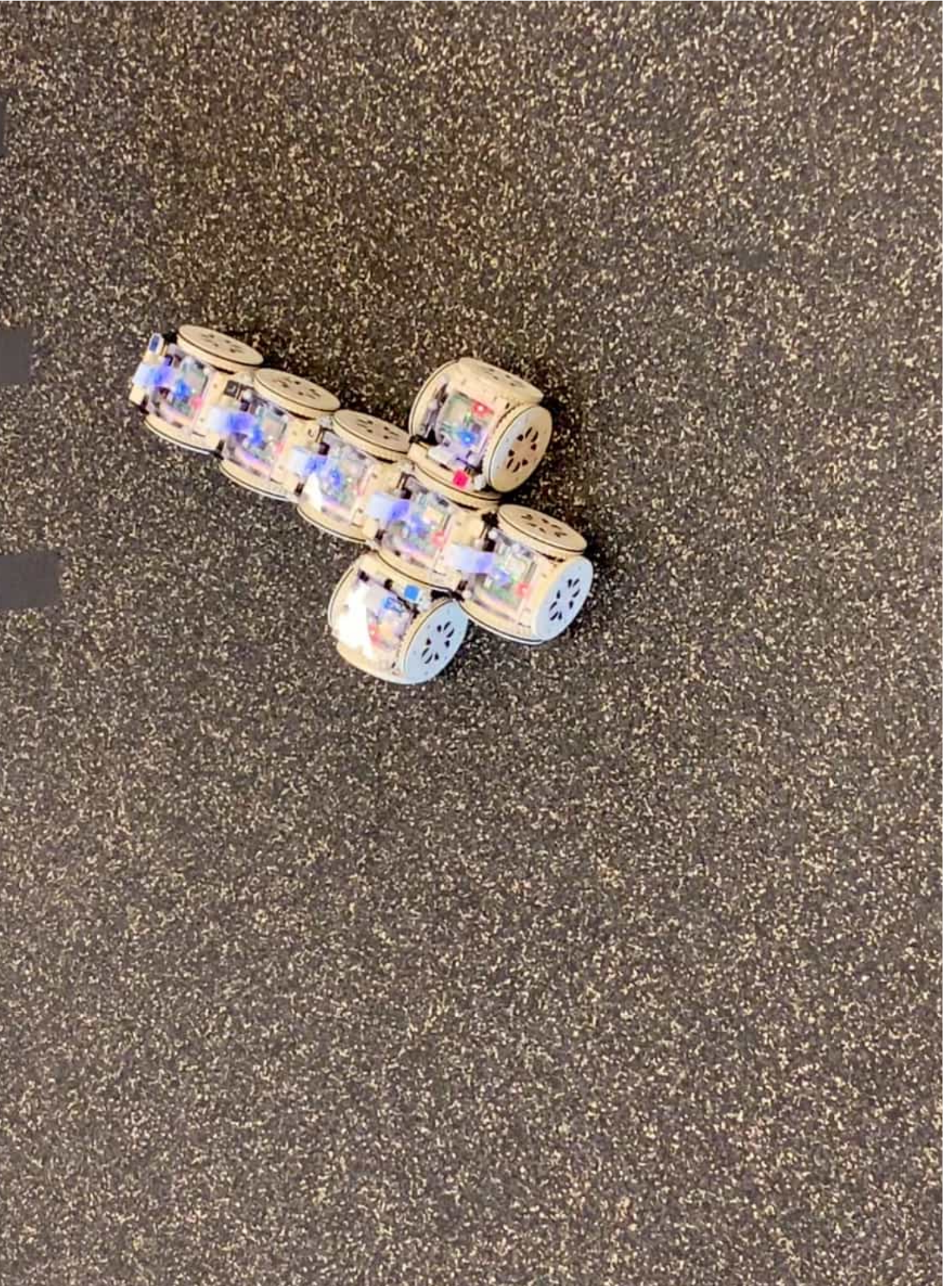}}
    \quad
    \subfloat[\label{fig:task1-goal}]{\includegraphics[height=0.30\textwidth]{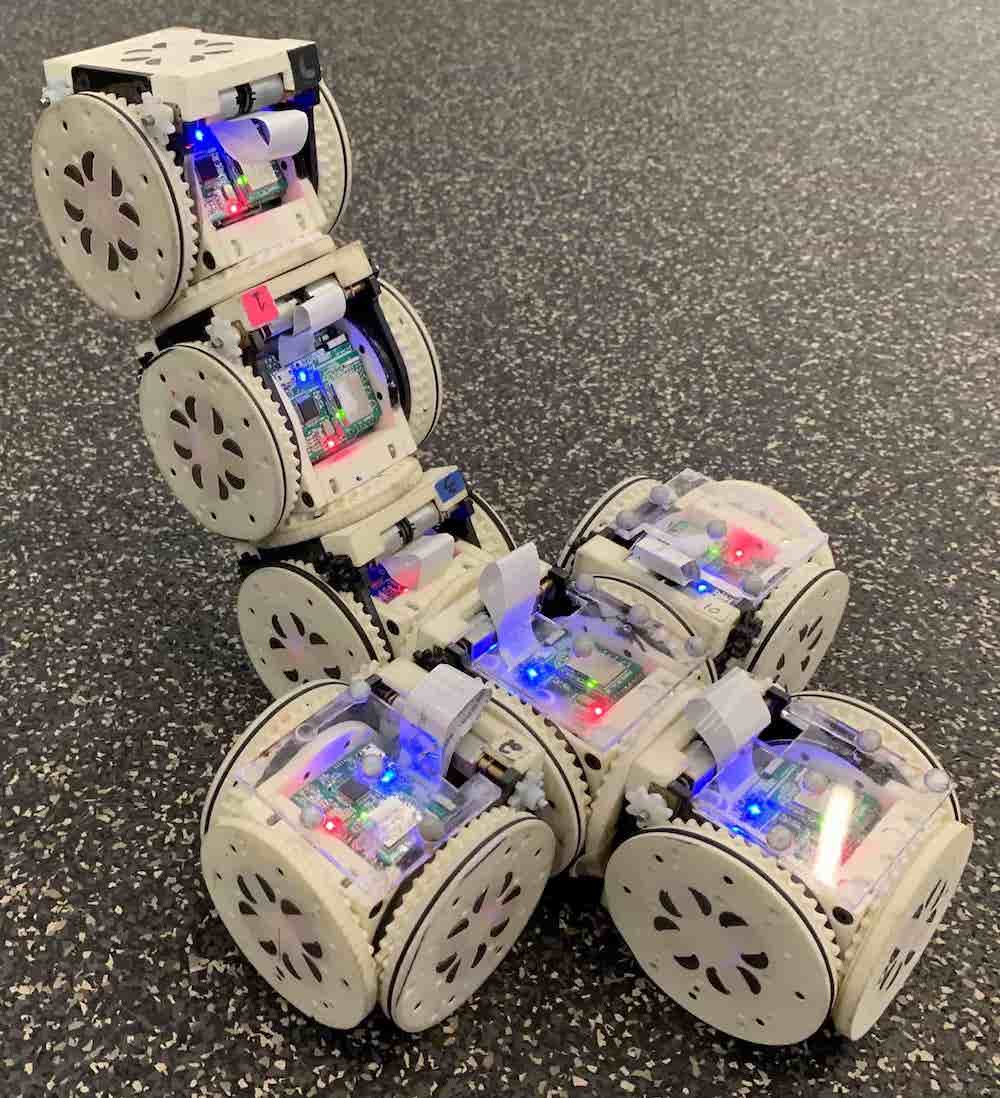}}
    \quad
    \subfloat[\label{fig:task1-unity}]{\includegraphics[height=0.30\textwidth]{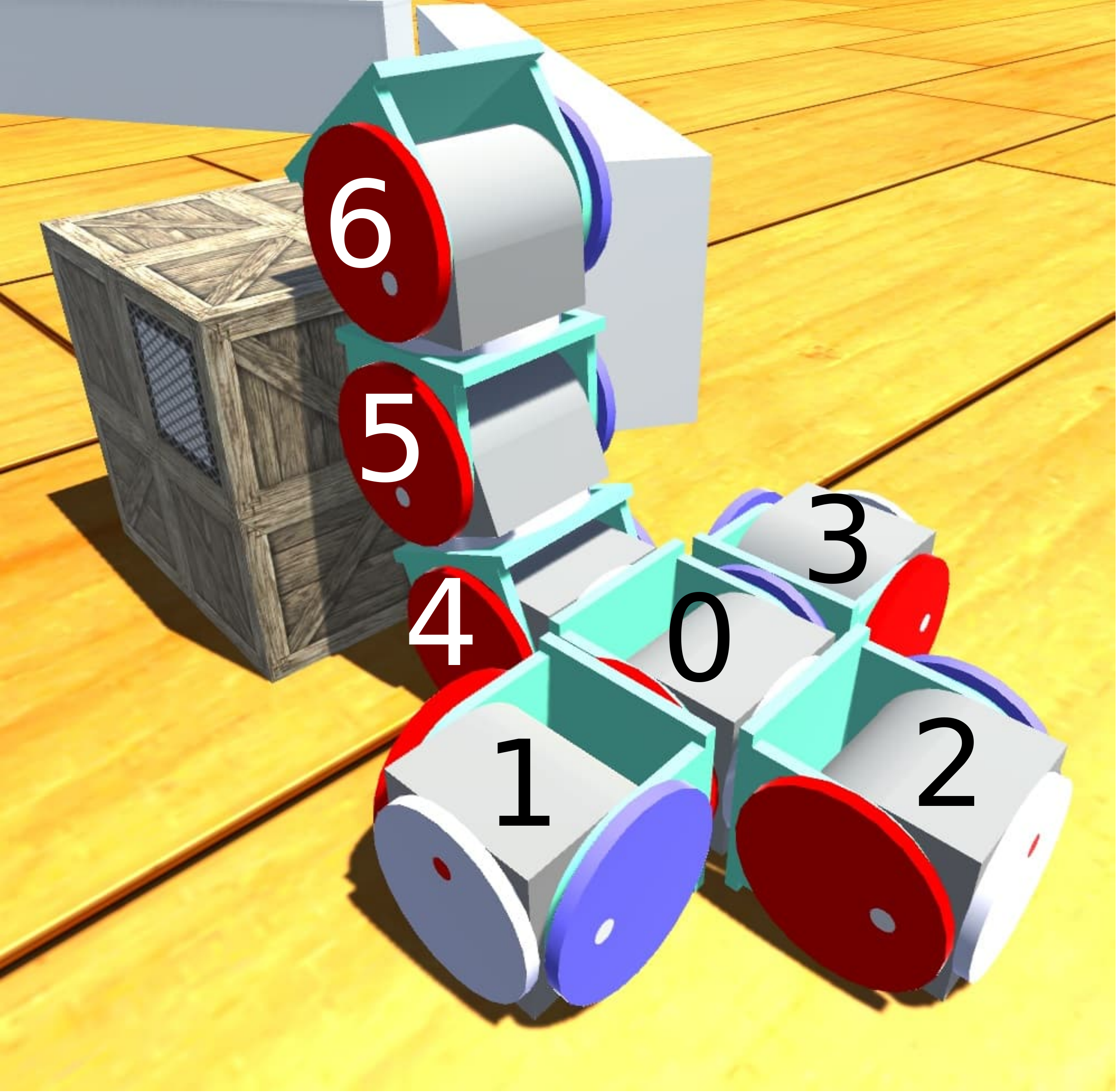}}
    \caption{SMORES-EP hardware mobile manipulator self-assembly: (a) Execute actions $(0, \text{\textbf{B}}, 1, \text{\textbf{L}})$ and $(5, \text{\textbf{B}}, 1, \text{\textbf{R}})$; (b) Execute actions $(2, \text{\textbf{B}}, 1, \text{\textbf{T}})$ and $(6, \text{\textbf{T}}, 1, \text{\textbf{B}})$; (c) Execute action $(4, \text{\textbf{T}}, 6, \text{\textbf{B}})$; (d) Execute action $(3, \text{\textbf{T}}, 4, \text{\textbf{B}})$. (e) --- (f) The final assembly. (g) The target kinematic topology.}
    \label{fig:task1-process}
\end{figure*}

The last step is to approach $c_j$ by moving in a straight line which
is similar to the controller used in navigation step following a given
trajectory. If $c_i$ is either TOP Face or BOTTOM Face, then module
$m_i$ will first adjust $c_i$ to the right position and then keep
moving pushing $m_j$ until $c_i$ and $c_j$ are fully
connected. Otherwise, when $c_i$ is either LEFT Face or RIGHT Face, a
helping module shown in Fig.~\ref{fig:helping-module} is needed. A new
assembly action $(m_{\mathrm{H}}, \text{\textbf{T}}, m_i, \bar{c}_i)$ where $m_{\mathrm{H}}$
is a helping module and $\bar{c}_i$ is LEFT Face if $c_i$ is RIGHT
Face, or the vice versa. After $m_{\mathrm{H}}$ is connected with $m_i$, $m_i$ is
lifted so that it can adjust $c_i$ to the right position, delivered to
its destination, and then placed down. Finally, $m_{\mathrm{H}}$ keeps moving to
push $m_i$ to approach $m_j$ for docking. The helping module is a
SMORES-EP module equipped with an extra mass on the BOTTOM Face as a
counter balance while lifting a module. In our setup, there is only
one helping module, and, for each assembly action group that is a
queue, the helping module offers help sequentially. It is possible to
have more to execute helping actions in parallel.  While moving
modules translate, the controller should keep the orientation of these
modules fixed as docking requires the magnet faces to mate
properly. To ease this requirement on the SMORES-EP modules the
EP-Faces have a relatively large area of acceptance.

\begin{figure}[t!]
    \centering
    \includegraphics[width=0.4\textwidth]{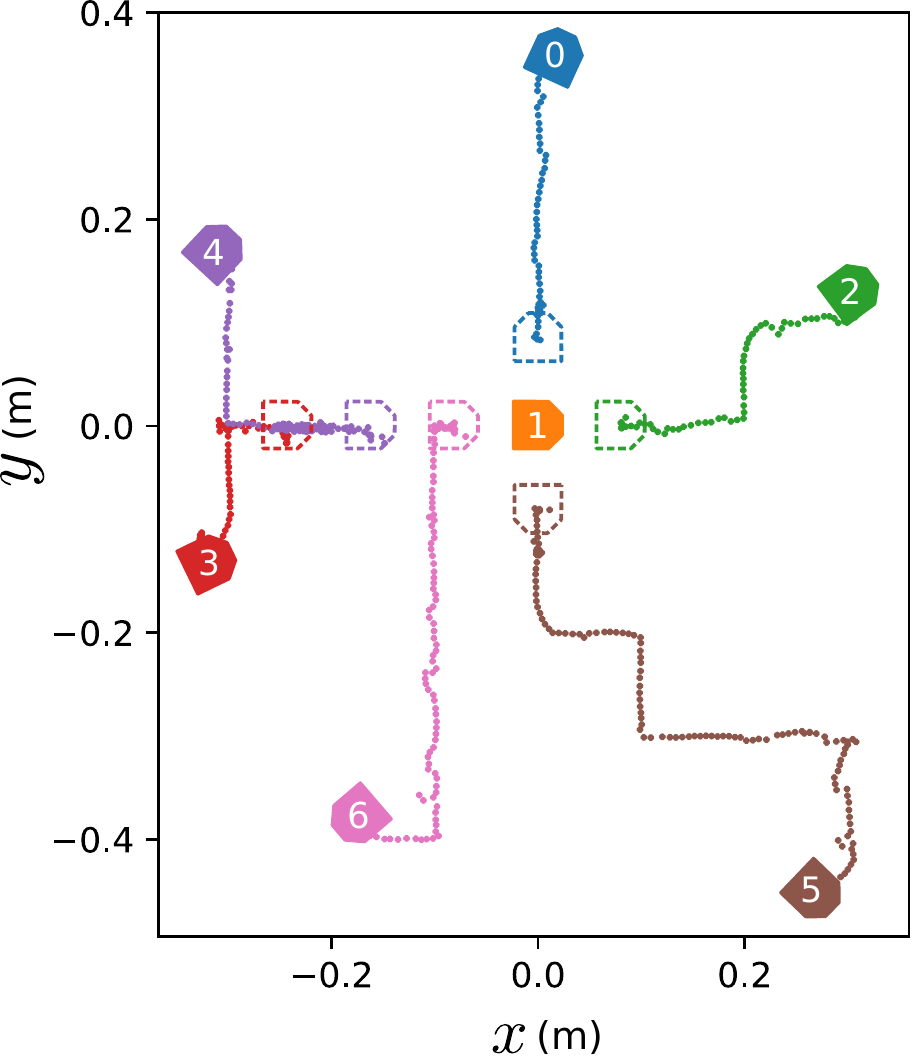}
    \caption{Actual path of each module for Task 1.}
    \label{fig:task1-path}
\end{figure}

\section{Experiments}
\label{sec:exp}

Our framework and algorithm are demonstrated with SMORES-EP modules on three tasks
to evaluate the hardware and software integration. All paths for
modules are generated on an empty grid map where each grid cell is a
square \SI{10}{cm}$\times$\SI{10}{cm}. We also show this framework can be
used to speed up the self-reconfiguration process proposed
by~\cite{Liu-smores-reconfig-ral-2019}.

\begin{table}[t]
    \centering
    \caption{Initial Locations of All Modules in Task 1}
    \label{tab:task1-init-pose}
    \begin{tabular}{c c c c}
    \toprule
    Module &   $x$ (m)  &   $y$ (m)  & $\theta$ (rad) \\
    \midrule
    Module 0  &  0.017 &  0.357 &  1.142  \\
    Module 1  &  0.0   &  0.0   &  0.0    \\
    Module 2  &  0.305 &  0.129 &  0.641  \\
    Module 3  & -0.318 & -0.132 &  0.454  \\
    Module 4  & -0.318 &  0.158 &  0.823  \\
    Module 5  &  0.264 & -0.448 & -0.763  \\
    Module 6  & -0.172 & -0.380 & -2.431  \\
    \bottomrule
    \end{tabular}
\end{table}

\subsection{Task 1: Mobile Manipulator}
The first task is to form a mobile vehicle with an arm that can reach
higher locations as shown in Fig.~\ref{fig:task1-unity}. There are
seven modules involved with Module 1 selected as the root module
$m_\tau$ which is closest to the center (Fig.~\ref{fig:task1-step1}). The
initial locations of all modules with respect to the root module is
shown in Table \ref{tab:task1-init-pose}. In the target topology
(Fig.~\ref{fig:task1-unity}), the root module $\tau$ is computed as
Module 0.  Hence Module 0 in the target topology $G=(V,E)$ is mapped
to Module 1 in the set of modules $\mathcal{M}$ on the ground. The
mapping $f: V\rightarrow \mathcal{M}$ that is $0\to 1$, $1\to 5$,
$2\to 2$, $3\to 0$, $4\to 6$, $5\to 4$, and $6\to 3$ is then derived by
Kuhn-Munkres algorithm.

\begin{figure}[t]
  \centering
    \subfloat[]{\includegraphics[width=0.22\textwidth]{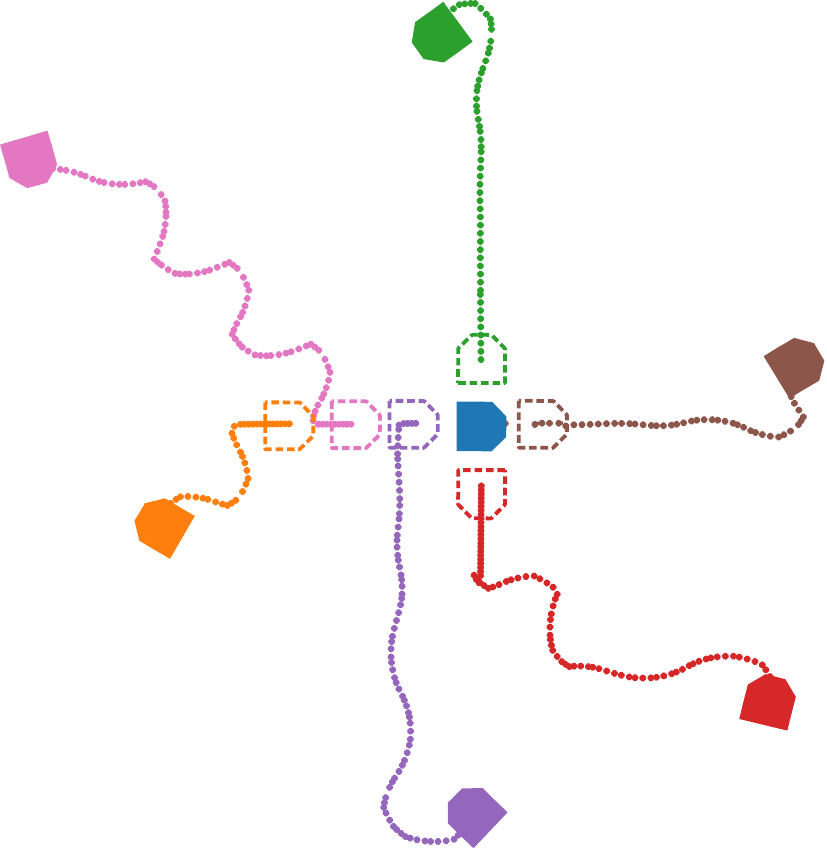}}
    \hfill
    \subfloat[]{\includegraphics[width=0.24\textwidth]{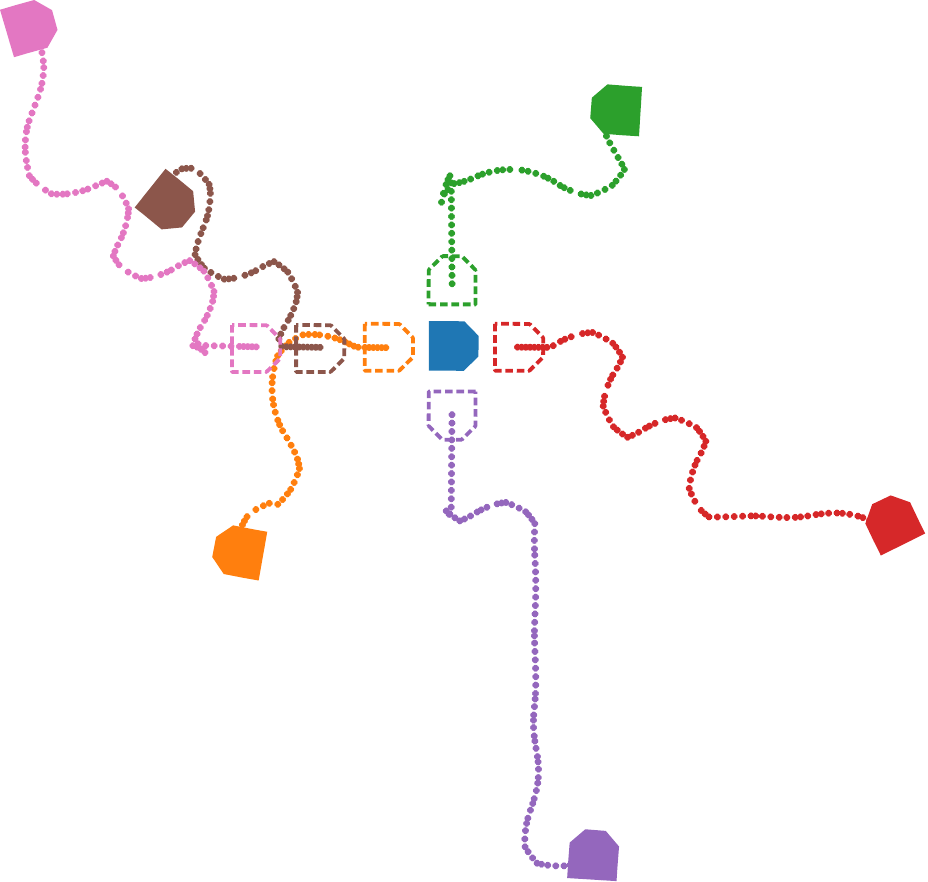}}
    \caption{Two additional experiments for Task 1.}
    \label{fig:task1-more}
\end{figure}

The assembly sequence starts from all vertices $v\in V$ with depth of
one in the target topology which include Module 0, Module 2, Module 5,
and Module 6. Because the root module is involved in this step, the
assembly actions are separated into two subgroups. Module 0 and Module
5 start moving first to dock with LEFT Face and RIGHT Face of Module 1
respectively (Fig.~\ref{fig:task1-step1}), then Module 2 and Module 6
begin the docking process with TOP Face and BOTTOM Face of the root
module (Fig.~\ref{fig:task1-step2}). In this way, even Module 1 can be
moved slightly after docking with Module 0 and Module 5, Module 2 and
Module 6 can still dock with Module 1 successfully. Lastly, Module 4
and Module 3 execute the assembly actions (Fig.~\ref{fig:task1-step3}
and Fig.~\ref{fig:task1-step4}).  It takes 130 seconds to finish the
whole assembly process with paths shown in
Fig. \ref{fig:task1-path}. Two additional experiments in simulation
with different initial settings are shown in
Fig.~\ref{fig:task1-more}.

\begin{figure}[t]
    \centering
    \subfloat[]{\includegraphics[width=0.15\textwidth]{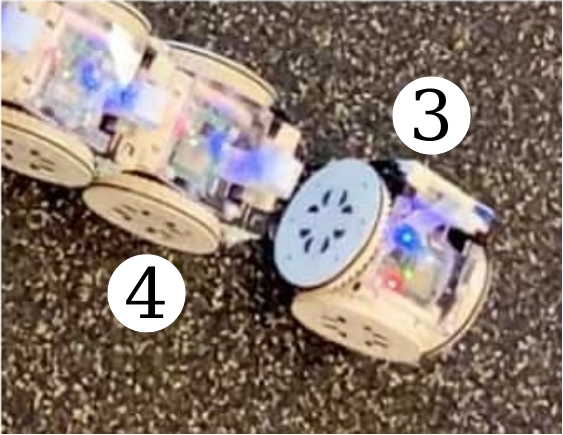}\label{fig:adj-1}}
    \hfill
    \subfloat[]{\includegraphics[width=0.15\textwidth]{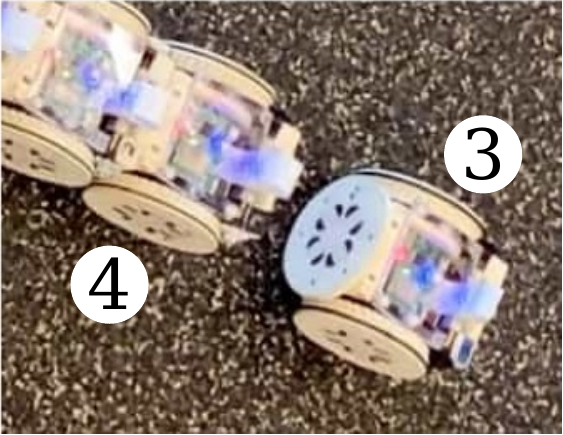}\label{fig:adj-2}}
    \hfill
    \subfloat[]{\includegraphics[width=0.15\textwidth]{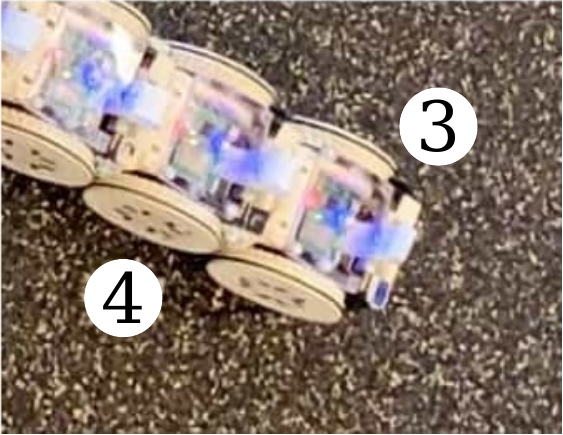}\label{fig:adj-3}}
    \caption{Adjustment of the position and orientation before
      executing assembly action
      $(3, \text{\textbf{T}}, 4, \text{\textbf{B}})$. (a) Module 3
      finished navigation process and started to adjust its pose. (b)
      $y_3^\prime$ and $\theta_3^\prime$ have been adjusted and it
      started to approach the goal for docking. (c) The docking
      process of Module 3 was accomplished.}
    \label{fig:controller_adj}
\end{figure}

\begin{figure}[t!]
    \centering
    \subfloat[\label{fig:y_yaw_ctrl}]{\includegraphics[width=0.24\textwidth]{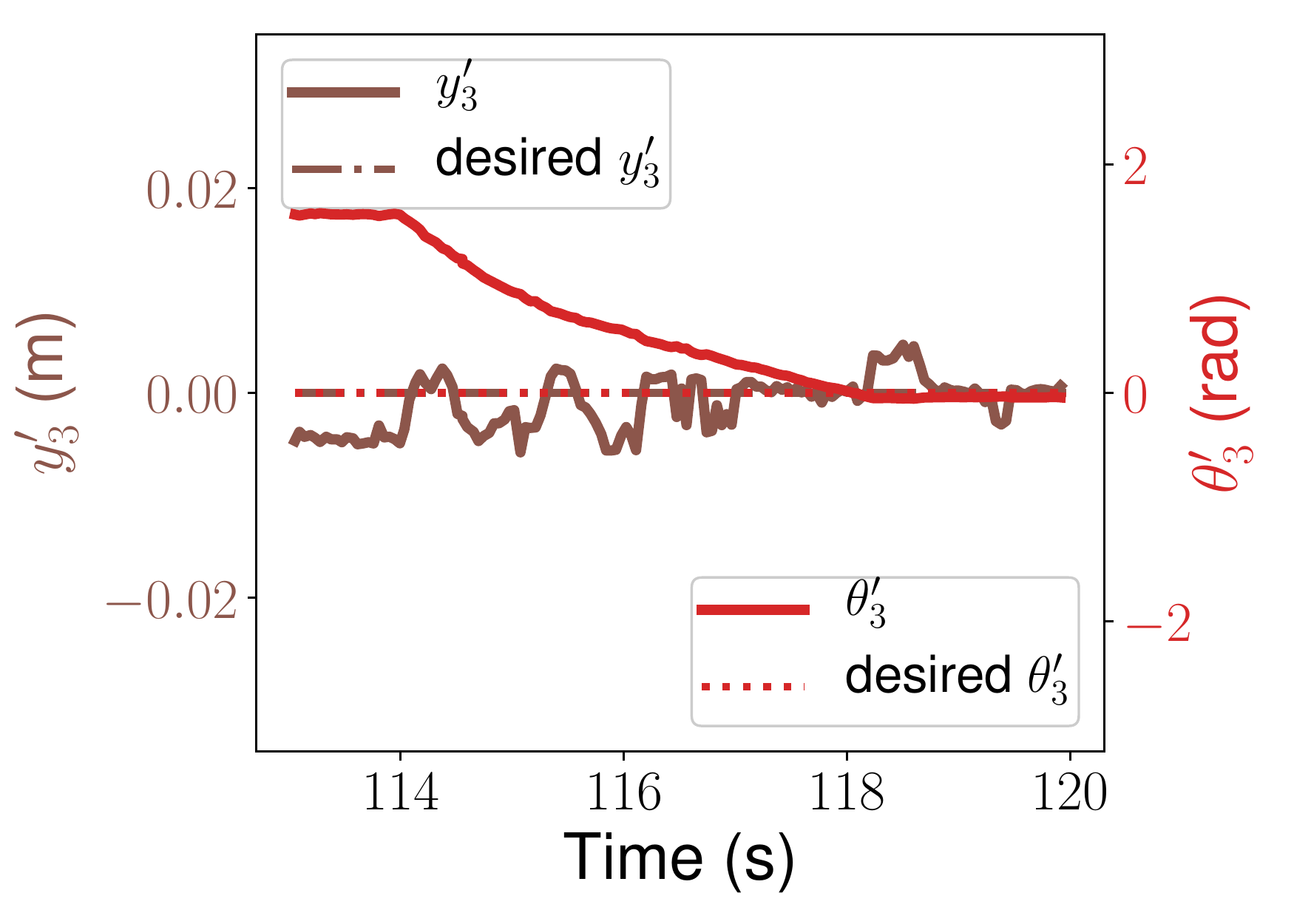}}
    \subfloat[\label{fig:x_ctrl}]{\includegraphics[width=0.24\textwidth]{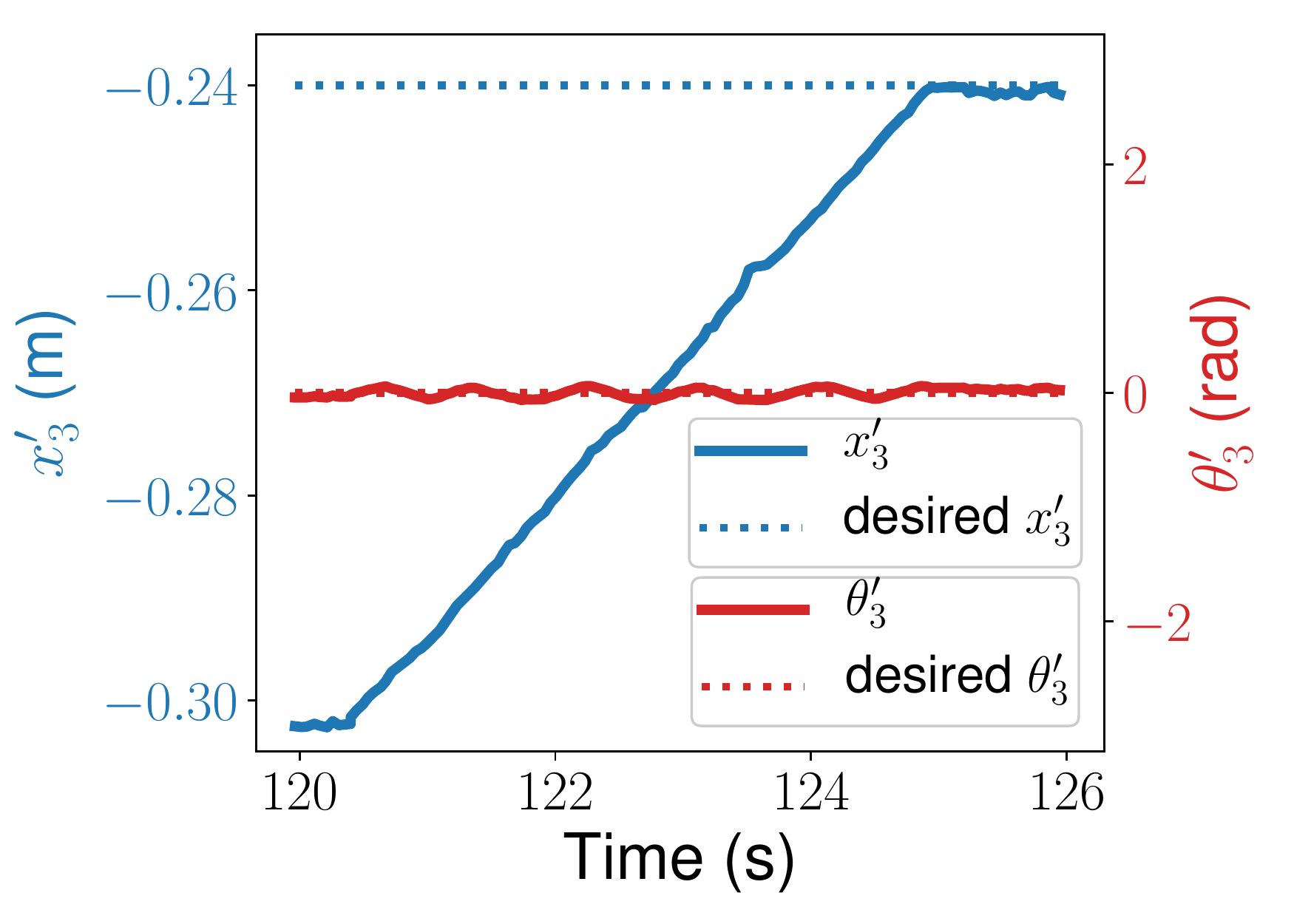}}
    \caption{Pose adjustment of Module 3 before docking in Task 1: (a)
      adjusting $y_3^\prime$ and $\theta_3^\prime$; (b) adjusting
      $x_3^\prime$ while maintaining the correct orientation.}
\end{figure}

\begin{figure*}[t]
    \centering
    \subfloat[]{\includegraphics[height=0.32\textwidth]{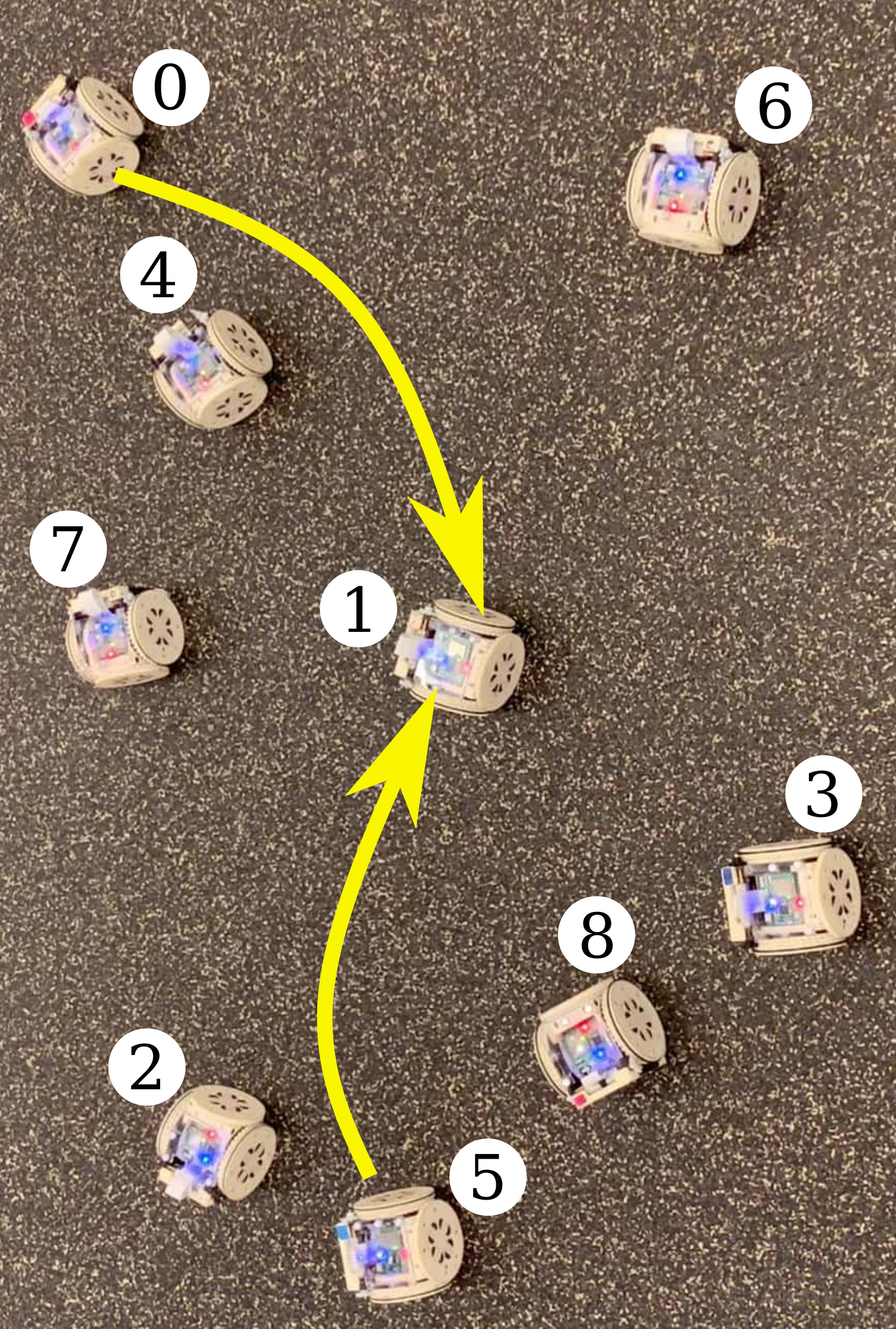}\label{fig:task2-step1}}
    \hfill
    \subfloat[]{\includegraphics[height=0.32\textwidth]{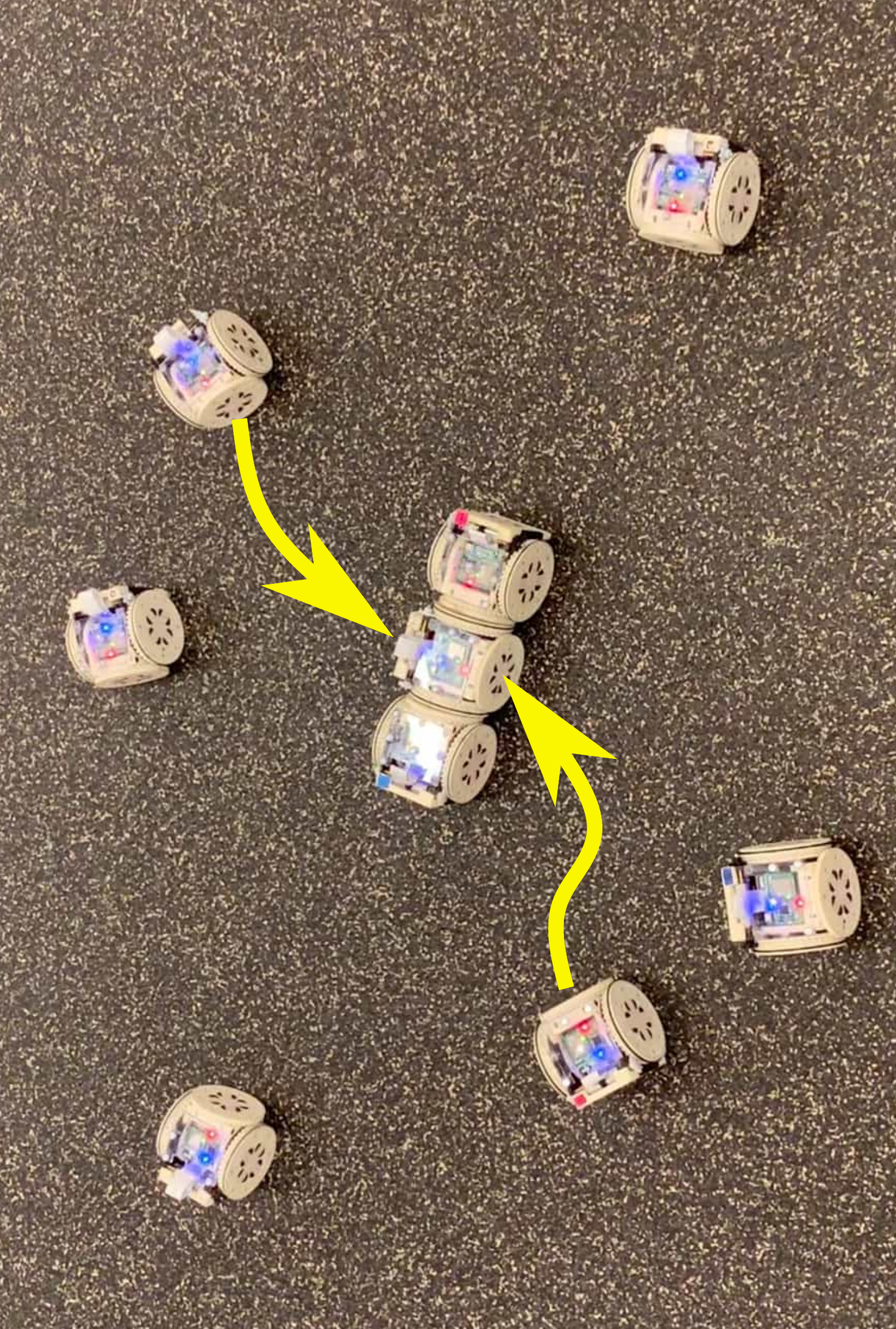}\label{fig:task2-step2}}
    \hfill
    \subfloat[]{\includegraphics[height=0.32\textwidth]{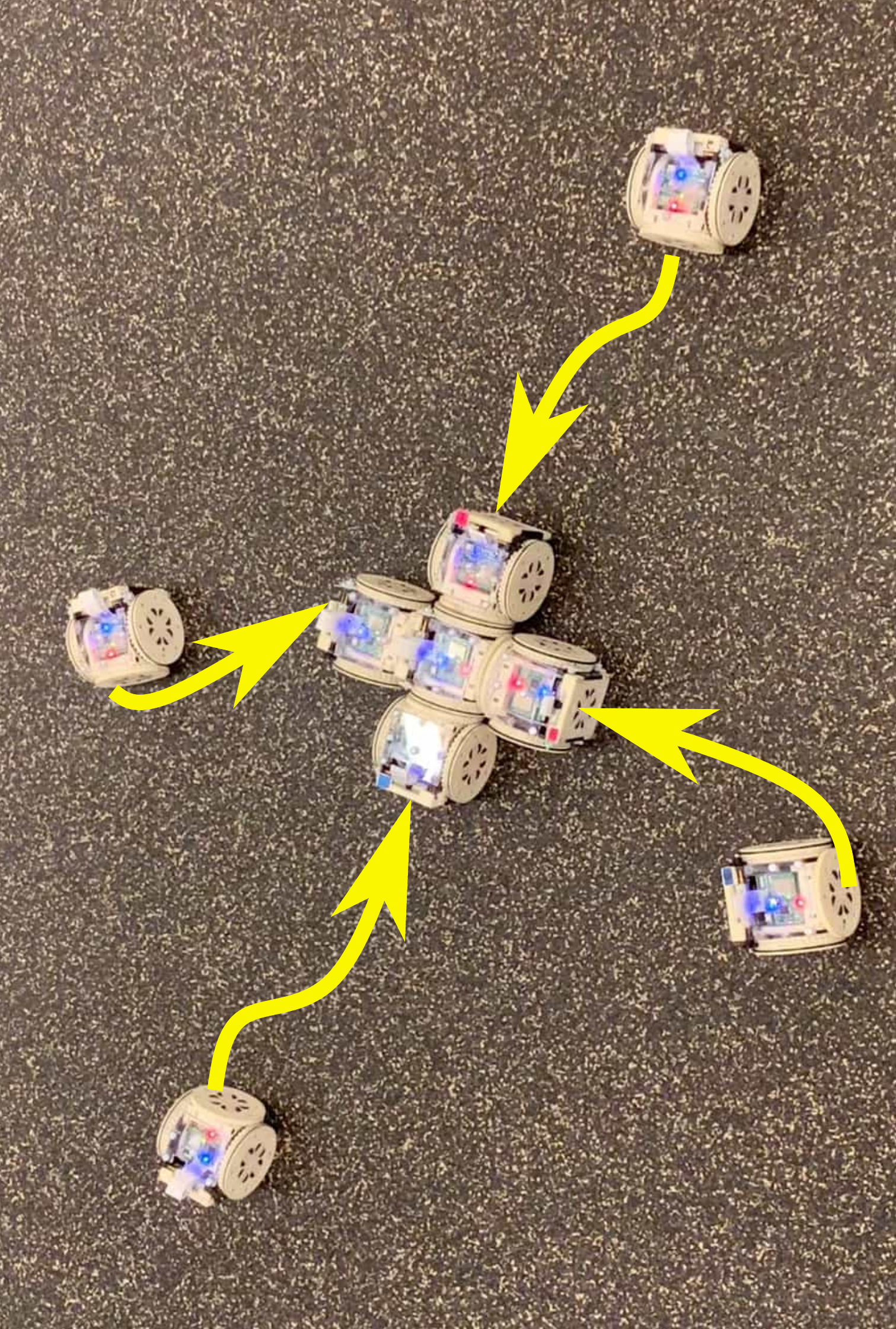}\label{fig:task2-step3}}
    \hfill
    \subfloat[]{\includegraphics[height=0.32\textwidth]{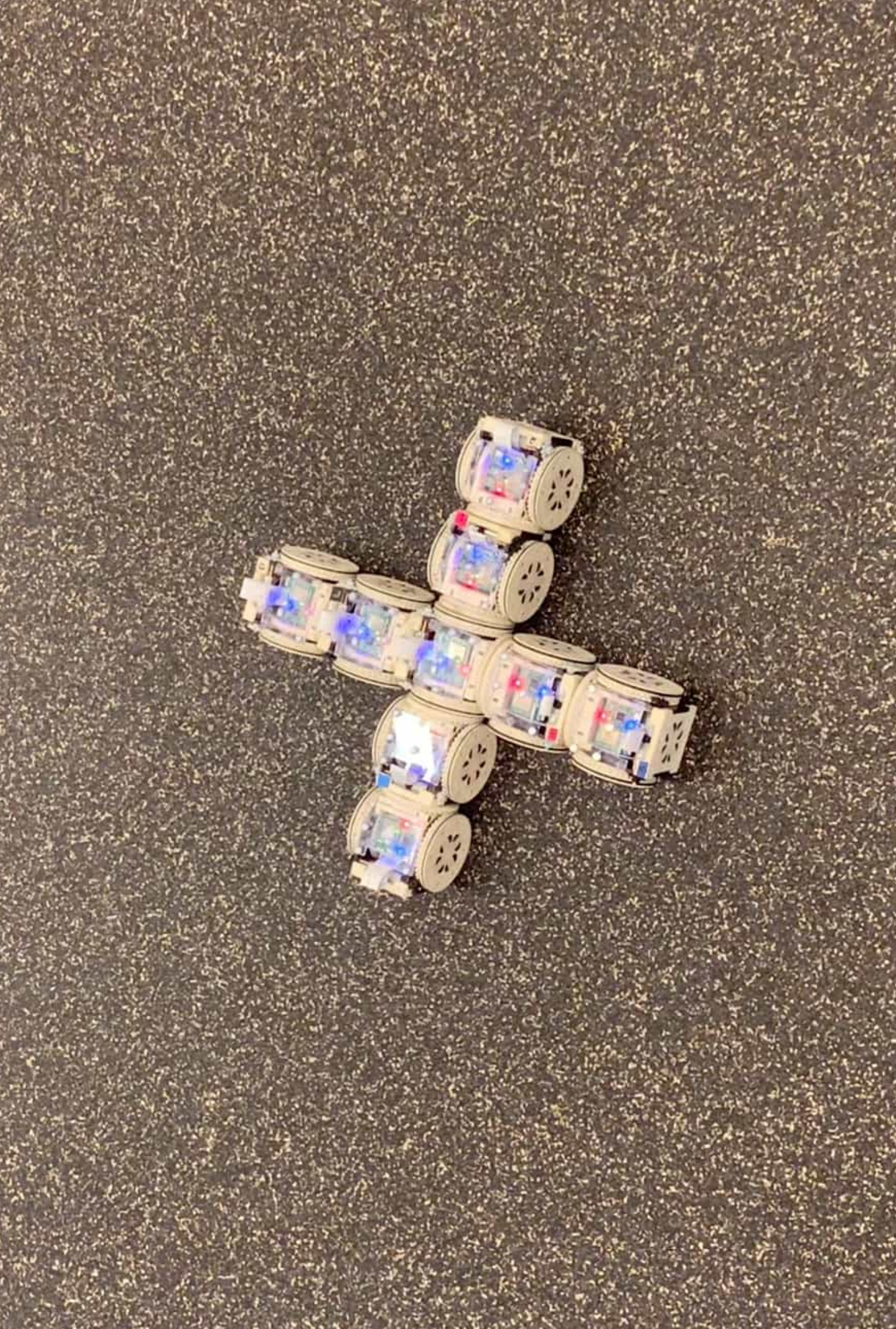}\label{fig:task2-assembly}}\\
    \subfloat[]{\includegraphics[width=0.3\textwidth]{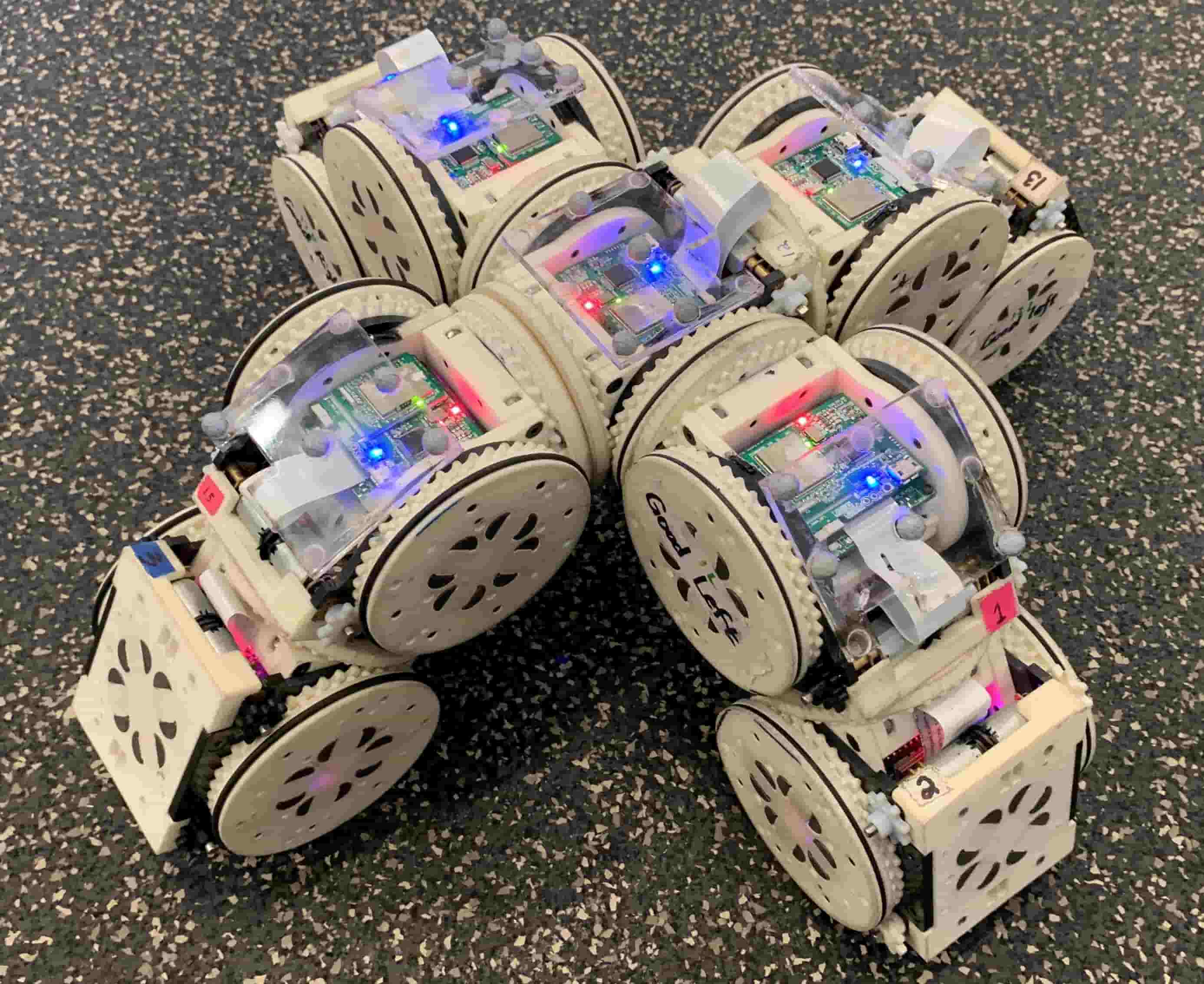}\label{fig:task2-goal}}
    \qquad
    \subfloat[\label{fig:task2-unity}]{\includegraphics[width=0.262\textwidth]{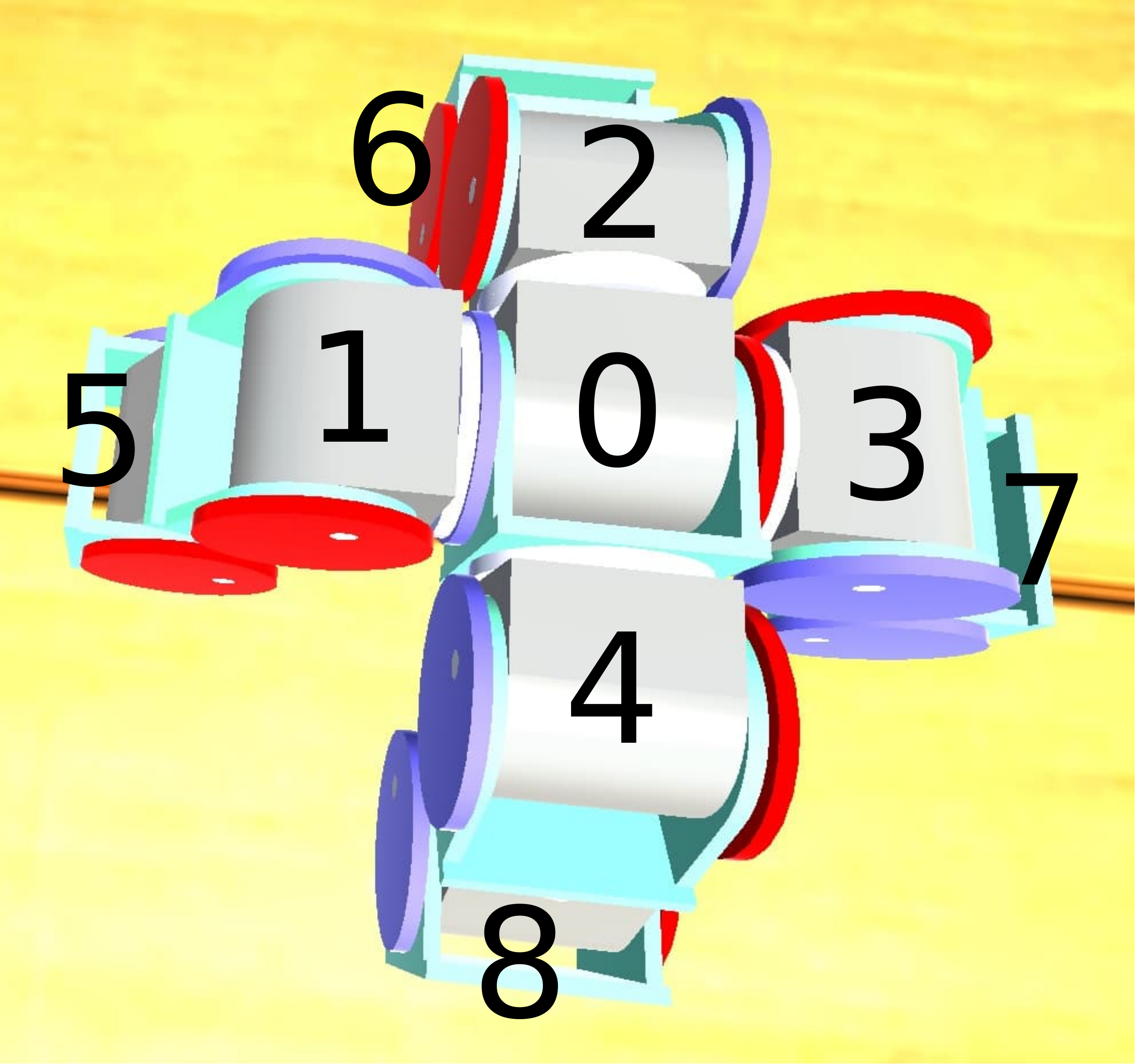}}
    \caption{SMORES-EP hardware holonomic vehicle self-assembly: (a)
      Execute assembly actions
      $(0, \text{\textbf{T}}, 1, \text{\textbf{L}})$ and
      $(5, \text{\textbf{T}}, 1, \text{\textbf{R}})$; (b) Execute
      assembly actions $(4, \text{\textbf{T}}, 1, \text{\textbf{B}})$
      and $(8, \text{\textbf{T}}, 1, \text{\textbf{T}})$; (c) Execute
      assembly actions $(3, \text{\textbf{T}}, 8, \text{\textbf{B}})$,
      $(2, \text{\textbf{T}}, 5, \text{\textbf{B}})$,
      $(7, \text{\textbf{T}}, 4, \text{\textbf{B}})$, and
      $(6, \text{\textbf{T}}, 0, \text{\textbf{B}})$. (d) --- (e)
      The final assembly. (f) The target kinematic topology.}
    \label{fig:task2-process}
\end{figure*}

\begin{figure}[t]
    \centering
    \includegraphics[width=0.4\textwidth]{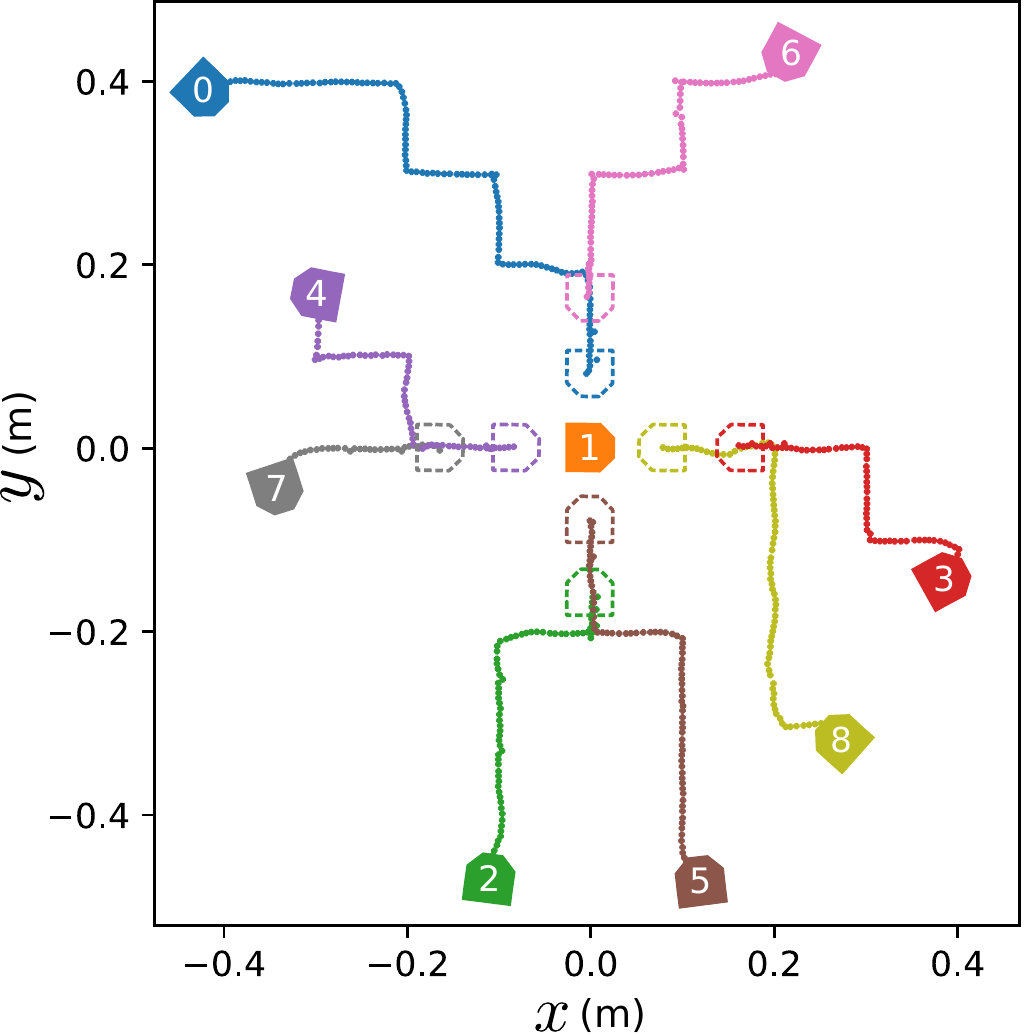}
    \caption{Actual path of each module for Task 2.}
    \label{fig:task2-path}
\end{figure}

In a docking process, the controller for pose adjustment and approach
ensures the success of docking. For example, for the last assembly
action $(3, \text{\textbf{T}}, 4, \text{\textbf{B}})$, Module 3 first
adjusts its body frame (Fig.~\ref{fig:adj-1}) so that its TOP Face is
aligned with BOTTOM Face of Module 4 (Fig.~\ref{fig:adj-2}) and the
performance is shown in Fig.~\ref{fig:y_yaw_ctrl}. The controller can
adjust the pose of Module 3 quickly and align its connector within
\SI{6}{\second} while almost keeping $x_3^\prime$ fixed. Then Module 3
moves forward to Module 4 for final docking (Fig.~\ref{fig:adj-3}) and
the performance is shown in Fig.~\ref{fig:x_ctrl}. It can be seen that
Module 3 can steadily approach Module 4 while maintaining its
orientation fixed (very little oscillation of $\theta_3^\prime$ around zero
value) so that the involved connector is always aligned.

\begin{figure}[t]
  \centering
    \subfloat[]{\includegraphics[width=0.23\textwidth]{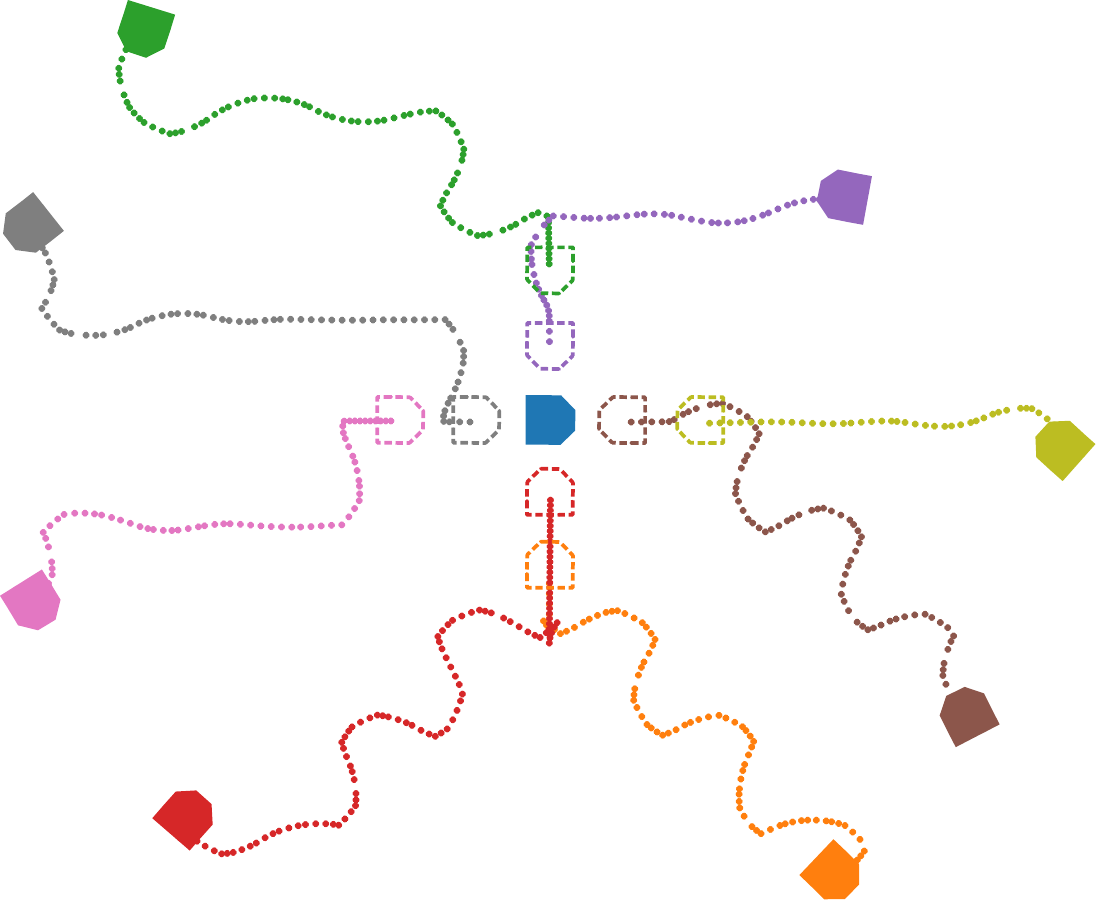}}
    \hfill
    \subfloat[]{\includegraphics[width=0.22\textwidth]{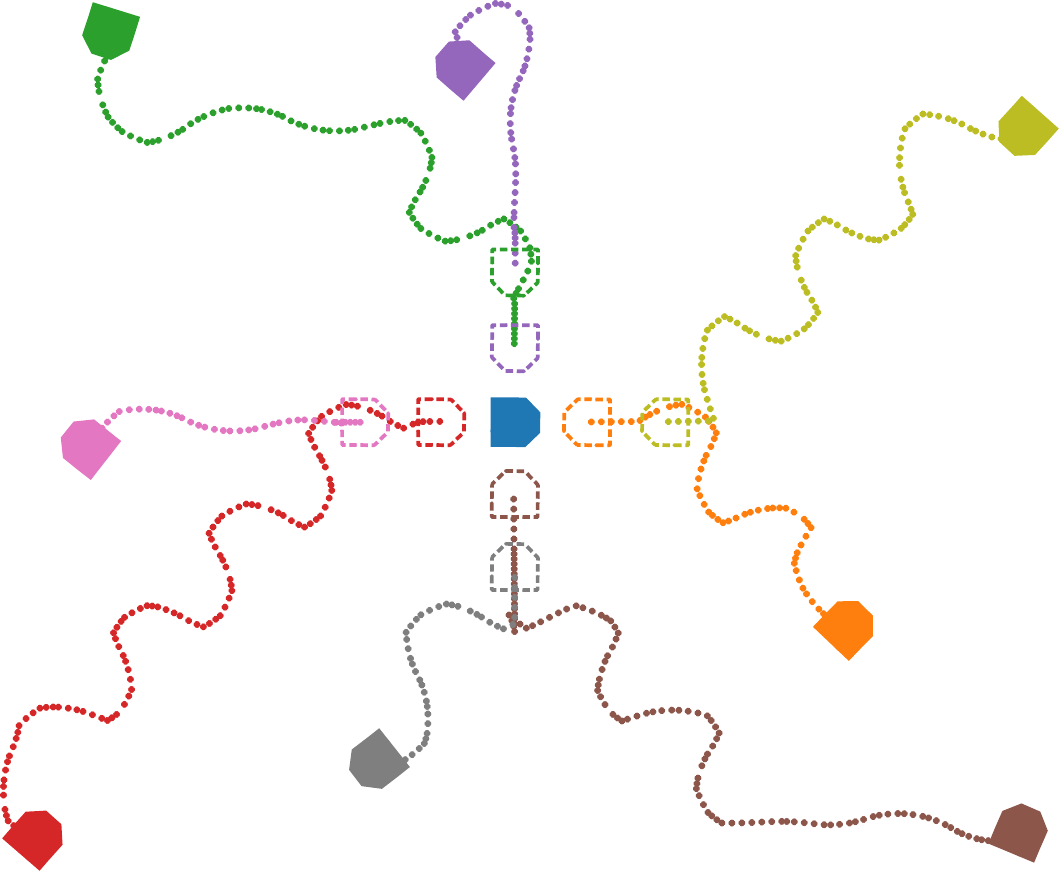}}
    \caption{Two additional experiments for Task 2.}
    \label{fig:task2-more}
\end{figure}

\begin{figure*}[t]
    \centering
    \subfloat[\label{fig:task3-step1}]{\includegraphics[height=0.2\textwidth]{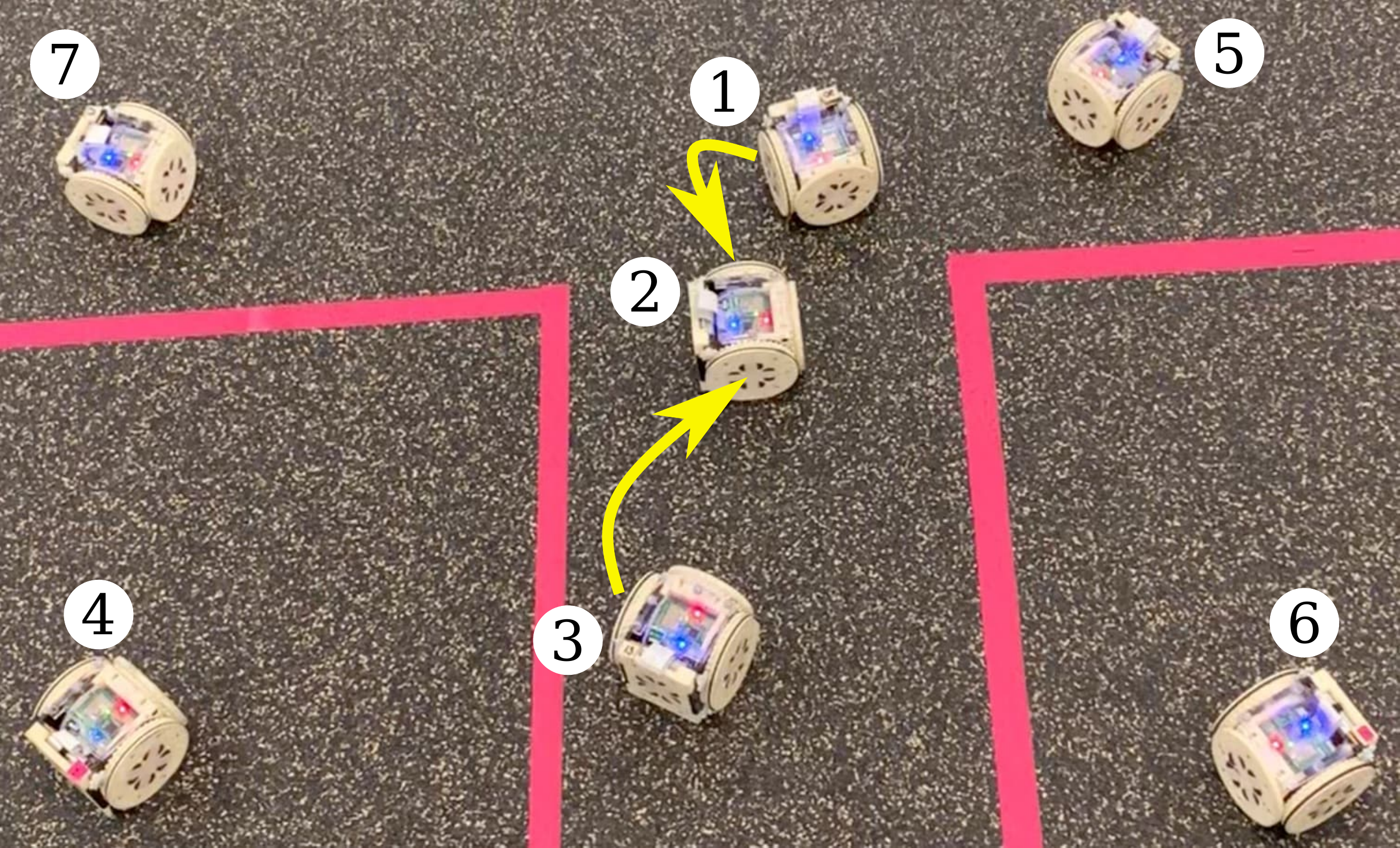}}
    \hfill
    \subfloat[\label{fig:task3-step2}]{\includegraphics[height=0.2\textwidth]{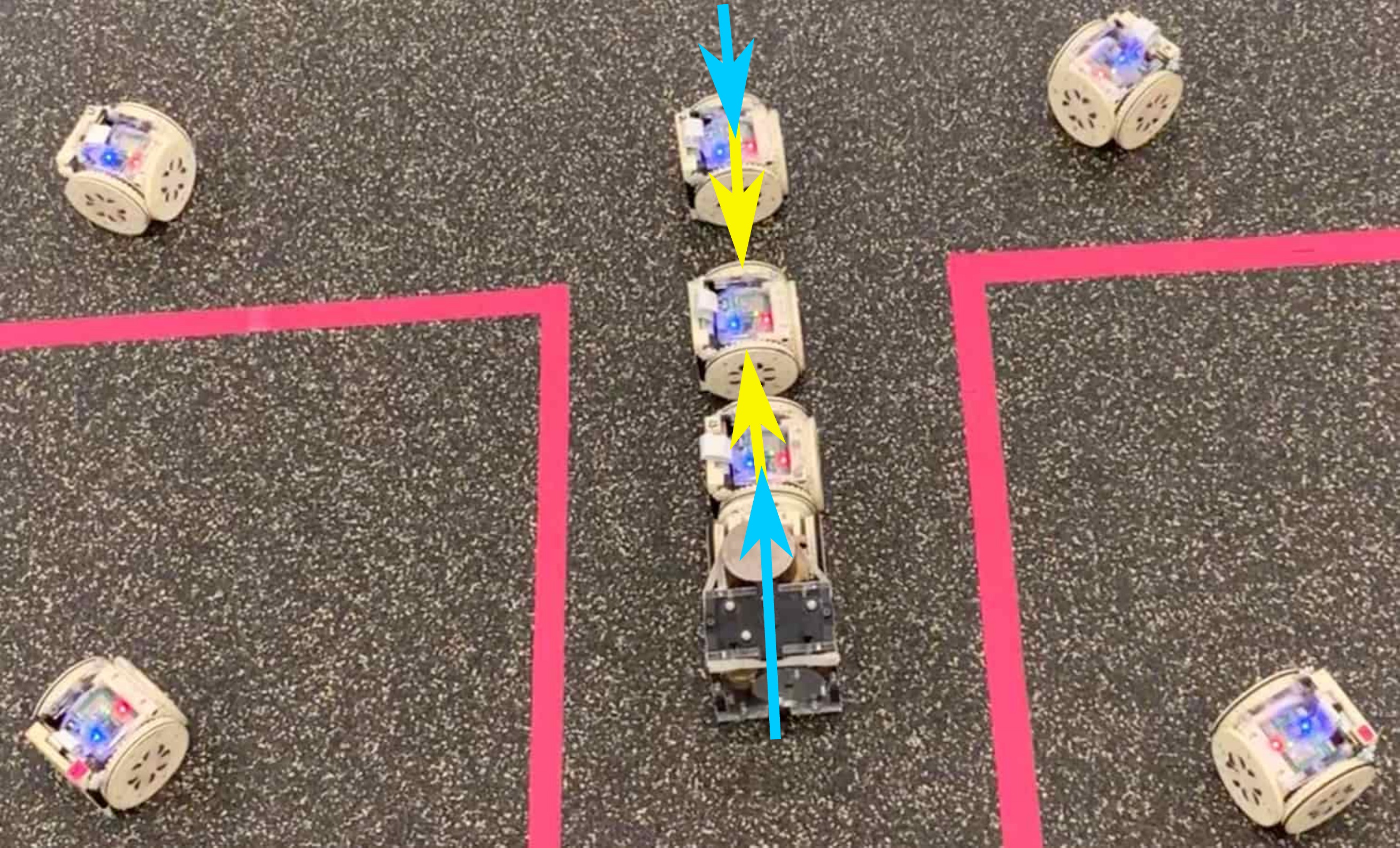}}
    \hfill
    \subfloat[\label{fig:task3-step3}]{\includegraphics[height=0.2\textwidth]{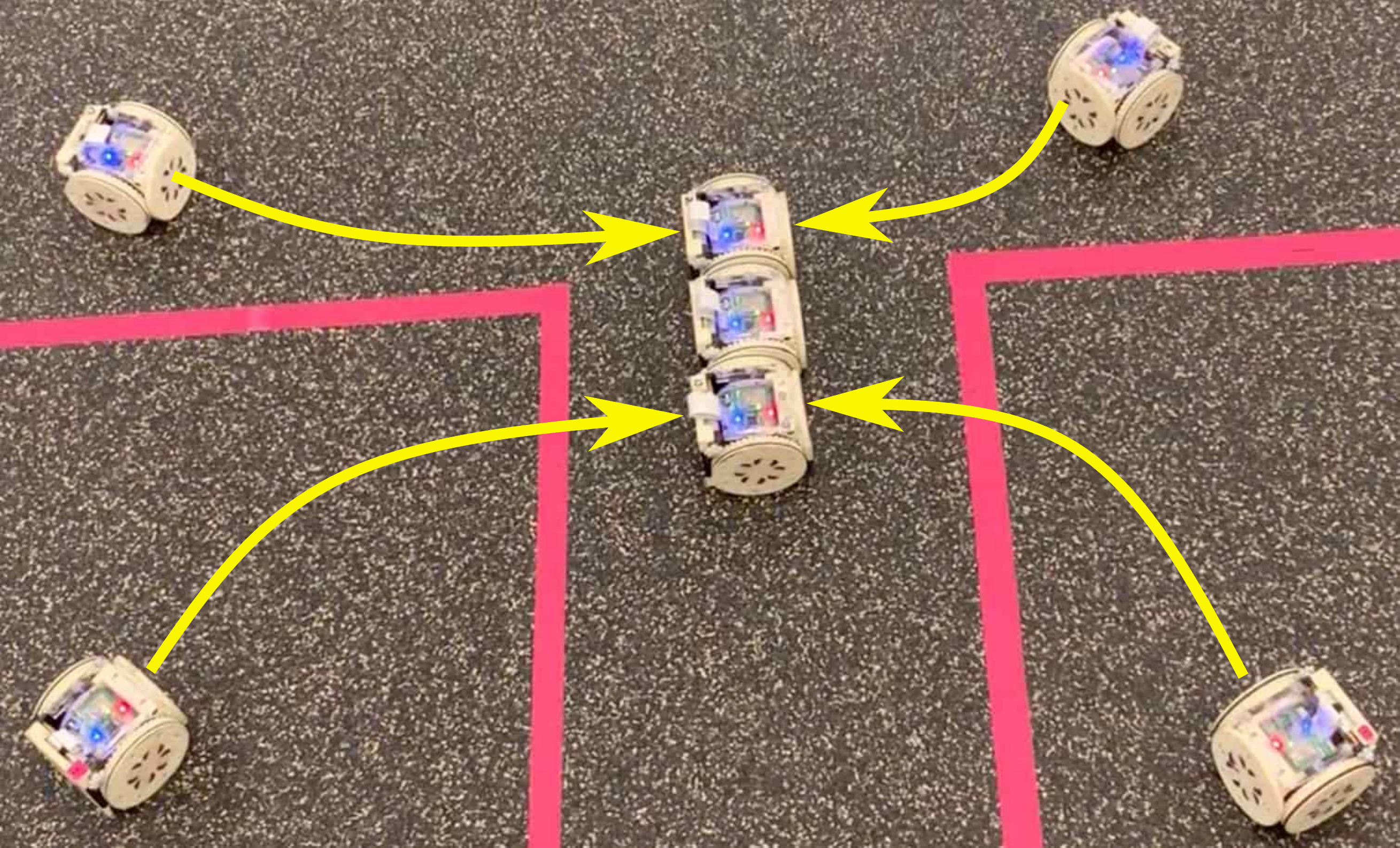}}\\
    \subfloat[\label{fig:task3-assembly}]{\includegraphics[height=0.2\textwidth]{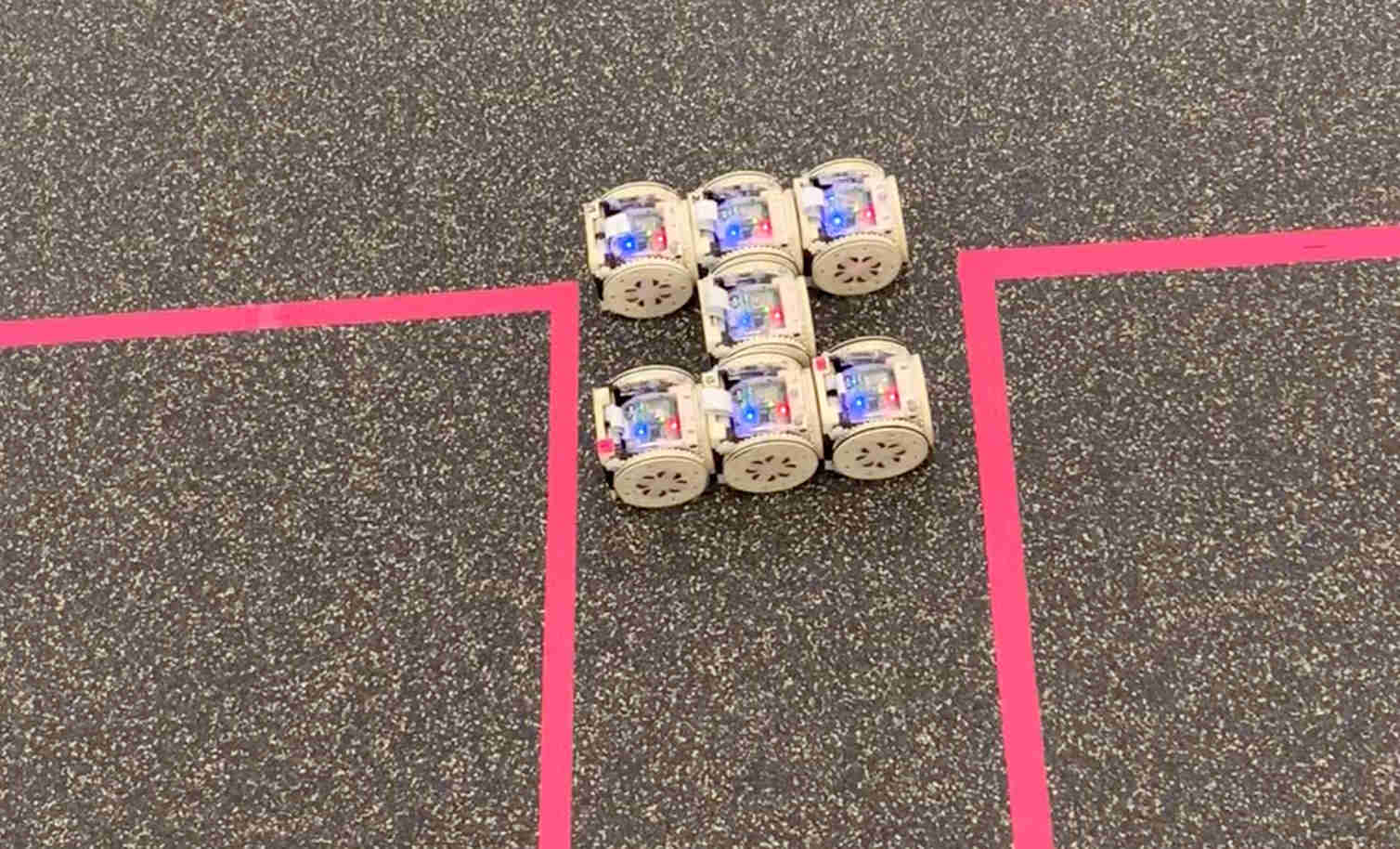}}
    \hfill
    \subfloat[\label{fig:task3-goal}]{\includegraphics[height=0.2\textwidth]{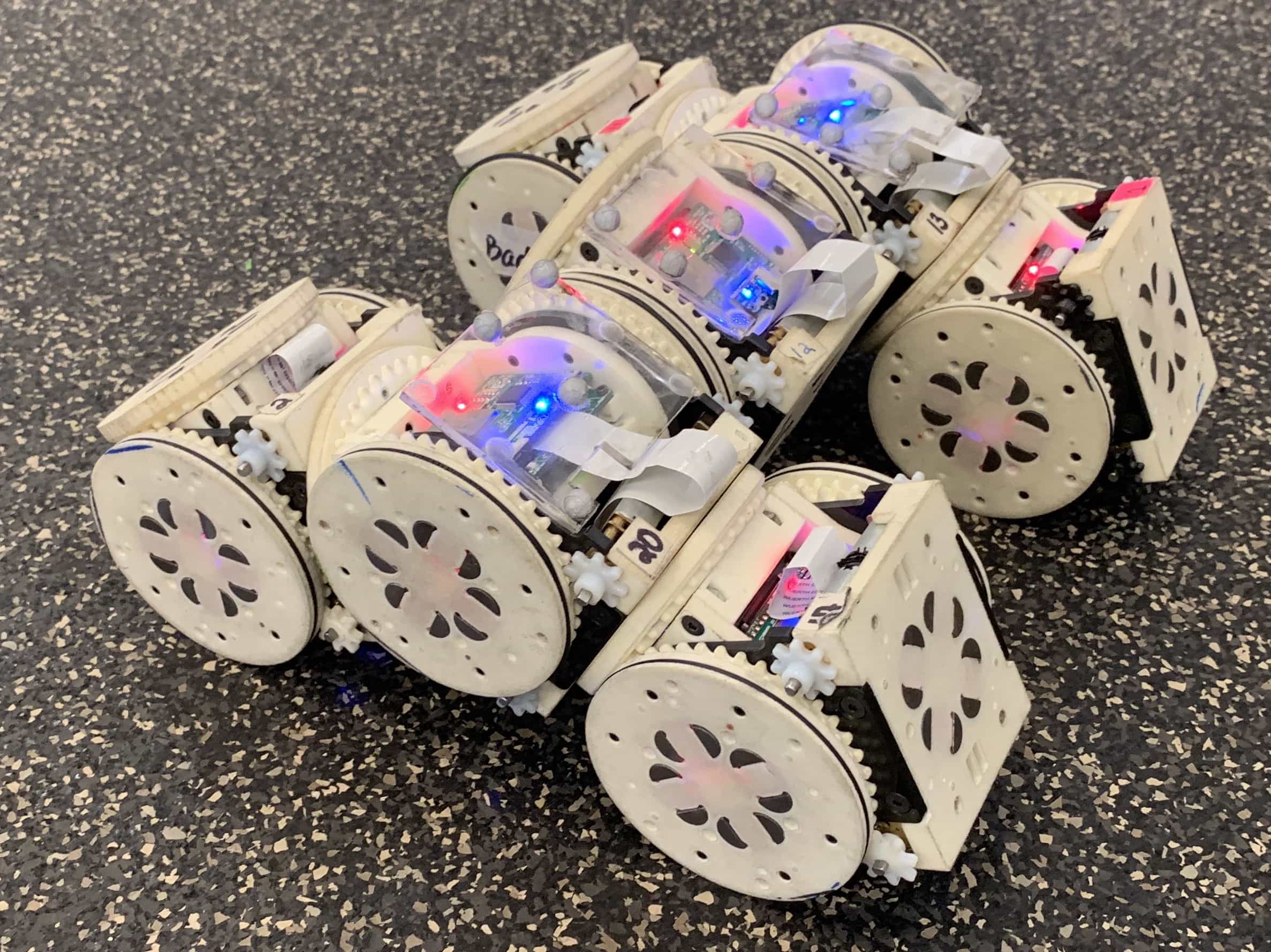}}
    \hfill
    \subfloat[\label{fig:task3-unity}]{\includegraphics[height=0.2\textwidth]{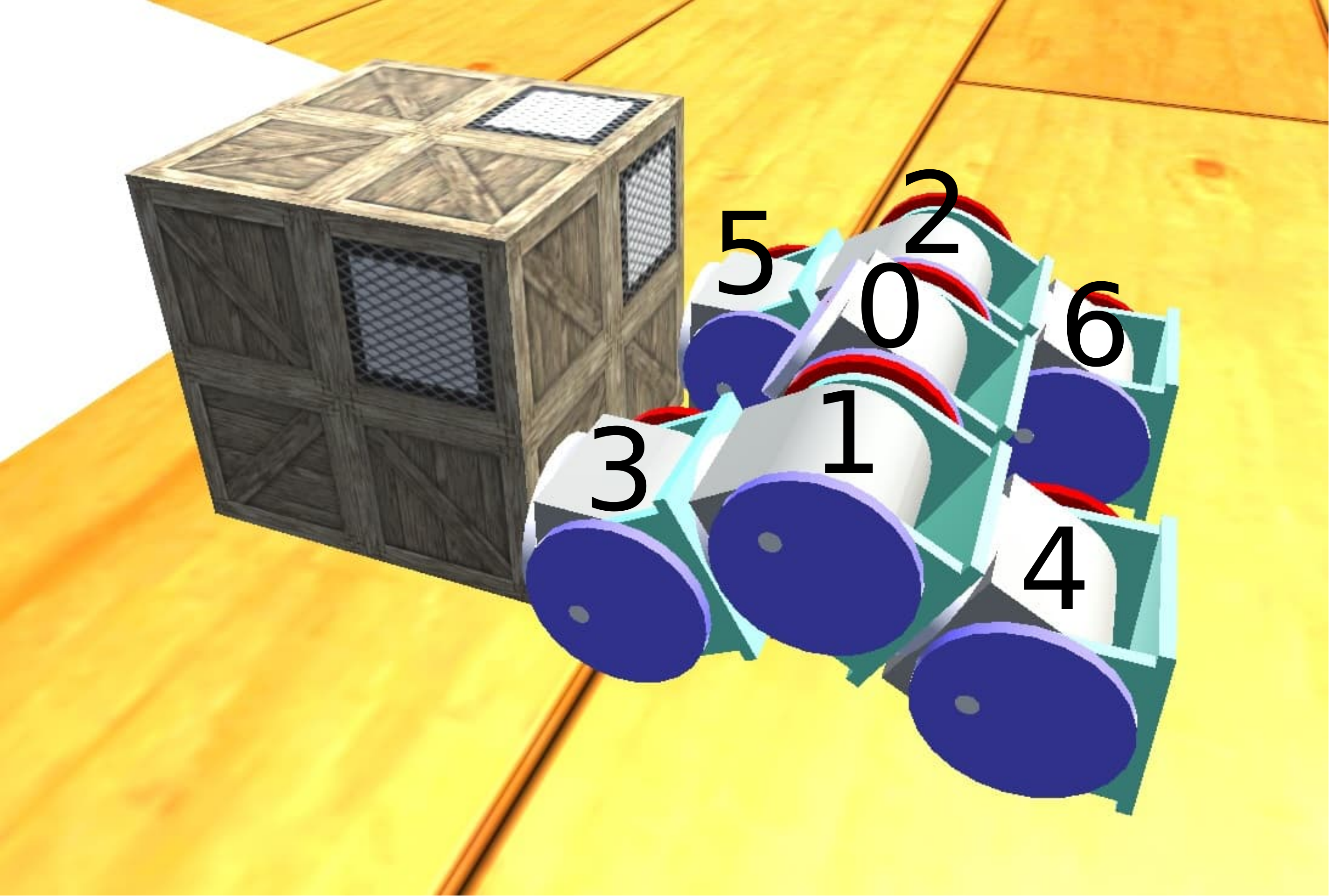}}
    \caption{SMORES-EP hardware RC car self-assembly: (a) Execute
      actions $(1, \text{\textbf{R}}, 2, \text{\textbf{L}})$ and
      $(3, \text{\textbf{L}}, 2, \text{\textbf{R}})$; (b) A helping
      module is used to execute the current docking action; (c) Execute
      actions $(4, \text{\textbf{T}}, 3, \text{\textbf{B}})$,
      $(7, \text{\textbf{T}}, 1, \text{\textbf{B}})$,
      $(6, \text{\textbf{B}}, 3, \text{\textbf{T}})$, and
      $(5, \text{\textbf{B}}, 1, \text{\textbf{T}})$. (d) --- (e) The final assembly. (f) The target kinematic topology.}
    \label{fig:task3-process}
\end{figure*}

\subsection{Task 2: Holonomic Vehicle}

\begin{table}[t]
    \centering
    \caption{Initial Locations of All Modules in Task 2}
    \label{tab:task2-init-pose}
    \begin{tabular}{c c c c}
    \toprule
    Module &   $x$(m)  &   $y$(m)  & $\theta$(rad) \\
    \midrule
    Module 0  & -0.421 &  0.388 & -0.760  \\
    Module 1  &  0.0   &  0.0   &  0.0    \\
    Module 2  & -0.110 & -0.469 &  1.445  \\
    Module 3  &  0.386 & -0.143 &  0.509  \\
    Module 4  & -0.343 &  0.082 & -2.961  \\
    Module 5  &  0.118 & -0.472 &  1.700  \\
    Module 6  &  0.215 &  0.428 & -2.058  \\
    Module 7  & -0.342 & -0.044 & -1.249  \\
    Module 8  &  0.270 & -0.317 &  2.416  \\
    \bottomrule
    \end{tabular}
\end{table}

The second task assembles nine modules into a holonomic vehicle in
order to move as in Fig.~\ref{fig:task2-unity}. Based on the initial
locations of all modules, the root module $m_\tau$ is then selected as
Module 1. Then the pose of every module with respect to $m_\tau$ is
computed and shown in Table~\ref{tab:task2-init-pose}. The root module
of the target topology $G=(V,E)$ in Fig.~\ref{fig:task2-unity} is
Module 0. Given Module 0 in the target topology is mapped to Module 1
in this set of modules $\mathcal{M}$ on the ground, the optimal
mapping $f: V\rightarrow \mathcal{M}$ is derived as $0\to 1$,
$1\to 0$, $2\to 8$, $3\to 5$, $4\to 4$, $5\to 6$, $6\to 3$, $7\to 2$,
and $8\to 7$. The assembly process is shown in
Fig. \ref{fig:task2-step1} --- Fig. \ref{fig:task2-step3} and the
final assembly is shown in Fig. \ref{fig:task2-assembly} and
Fig. \ref{fig:task2-goal}. The behavior that the group of assembly
actions is separated into two when root module is involved can be seen
to ensure the success of the docking process. Then the rest of the
assembly actions can be executed in parallel. Module 2, 3, 6, and 7
begin moving at the same time to locations close to their
destinations, adjust their poses, and finally approach to execute
docking actions. It takes 103 seconds in total to finish the whole
assembly process. With our planner and controllers, the recorded
actual path of every module in the experiment is illustrated in
Fig.~\ref{fig:task2-path}. Two additional experiments in simulation
with different initial settings are shown in
Fig.~\ref{fig:task2-more}.

\begin{table}[t]
    \centering
    \caption{Initial Locations of All Modules in Task 3}
    \label{tab:task3-init-pose}
    \begin{tabular}{c c c c}
    \toprule
    Module &   $x$(m)  &   $y$(m)  & $\theta$(rad) \\
    \midrule
    Module 1  &  0.070 &  0.155 & -1.352  \\
    Module 2  &  0.0   &  0.0   &  0.0    \\
    Module 3  & -0.066 & -0.250 &  0.997  \\
    Module 4  & -0.299 &  0.170 &  0.811  \\
    Module 5  &  0.311 &  0.197 & -2.539  \\
    Module 6  &  0.330 & -0.373 & -2.728  \\
    Module 7  & -0.487 &  0.218 & -0.436  \\
    \bottomrule
    \end{tabular}
\end{table}

\begin{figure}[t!]
    \centering
    \includegraphics[width=0.4\textwidth]{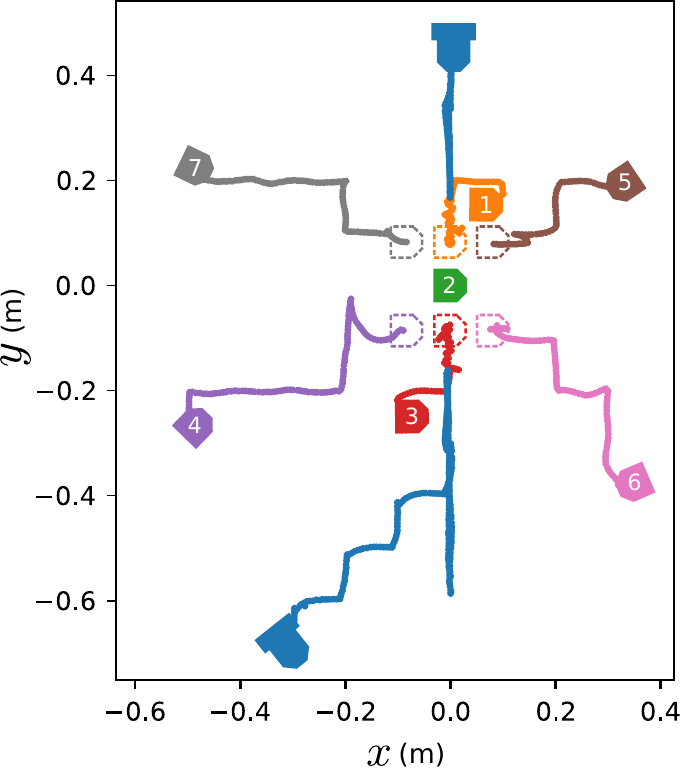}
    \caption{Actual path of each module in Task 3. The blue blocks without number labeled represent the helping modules.}
    \label{fig:task3-path}
\end{figure}

\begin{figure}[t]
  \centering
    \subfloat[]{\includegraphics[width=0.2\textwidth]{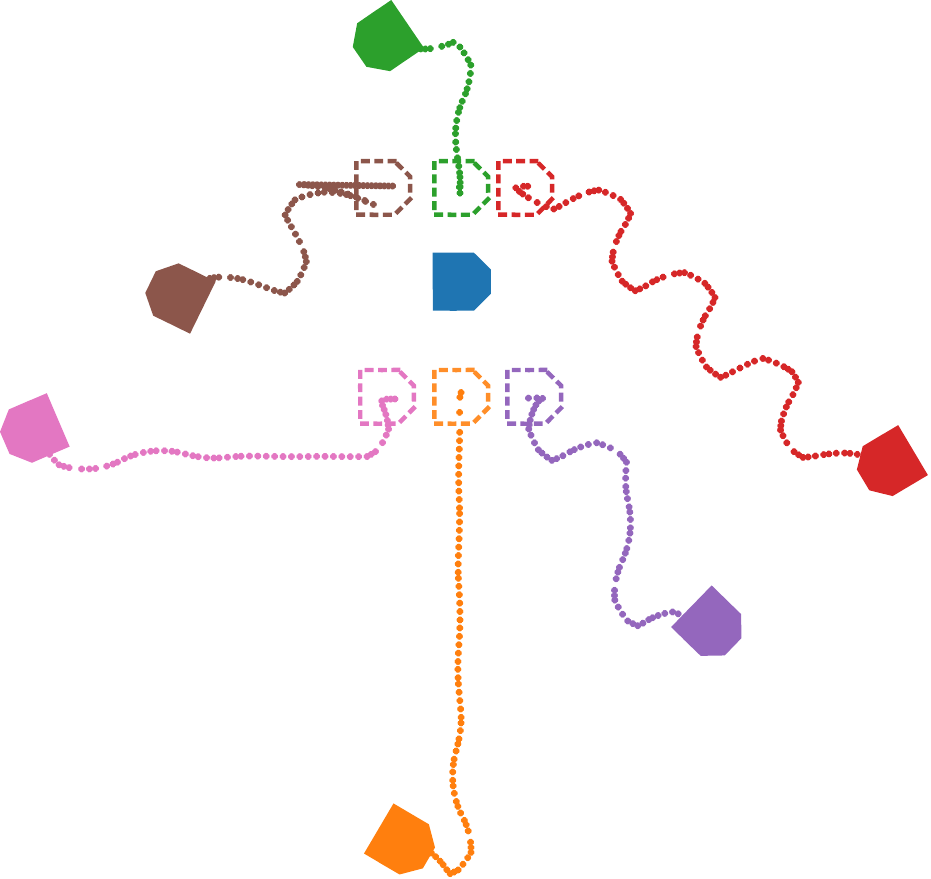}}
    \hfill
    \subfloat[]{\includegraphics[width=0.25\textwidth]{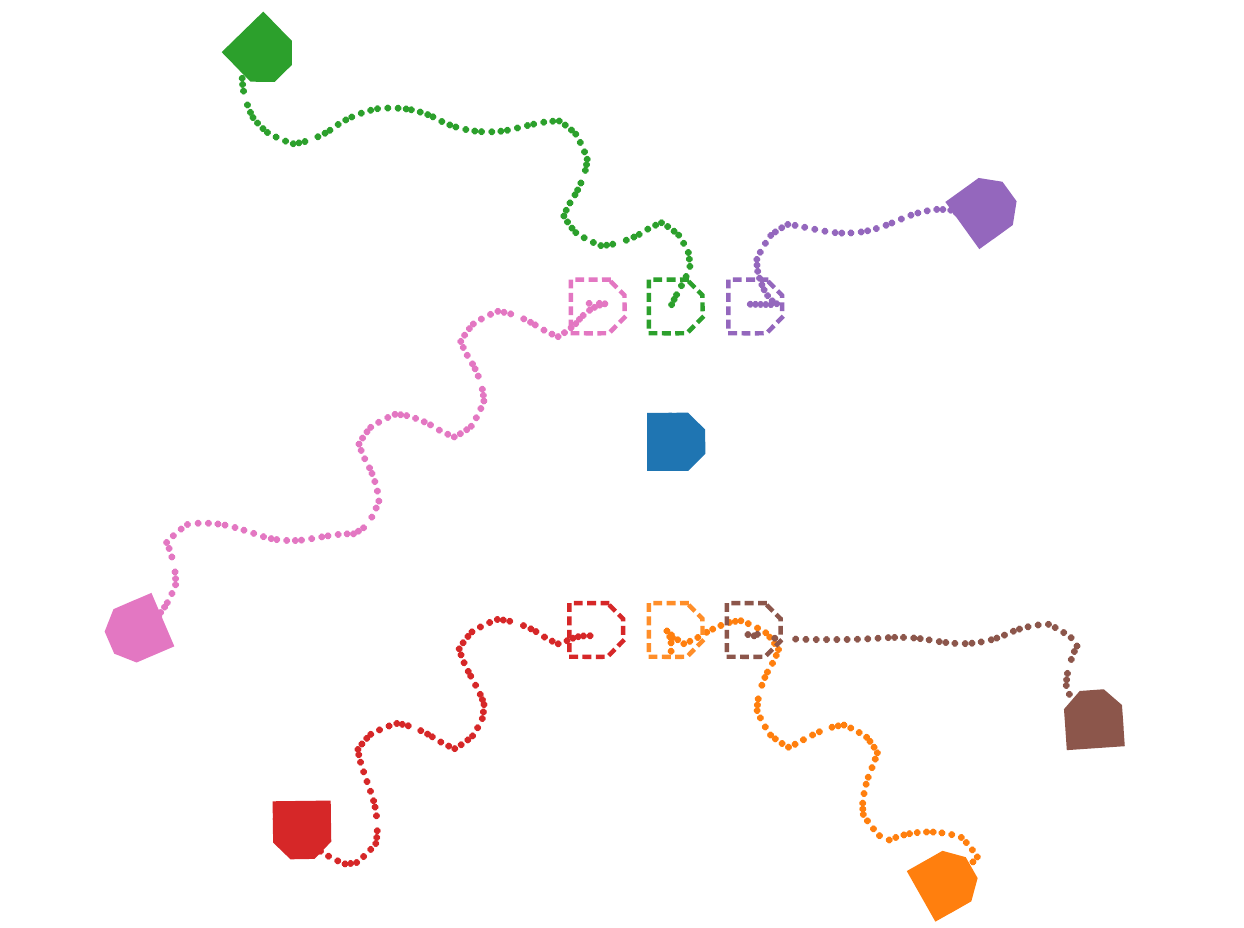}}
    \caption{Two additional experiments for Task 3.}
    \label{fig:task3-more}
\end{figure}

\subsection{Task 3: RC Car}
The last task assembles seven modules into a vehicle in order to push
heavy items shown in Fig.~\ref{fig:task3-unity}. The root module
$m_\tau$ is selected as Module 2 that is the closest module to the
center of the cluster. Then the initial locations of all modules with
respect to $m_\tau$ are shown in Table~\ref{tab:task3-init-pose}. The
root module $\tau$ of the target topology $G=(V,E)$ is Module 0, and
given $m_\tau$ is mapped to $\tau$, the mapping
$f:V\rightarrow \mathcal{M}$ that is $0\to 2$, $1\to 1$, $2\to 3$,
$3\to 5$, $4\to 7$, $5\to 6$, and $6\to 4$ is derived. The assembly
actions are shown in Fig.~\ref{fig:task3-step1} ---
Fig.~\ref{fig:task3-step3}. For the first step, we need to dock Module
1 RIGHT Face with Module 2 LEFT Face and dock Module 3 LEFT Face with
Module 2 RIGHT Face. These two assembly actions cannot be executed
directly since we cannot align the LEFT Face and RIGHT Face directly.
We need the help of a helping module shown in
Fig.~\ref{fig:task3-step2}. In this experiment, there is only one
helping module.  Similar to common docking actions, the helping module
first navigates to a location close to Module 3, then adjusts its pose
to align its TOP Face, and approaches Module 3 RIGHT Face for
docking. Then it lifts Module 3 so that Module 3 can adjust its LEFT
Face. Finally the helping module delivers Module 3 to the desired
location for docking with Module 2 RIGHT Face. This process repeats
for Module 1 RIGHT Face. After both Module 1 and Module 3 are docked
with the root module (Module 2), the rest of four modules can execute
assembly actions in parallel. It takes 260 seconds in total to finish
this assembly task, though 210 seconds are consumed by the helping
module. More helping modules working in parallel would decrease the
duration. The final assembly result is shown in
Fig.~\ref{fig:task3-assembly} and Fig.~\ref{fig:task3-goal}. The
recorded path in the experiment is shown in
Fig. \ref{fig:task3-path}. Two additional experiments in simulation
with different initial settings are also shown in
Fig.~\ref{fig:task3-more}.

\subsection{Reconfiguration Action Parallelization}
\label{sec:reconfig-parallel}

\begin{figure}[t]
  \centering
  \begin{subfloat}[]{
      \includegraphics[height=0.175\textwidth]{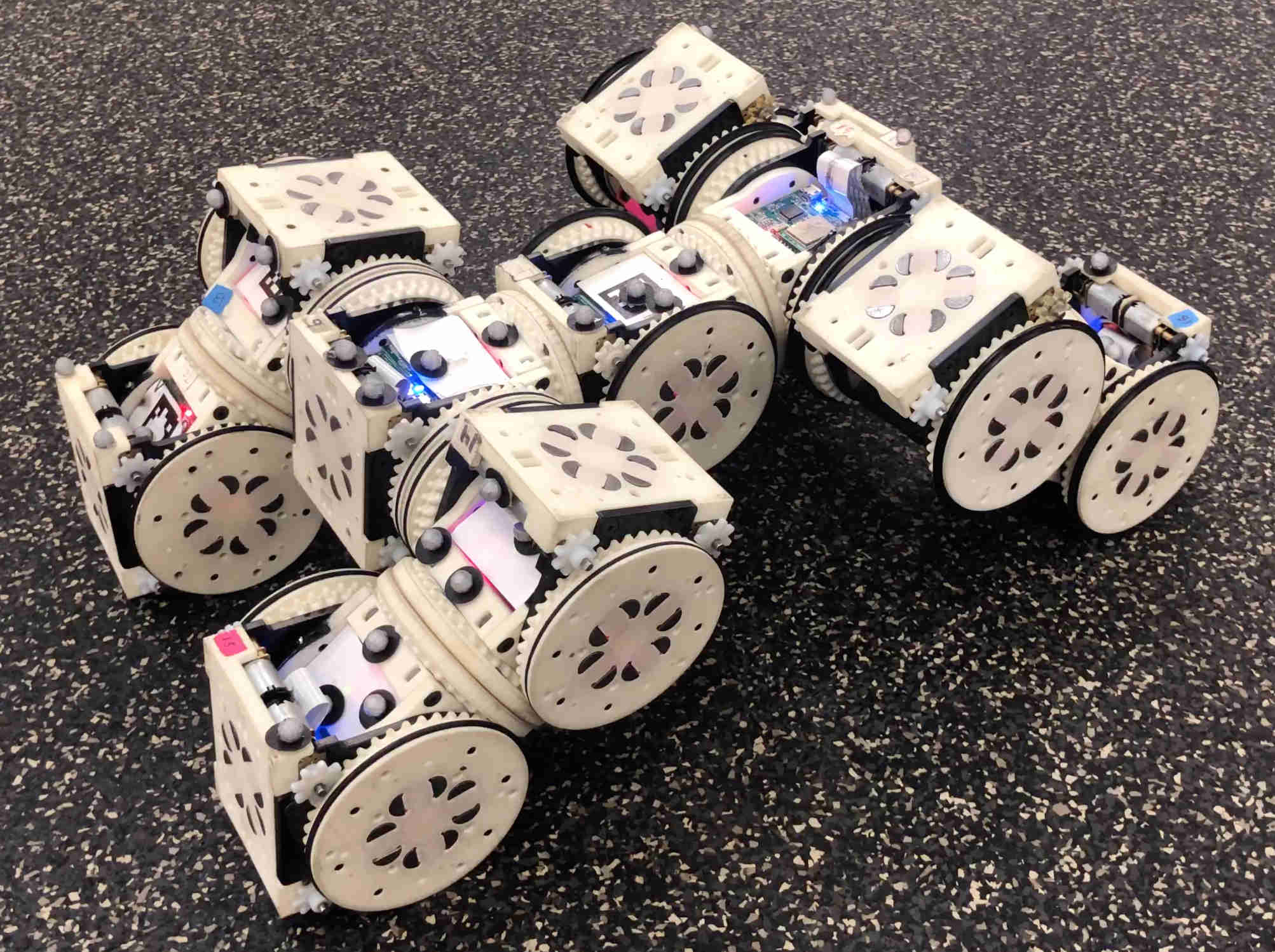}\label{fig:init_hardware}}
  \end{subfloat}
  \begin{subfloat}[]{
      {\includegraphics[height=0.175\textwidth]{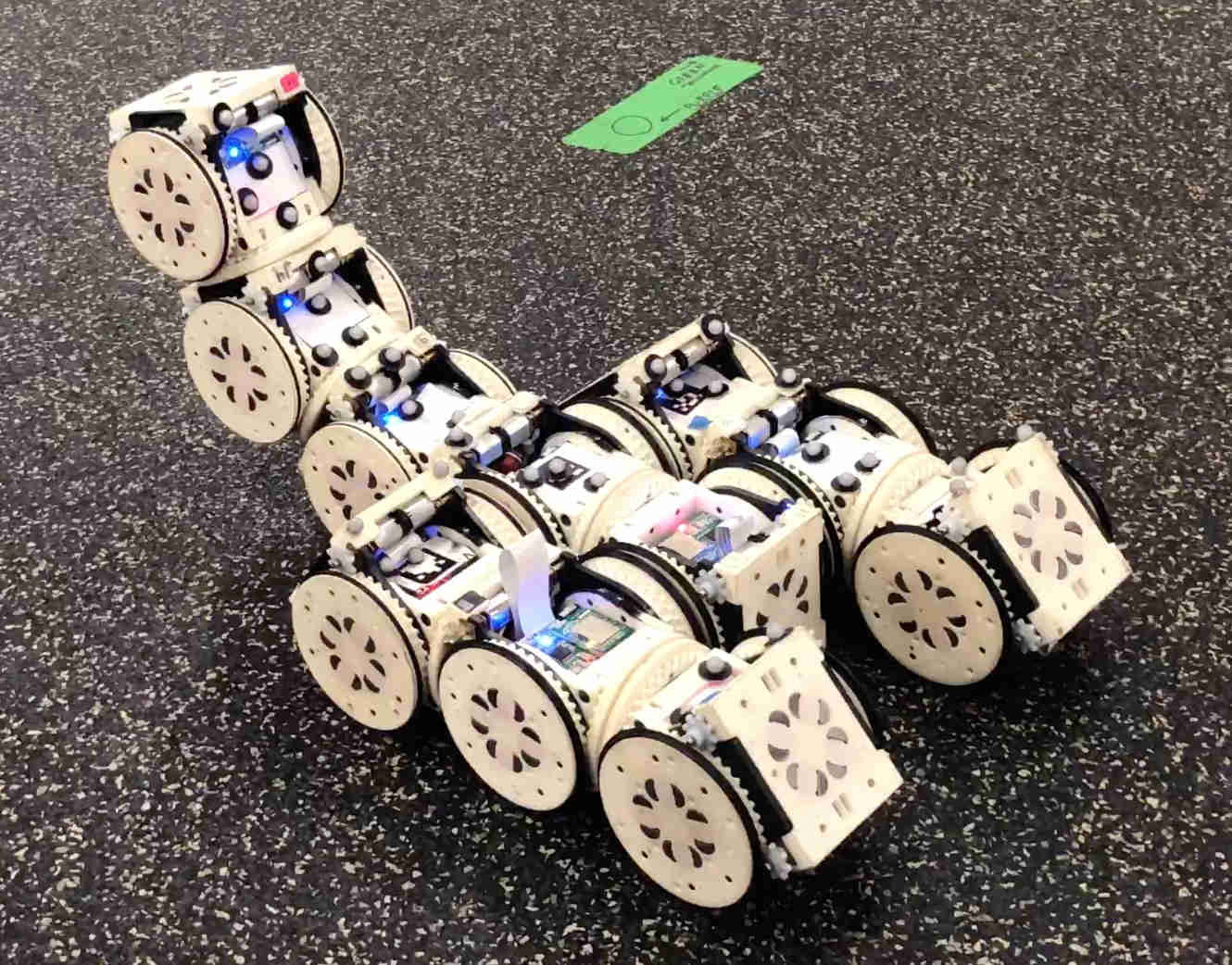}\label{fig:goal_hardware}}}
  \end{subfloat}
  \caption{Reconfigure a walker configuration (a) into a mobile
    vehicle with an arm configuration (b) with eleven SMORES-EP
    modules involved.}\label{fig:reconfig-hardware}
\end{figure}

\begin{figure}[t]
  \centering
  \begin{subfloat}[]{
      \includegraphics[height=0.11\textwidth]{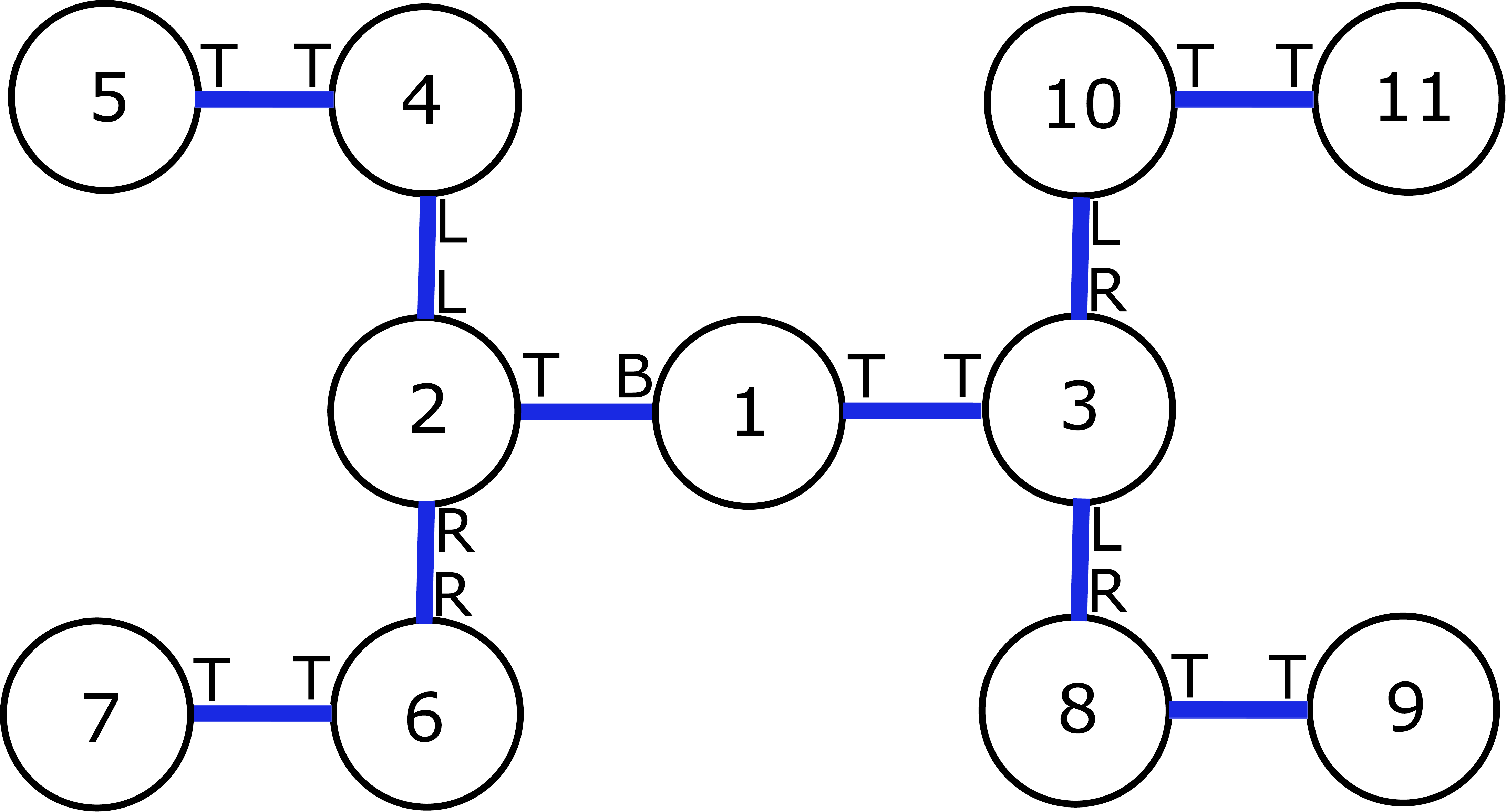}\label{fig:init_graph}}
  \end{subfloat}
  \begin{subfloat}[]{
      \includegraphics[height=0.11\textwidth]{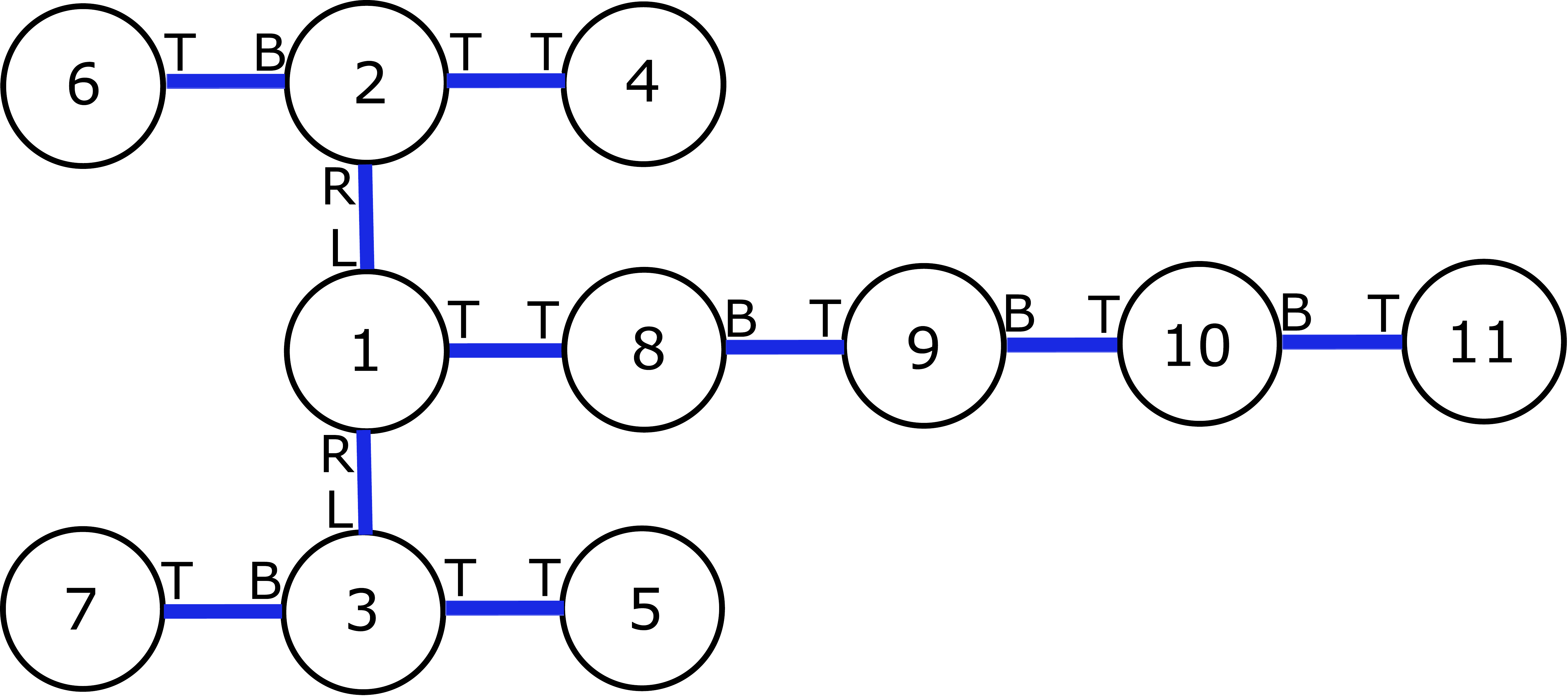}\label{fig:goal_graph}}
  \end{subfloat}
  \caption{(a) The graph representation of the walker configuration. (b)
    The graph representation of the mobile manipulator
    configuration.}\label{fig:reconfig-graph}
\end{figure}

The algorithm by~\cite{Liu-smores-reconfig-ral-2019} can output a
sequence of reconfiguration actions which can be parallelized by our
framework to speed up the process. These reconfiguration actions are
executed from leaf modules to the root module. In addition to the
docking action involved in a self-assembly action, a
self-reconfiguration action also contains an undocking action
beforehand. Given the initial configuration $G_i=(V_i, E_i)$ and goal
configuration $G_g=(V_g, E_g)$, a mapping $f:V_i\rightarrow V_g$ as well as its
inverse $f^{-1}:V_g\rightarrow V_i$ are computed, and each
$v_i\in V_i$ is mapped to a unique $v_g\in V_g$, and
$\tilde{v}_i^{c_i}$ and $\tilde{v}_g^{c_g}$ are their parents
respectively. Then a sequence of reconfiguration actions can be
generated. For example, given the task that is to reconfigure eleven
SMORES-EP modules from a walker (Fig.~\ref{fig:init_hardware}) into a
mobile vehicle with an arm (Fig.~\ref{fig:goal_hardware}), their graph
representations are shown in Fig.~\ref{fig:reconfig-graph} where
$\tau_i$ is Module 1 and $\tau_g$ is Module 1 respectively. The mapping
$f:V_i\rightarrow V_g$ is computed as $1\to 1$, $3\to 8$, $9\to 5$,
$8\to 3$, $2\to 9$, $11\to 4$, $10\to 2$, $4\to 6$, $6\to 10$,
$5\to 7$, and $7\to 11$, and the corresponding reconfiguration actions are
shown in Table~\ref{tab:reconfig-actions}.

\begin{table}[t]
  \caption{Reconfiguration Actions for Task 1}
  \label{tab:reconfig-actions}
  \begin{center}
    \begin{tabular}{|c|c|c|c|c|}
      \hline
      Action & ID & Face & ID & Face \\
      \hhline{|=|=|=|=|=|}
      Undock & 5 & TOP Face & 4 & TOP Face\\
      Dock   & 5 & TOP Face & 8 & BOTTOM Face\\
      \hline
      Undock & 7 & TOP Face & 6 & TOP Face\\
      Dock   & 7 & TOP Face & 6 & BOTTOM Face\\
      \hline
      Undock & 4 & LEFT Face & 2 & LEFT Face\\
      Dock   & 4 & TOP Face & 10 & BOTTOM Face\\
      \hline
      Undock & 6 & RIGHT Face & 2 & RIGHT Face\\
      Dock   & 6 & TOP Face & 2 & BOTTOM Face\\
      \hline
    \end{tabular}
  \end{center}
\end{table}

As shown by~\cite{Liu-smores-reconfig-ral-2019}, these actions can be
executed in sequential manner. This process can be parallelized by
formulating all docking actions as a self-assembly problem. We first
execute all undocking actions in Table~\ref{tab:reconfig-actions} so
that these modules are free to move. Then run the parallel algorithm
(Algorithm~\ref{alg:parallel-assembly}) in which the target topology
is the goal configuration, and the mapping is simply
$f^{-1}:V_g\rightarrow V_i$ derived from the reconfiguration
planner. The algorithm starts from modules in the goal configuration
with depth being 1, but we ignore existing connections which are not
constructed by docking actions in Table~\ref{tab:reconfig-actions}. By
the mapping, the depth of Module 4 and Module 5 in the goal
configuration are both 2, the depth of Module 6 is 3, and the depth of
Module 7 is 4. Hence, the output of the assembly process is first
$(5, \text{\textbf{T}}, 8, \text{\textbf{B}})$ and
$(4, \text{\textbf{T}}, 10, \text{\textbf{B}})$ in parallel, then
execute $(6, \text{\textbf{T}}, 2, \text{\textbf{B}})$, finally
execute $(7, \text{\textbf{T}}, 6, \text{\textbf{B}})$. Some docking
action may require other modules to move away to provide clearance as
demonstrated by~\cite{Liu-smores-reconfig-ral-2019}. The strategy can
also be used to handle self-assembly scenarios where modules are very
close to each other.

\section{Conclusion}
\label{sec:conclude}

In this paper, we present a parallel modular robot self-assembly
framework for kinematic topology which can significantly improve the
capability of modular robots to interact with the
environment. SMORES-EP is a hybrid self-reconfigurable modular robot
system. Each module has four DOFs and four EP-Face connectors in a
compact space, and a swarm of modules are controlled by a hybrid
architecture. Given a target kinematic topology, modules are mapped by
those in the target configuration in an optimal way and then the
assembly actions can be computed and executed in parallel. Motion
controllers are developed to ensure the success of docking among
modules. Hardware demonstrations show the effectiveness and robustness
of the framework. This framework can also speed up the
self-reconfiguration process by formulating all necessary docking
actions as a self-assembly problem.

Future work includes creating demonstrations of arbitrary 3D
structures with the SMORES-EP system. This is a capability which, in
principle, should be possible with small changes to the algorithm, but
is much harder to demonstrate using hardware with the concomitant
complications of constraints from actuator limitations for lifting
modules. Simulations with much larger structures would also
demonstrate the scalability of this algorithm. Moreover, additional
work needs to be done in order to move the system into the wild,
including outdoor localization and wireless communication. The current
hardware control architecture performs well but may encounter
scalability issues when many more modules are involved.  Hence
exploring control structures that leverage different computation
resources more efficiently is also important.




\begin{acknowledgements}
This work was funded by NSF grant number CNS-1329620.
\end{acknowledgements}

%


\bibliographystyle{spbasic}      
\bibliography{reference}   

%
%

\end{document}